\documentclass{dmathesis}
\usepackage{alphalph}
\usepackage[spaces,hyphens]{url}
\usepackage[colorlinks,allcolors=blue]{hyperref} %% optional
\usepackage{hyperref}
\usepackage{graphicx, subcaption}
\usepackage{hyphenat}
\usepackage{makecell}
\usepackage[dvipsnames,x11names]{xcolor}
\usepackage{rotfloat}
\usepackage{float}
\newcolumntype{P}[1]{>{\centering\arraybackslash}p{#1}}
\newcolumntype{M}[1]{>{\centering\arraybackslash}m{#1}}
\usepackage[ noend,ruled, linesnumbered]{algorithm2e}
\usepackage{mathtools}
\usepackage{amsmath}
\usepackage{amssymb}
\usepackage{bm}
\usepackage{array}
\usepackage[utf8]{inputenc}
\usepackage{soul}
\usepackage{textcmds}
\usepackage{csquotes}
\usepackage[dvipsnames]{xcolor}
\usepackage{color}

\usepackage[english]{babel}

\PassOptionsToPackage{unicode}{hyperref}

\makeatletter
\let\oldnl\nl% Store \nl in \oldnl
\newcommand{\nonl}{\renewcommand{\nl}{\let\nl\oldnl}}% Remove line number for one line
\makeatother
\usepackage{xcolor,colortbl}
\newcommand{\rednote}[1]{\textcolor{red}{1}}
\restylefloat{figure}
\usepackage[squaren]{SIunits}
\usepackage{amsmath}
\usepackage{emptypage}
\usepackage{hyperref}
\usepackage{cite}
\usepackage{filecontents}
\usepackage{longtable}
\usepackage{listings}
\usepackage{color}
\usepackage{multirow}
\usepackage[resetlabels]{multibib}
\usepackage{etoolbox}
\usepackage{float}
\usepackage{amsmath}

\usepackage[T1]{fontenc}  % access \textquotedbl
\usepackage{textcomp}     % access \textquotesingle
\usepackage{textcmds}

\usepackage{ragged2e}
\usepackage{graphicx}
\usepackage[capitalise]{cleveref}

\usepackage{tikz}
\usepackage{booktabs}
\usetikzlibrary{calc}

\newcommand{\tikzmark}[1]{\tikz[overlay,remember picture] \node (#1) {};}

\usepackage[most]{tcolorbox}

\usepackage{graphicx}
\apptocmd{\thebibliography}{\csname phantomsection\endcsname\addcontentsline{toc}{chapter}{\bibname}}{}{}
\newcites{ref}{Bibliography}
\newcites{pub}{List of Publication}
\usepackage{url}
\usepackage{ragged2e}
\usepackage[font={footnotesize}]{caption}
\usepackage[export]{adjustbox}
\usepackage{csquotes}%For quotation marks
\usepackage{datetime}
\newdateformat{monthyeardate}{%
  \monthname[\THEMONTH] \THEYEAR}

\definecolor{codegreen}{rgb}{0,0.6,0}
\definecolor{codegray}{rgb}{0.5,0.5,0.5}
\definecolor{codepurple}{rgb}{0.58,0,0.82}
\definecolor{backcolour}{rgb}{0.95,0.95,0.92}
\lstdefinestyle{mystyle}{
    backgroundcolor=\color{backcolour},   
    commentstyle=\color{codegreen},
    keywordstyle=\color{magenta},
    numberstyle=\tiny\color{codegray},
    stringstyle=\color{codepurple},
    basicstyle=\footnotesize,
    breakatwhitespace=false,         
    breaklines=true,                 
    captionpos=b,                    
    keepspaces=true,                 
    numbers=left,                    
    numbersep=3pt,                  
    showspaces=false,                
    showstringspaces=false,
    showtabs=false,                  
    tabsize=2
}
\usepackage[ruled, linesnumbered]{algorithm2e}
\usepackage{paracol}

\newcolumntype{E}{>{\collectcell\ColCell}m{0.35cm}<{\endcollectcell}}  %Cell width

\hypersetup{urlcolor=black, colorlinks=true, citecolor=darkgray, linkcolor=black,}

\usepackage{fancyhdr}
\usepackage{epsfig}
\usepackage{cite}
\usepackage{float}
\usepackage{graphicx}
\usepackage{color}
\usepackage{amsmath}
\usepackage{theorem}
\usepackage{amssymb}
\usepackage{latexsym}
\usepackage{epic}
\usepackage{hyperref}
\usepackage{caption}
\usepackage{subcaption}
\usepackage{listings}
\usepackage{paralist}
\usepackage{verbatim}
\usepackage{acro}
\usepackage{url}
\usepackage{longtable}
\usepackage{upgreek}

\usepackage[acronym,nonumberlist,shortcuts]{glossaries}
\makeglossaries

\newcounter{ind}
\def\eqlabon{
\setcounter{ind}{\value{equation}}\addtocounter{ind}{1}
\setcounter{equation}{0}
\renewcommand{\theequation}{\arabic{chapter}%
         .\arabic{section}.\arabic{ind}\alph{equation}}}
\def\eqlaboff{
\renewcommand{\theequation}{\arabic{chapter}%
         .\arabic{section}.\arabic{equation}}
\setcounter{equation}{\value{ind}}}
{\theorembodyfont{\rmfamily}}
{\theorembodyfont{\rmfamily}}
{\theorembodyfont{\rmfamily}}
{\theorembodyfont{\rmfamily}}
{\theorembodyfont{\rmfamily}}
{\theorembodyfont{\rmfamily}}
{\theorembodyfont{\rmfamily}}

\usepackage[acronym]{glossaries}
\usepackage[xindy]{imakeidx}
\usepackage{array}
\usepackage{setspace}
\usepackage{threeparttable} % Add this to your preamble if not already present
\newacronym{fca}{FCA}{Formal Concept Analysis}

\newacronym{rmse}{RMSE}{Root Mean Square Error}

\makeglossaries
\makeindex
\begin{document}

\pagenumbering{roman}
%\pagenumbering{arabic}

\setcounter{page}{1}

\newpage

\thispagestyle{empty}
\setcounter{page}{0}
\vglue 0in
\begin{center}
    % {\Large{\bf A Deep RL based framework for targeted white matter tractography}}\\[7ex]
    {\Large{\bf A Deep RL based Framework for Targeted White Matter Tractography}}\\[7ex]
    {\normalsize \textit{A Thesis Submitted}}\\[1ex]
    {\normalsize in Accordance with the requirements}\\[1ex]
    {\normalsize for the degree of}\\[4ex]
    {\large {\bf {\em{MTECH BY RESEARCH}} }}\\[2ex]
    {\normalsize  \textit{By}}\\[2ex]
    {\large{\bf Ankita~Joshi}}\\[1ex]
    {\normalsize{(\textit{S22041})}}\\[4ex]
    % {\large{Supervisor}}\\[2ex]
    % {\large	\textbf{Dr.~Aditya~Nigam}}\\[4ex]

\end{center}

\vglue 2ex plus 1fill  % fill to keep bottom of page from creeping up
\vglue 4ex
\centerline{\includegraphics[scale=0.3]{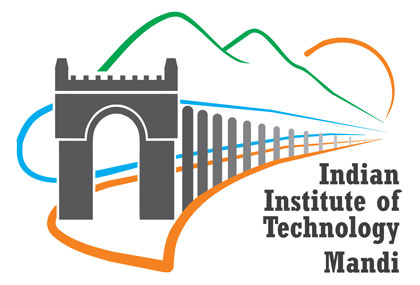}}%[width=3.5cm]

\begin{center}
{\normalsize \textit{to the}}\\[2ex]
{\large \bf SCHOOL OF COMPUTING AND ELECTRICAL}\\[1ex]
{\large \bf ENGINEERING}\\[1ex]
{\large \bf INDIAN INSTITUTE OF TECHNOLOGY MANDI}\\[1ex]
{\bf December, 2025}
\end{center}
\clearpage
\thispagestyle{empty}
%~\clearpage
 %\includegraphics[width=0.4\linewidth]{images/logo.jpg}
%%%%%%%%%%%%%%%%%%%%%%%%%%%%%%%%%%%%%%%%%%%%%%%%%%%%%%%%%%%%%%%%%%%%%%%%%%%
%%%%%%%%%%%%%%%%%%%%%%%%%%%%% Blank Page  %%%%%%%%%%%%%%%%%%%%%%%%%%%%%%%%%
%%%%%%%%%%%%%%%%%%%%%%%%%%%%%%%%%%%%%%%%%%%%%%%%%%%%%%%%%%%%%%%%%%%%%%%%%%%
\vspace*{\fill}
\begin{center}
    \thispagestyle{empty}
    Copyright \textcopyright~Ankita Joshi\\
    Indian Institute of Technology Mandi \\ 
    175005 - INDIA\\
    All rights reserved \\
    2025
\end{center}
\vspace*{\fill}

% \newpage
% \thispagestyle{empty}
% \begin{center}
%     \vspace*{10cm}
%     \textit{\large {I dedicate this thesis to:}}\\
    
%     {\large My parents}% My Guide and my parents
    
% \end{center}
\clearpage
\thispagestyle{empty}
\newpage

\thispagestyle{empty}
\setcounter{page}{0}
\vspace*{2cm}
\centerline{\Large \bf Declaration by the Research Scholar}
\vspace*{2cm}
I hereby declare that the entire work embodied in this thesis entitled \textbf{\qq{A Deep RL based Framework for Targeted White
Matter Tractography}} is the result of investigations carried out by me in the School of Computing and Electrical Engineering, Indian Institute of Technology Mandi, Mandi, H.P., India, under the supervision of \textbf{Dr.~Aditya~Nigam}, Associate Professor, School of Computing and Electrical Engineering, Indian Institute of Technology, Mandi, Mandi, India.

I also declare that it has not been submitted elsewhere for the award of any degree or diploma. In keeping with general practice, due acknowledgments have been made wherever the work described is based on the findings of other investigators. Any omissions that might have occurred due to oversight or error in judgment are regretted. \\[2.5cm]

\noindent \begin{minipage}[c]{0.4\columnwidth}
Date: 09/12/2025\\
Place: IIT Mandi, \\
Himachal Pradesh, India
\end{minipage}
\noindent \begin{minipage}[c]{0.6\columnwidth}
\begin{flushright}
	\textbf{Ankita Joshi,}\\
    Enrollment No.: S22041\\
    School of Computing and Electrical Engineering,\\
    Indian Institute of Technology Mandi,\\
    Mandi, Himachal Pradesh, India - 175075.
\end{flushright}
\end{minipage}

\clearpage
\thispagestyle{empty}

\clearpage
\thispagestyle{empty}
%~\clearpage

\newpage
 
 \thispagestyle{empty}
\setcounter{page}{0}
\vspace*{2cm}
\centerline{\Large \bf Thesis Certificate}
\vspace*{2cm}
This is to certify that the thesis titled \textbf{\qq{A Deep RL based Framework for Targeted White Matter Tractography}}, submitted by \textbf{Ankita Joshi}, Enrollment No. S22041, to \textbf{Indian Institute of Technology Mandi}, for the award of the degree of \textbf{Master of Technology by research}, is a bona fide record of the research work done by him under our guidance. The contents of this thesis, in full or in parts, have not been submitted to any other institute or university for the award of any degree or diploma. \\[2.25cm]

\noindent \begin{minipage}[c]{0.5\columnwidth}
\textbf{Dr.~Aditya~Nigam} (Guide)\\
    % Research Guide\\
    Associate Professor\\
    School of Computing and \\Electrical Engineering,\\
    Indian Institute of Technology Mandi,\\
    Himachal Pradesh, India - 175005.
\end{minipage}
\noindent \begin{minipage}[c]{0.5\columnwidth}
\begin{flushright}
	\textbf{Dr. Arnav Bhavsar} (Co-guide)\\
    % Research Co-guide\\
    Professor\\
    School of Computing and \\Electrical Engineering,\\
    Indian Institute of Technology Mandi,\\ 
    Himachal Pradesh, India - 175075.
\end{flushright}
\end{minipage}

%\vspace{1.75cm}
%\noindent \begin{minipage}[c]{0.4\columnwidth}
%Date: \\
%Place: IIT Mandi, \\
%Himachal Pradesh, India
%\end{minipage}
%\noindent \begin{minipage}[c]{0.6\columnwidth}
%\begin{flushright}
%	\textbf{Dr. Hitesh Shrimali}\\
%    Associate Professor\\
%    School of Computing and Electrical Engineering,\\
%    Indian Institute of Technology Mandi,\\
%    Mandi, Himachal Pradesh, India - 175075.
%\end{flushright}
%\end{minipage}

\clearpage
\thispagestyle{empty}

%\newpage
%\thispagestyle{empty}
%\begin{center}
%    \vspace*{2cm}
%\end{center}
%%%%%%%%%%%%%%%%%%%%%%%%%%%%%%%%%%%%%%%%%%%%%%%%%%%%%%%%%%%%%%%%%%%%%%%%%%%
%%%%%%%%%%%%%%%%%%          The Certificate page         %%%%%%%%%%%%%%%%%%
%%%%%%%%%%%%%%%%%%%%%%%%%%%%%%%%%%%%%%%%%%%%%%%%%%%%%%%%%%%%%%%%%%%%%%%%%%%

%
%\newpage
%\thispagestyle{empty}
%\begin{center}
%    \vspace*{2cm}
%\end{center}

%%%%%%%%%%%%%%%%%%%%%%%%%%%%%%%%%%%%%%%%%%%%%%%%%%%%%%%%%%%%%%%%%%%%%%%%%%%
%%%%%%%%%%%%%%%%%%     The acknowledgments page         %%%%%%%%%%%%%%%%%%
%%%%%%%%%%%%%%%%%%%%%%%%%%%%%%%%%%%%%%%%%%%%%%%%%%%%%%%%%%%%%%%%%%%%%%%%%%%
 \chapter*{Acknowledgements}

I express my deepest gratitude to my supervisor, Dr. Aditya Nigam, for his continuous support, insightful feedback, and encouraging words throughout the course of this research. His unwavering support and guidance have been instrumental in this journey.
I am also sincerely grateful to my co-supervisor, Prof. Arnav Bhavsar, for his support and valuable inputs that contributed significantly to shaping this work. I am also grateful to Dr. Ranjeet Ranjan Jha for his support and guidance which greatly helped me navigate the challenges of research with clarity and confidence.

I would like to thank Anoushkrit Goel for being a constant source of encouragement and belief. His unwavering support made this journey far more enriching.
I am also thankful to Ashutosh Sharma for being a dedicated and inspiring teammate, making our collaboration both smooth and productive.
I am deeply grateful to the faculty members of IIT Mandi, particularly Prof. Dileep AD and Dr. Erwin Fuhrer, for their inspiring teaching and for fostering a challenging yet intellectually enriching environment.
I also wish to acknowledge Geetanjali Sharma for being a wonderful friend and a strong support system throughout my time at IIT Mandi. My heartfelt thanks also go to my labmates and friends — Sushovan Jena, Shilpa Chandra, Soma Ma’am, Abhishek Tandon, and Munish Daroch — for their camaraderie and support.
Lastly, I am profoundly thankful to my family for their unconditional love, support, and belief in me, which have been the true foundation of all my endeavours.

\begin{minipage}[r]{0.95\columnwidth}
\begin{flushright}
	\textit{\footnotesize - Ankita Joshi}   
\end{flushright}
\end{minipage}

\newpage
\thispagestyle{empty}

%%%%%%%%%%%%%%%%%%%%%%%%%%%%%%%%%%%%%%%%%%%%%%%%%%%%%%%%%%%%%%%%%%%%%%%%%%%
%%%%%%%%%%%%%%%%%%           The abstract page         %%%%%%%%%%%%%%%%%%%%
%%%%%%%%%%%%%%%%%%%%%%%%%%%%%%%%%%%%%%%%%%%%%%%%%%%%%%%%%%%%%%%%%%%%%%%%%%%
\newpage
\thispagestyle{empty}
%\vspace{-2cm}
 \chapter*{\center Abstract}
\addcontentsline{toc}{chapter}{Abstract}
%\begin{center}
   % \textbf{\LARGE Abstract}
%\end{center}
\vspace{-1cm}
Fiber tractography is a computational technique that estimates the three-dimensional trajectories of white-matter fibers by utilizing directional information derived from diffusion MRI. 
Building on this ability to reconstruct the brain’s structural pathways, tractography has become a crucial component of modern neuroimaging, enabling detailed, non-invasive mapping of structural connectivity and supporting a wide range of neurological research and clinical applications.

However, despite its importance, tractography remains a challenging task due to the inherent complexity of white matter structure and its susceptibility to false positives, which can lead to the misrepresentation of critical pathways.

To overcome these limitations, recent approaches have increasingly turned to deep learning, particularly supervised learning and reinforcement learning (RL). While supervised learning requires precisely annotated ground-truth fibers which are often unavailable or unreliable, RL offers the advantage of learning policies without such annotations. Despite the benefits, RL methods still face challenges in terms of performance and generalization, motivating the need for novel strategies to enhance their effectiveness.

In this thesis, we propose a hybrid framework that integrates reinforcement learning with supervised learning for refining RL policies, specifically tailored for tract-specific tractography. Notably, our framework does not rely on ground-truth fibers for training. Moreover, the tract-specific formulation bypasses the need for an explicit segmentation process, simplifying the overall pipeline.

Our work includes two main contributions, each building upon the previous. \textbf{First}, we introduce a \textit{hybrid} approach that combines reinforcement learning with supervised learning (specifically, GPT-based policy learning) to refine policies in a tract-specific context. 
\textbf{Second}, we propose a \textit{scalable} framework for data-driven multi-policy fusion, which leverages the complementary strengths of multiple RL policies to improve tractography performance and robustness.

We demonstrate the effectiveness of our framework through extensive validation on benchmark public datasets including \textbf{TractoInferno}, \textbf{HCP}, and \textbf{ISMRM-2015}, highlighting its ability to \textit{generalize} across data sources and \textit{accurately} reconstruct brain white matter tracts.
We believe that these contributions represent significant advancements in the field of tractography, improving \textit{robustness}, \textit{reliability}, and \textit{accuracy} while reducing dependence on ground-truth annotations.

\tableofcontents

\newacronym{gcd}{GCD}{Greatest Common Divisor}

\newacronym{lcm}{LCM}{Least Common Multiple}
\listoffigures 
\addcontentsline{toc}{chapter}%{\numberline{}List of Figures}
{List of Figures}
\listoftables
\addcontentsline{toc}{chapter}%{\numberline{}List of Tables}
{List of Tables}

%\printglossaries
%\printacronyms

%\include{abbreviation.tex}
%\printglossari
%\addcontentsline{toc}{chapter}{Abbreviations}

%\clearpage
%\thispagestyle{empty}
%~\clearpage
%\printglossaries

%\setcounter{equation}{0}
\chapter*{Abbreviations}
\addcontentsline{toc}{chapter}{Abbreviations}

\begin{center}
% \begin{tabular}{ll}
\begin{longtable}{lll}
        DWI & Diffusion Weighted Image & \hyperlink{page.4}{Pg. 4} \\
        MRI & Magnetic Resonance Imaging & \hyperlink{page.1}{Pg. 1} \\
        CT & Computed Tomography & \hyperlink{page.1}{Pg. 1} \\
        PET & Positron Emission Tomography & \hyperlink{page.1}{Pg. 1} \\
        fMRI & Functional Magnetic Resonance Imaging & \hyperlink{page.2}{Pg. 2} \\
        dMRI & Diffusion Magnetic Resonance Imaging & \hyperlink{page.2}{Pg. 2} \\
        HARDI & High Angular Resolution Diffusion Imaging & \hyperlink{page.5}{Pg. 5} \\
        DTI & Diffusion Tensor Imaging & \hyperlink{page.4}{Pg. 4} \\
        AD & Axial Diﬀusivity & \hyperlink{page.6}{Pg. 6} \\
        FA & Fractional Anisotropy & \hyperlink{page.6}{Pg. 6} \\
        MD & Mean Diffusivity & \hyperlink{page.6}{Pg. 6} \\
        RD & Radial Diffusivity & \hyperlink{page.6}{Pg. 6} \\
        TBI & Traumatic Brain Injury & \hyperlink{page.6}{Pg. 6} \\
        fODF & Fiber Oriented Distribution Function & \hyperlink{page.5}{Pg. 5} \\
        CSD & Constrained Spherical Deconvolution & \hyperlink{page.4}{Pg. 4} \\
        RL & Reinforcement Learning & \hyperlink{page.13}{Pg. 13} \\
        DRL & Deep Reinforcement Learning & \hyperlink{page.13}{Pg. 13} \\ 
        DDPG & Deep Deterministic Policy Gradient & \hyperlink{page.15}{Pg. 15} \\
        TD3 & Twin Delayed Deep Deterministic Policy Gradient & \hyperlink{page.15}{Pg. 15} \\
        SAC & Soft Actor Critic & \hyperlink{page.15}{Pg. 15} \\
        GPT & Generative Pretrained Transformer & \hyperlink{page.15}{Pg. 15} \\
        HCP & Human Connectome Project & \hyperlink{page.36}{Pg. 36} \\
        ISMRM & International Society for Magnetic Resonance in Medicine & \hyperlink{page.36}{Pg. 36} \\
        T-RLF & Tract-RLFormer & \hyperlink{page.37}{Pg. 37} \\
        AF & Arcuate Fasciculus & \hyperlink{page.37}{Pg. 37} \\
        CC & Corpus Callosum & \hyperlink{page.37}{Pg. 37} \\
        CST & Cortico-Spinal Tract & \hyperlink{page.37}{Pg. 37} \\
        PYT & Pyramidal Tract & \hyperlink{page.37}{Pg. 37} \\
        OR & Optical Radius & \hyperlink{page.47}{Pg. 47} \\
        EDS & Episodic Data Selection & \hyperlink{page.58}{Pg. 58} \\
        MCPFT & Multi-critic Policy Finetuning & \hyperlink{page.58}{Pg. 58} \\

\end{longtable}
\end{center}
\newpage
\thispagestyle{empty}

% \chapter*{Abbreviations}
% SQL      &- Structural Query Language \\
% POS      &- Part of Speech \\

%\newpage
%\vspace*{4cm}
% \begin{center}
% \begin{tabular}{ll}

%         SI      &- Silhouette Index \\
%         DTs    &- Description Topics  \\
%         HAC     &- Hierarchical Agglomerative Clustering \\
%         ML    &- Machine Learning \\
%         DL      &- Deep Learning \\
%         HDFS     &- Hadoop Distribute File System \\
%         YARN      &- Yet Another Resource Negotiator \\
%         DLB       &- Dynamic Load Balancing \\
%         SLB      &- Static Load Balancing \\
%         BFS       &- Breadth-First-Search \\
%         MR-Ganter+      &- Map-Reduce-Ganter+\\
%         minTh       &- Minimum Threshold \\
%         TN     &- True Negative \\
%         FP      & - False Positive \\
%         TP      &- True Positive \\
%         LSB     &- Least Significant Bit \\
%         MSB      &- Most Significant Bit \\
%         RBM      &- RoaringBitmap \\
%         minSupp    &- Minimum Support Threshold\\
%         BC       &- Border Concept \\
%         FFD     &- First-Fit-Decreasing \\
%         SLA     &- Service Level Agreement \\

% \end{tabular}
% \end{center}

%~\clearpage

%\addcontentsline{toc}{chapter}{\numberline{}Abbreviations}
\thispagestyle{empty}
\pagenumbering{arabic}

%  ~\clearpage
 % \end{pagenumbering}
\setcounter{page}{1}

\eqlabon
\eqlaboff
\chapter{Introduction}\label{introduction}

\section{Artificial Intelligence in Medical Imaging}

Artificial intelligence has found widespread application in the medical domain, spanning drug discovery and delivery, diagnosis, treatment planning, robot-assisted surgery, and guided interventions.
Advances in machine learning and deep learning approaches have contributed to improved outcomes, greater efficiency, faster treatment pipelines, better planning, and reduced costs.
Medical imaging, in particular, has become an essential part of modern healthcare, allowing doctors to study the structure and function of the human body without invasive procedures. Modalities such as MRI, CT, PET, ultrasound, and X-ray generate vast amounts of data, often far exceeding what can be feasibly analyzed through manual interpretation.
In recent years, deep learning approaches have achieved remarkable success in image classification \cite{krizhevsky2012imagenet}, detection \cite{ren2015faster}, segmentation \cite{ronneberger2015u}, and reconstruction \cite{ledig2017photo}, making them highly effective for managing these complexities. These learnable techniques provide data-driven insights that complement and augment clinical expertise, enabling a shift in the way clinicians and researchers interpret and make use of complex medical images.

\subsection{Introduction to Brain Imaging}

The human brain is a complex organ responsible for controlling a wide range of functions, including motor functions, sensory processing, cognitive operations, emotional regulation, and essential autonomic functions such as respiration and cardiovascular functions. To understand this complexity, brain imaging plays a crucial role, as it enables researchers and clinicians to visualize brain structure and function in detail. Such visualization is vital for diagnosing and studying neurological conditions, guiding treatments, and advancing neuroscience research.

\begin{figure}[h]
    \centering
    \includegraphics[width=\linewidth]{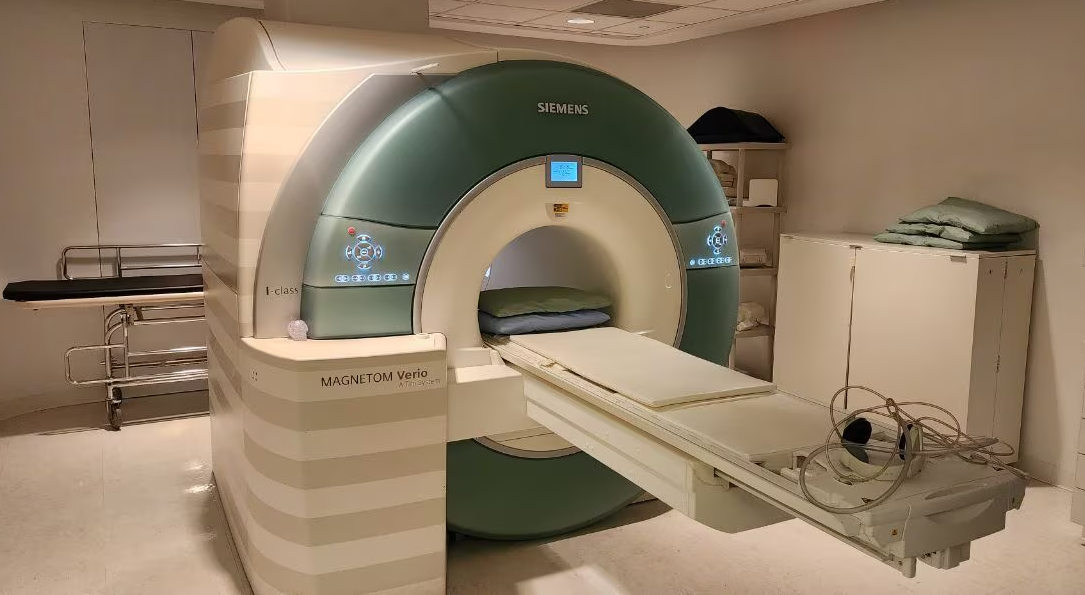}
    \caption{Siemens Magnetom Verio 3T MRI Scanner \cite{blockimaging3tmri}.} 
    % Source : https://www.blockimaging.com/blog/siemens-3t-mri-scanners-compared}
    \label{fig:dti}
\end{figure}

Different imaging techniques, such as Magnetic Resonance Imaging (MRI) \cite{hashemi2010mri}, Computed Tomography (CT) \cite{hsieh2003computed}, and Positron Emission Tomography (PET) \cite{phelps2004molecular}, provide complementary information about brain health, structure, and function.
Among these, \emph{MRI} is currently the most common imaging modality for brain tissue evaluation because of its excellent soft-tissue contrast and high spatial resolution \cite{heiss2011multimodality}. \emph{CT} scans, by contrast, are particularly important in emergency settings for rapid assessment of injuries and acute conditions such as stroke, though they offer lower resolution compared to MRI and involve radiation exposure. \emph{PET} scans, on the other hand, are used to visualize metabolic activity in the brain, aiding in the diagnosis of disorders like Alzheimer’s disease and Parkinson’s disease. However, PET also involves radiation exposure and is therefore less commonly used than functional MRI for studies of brain activity.
Within MRI, two notable techniques are \emph{Diffusion MRI (dMRI)} and \emph{Functional MRI (fMRI)}. dMRI provides insights into the structural connections of the brain, particularly white matter, by measuring the diffusion of water molecules in tissues. fMRI, in contrast, helps in understanding how different brain regions interact during specific tasks. It measures brain activity by detecting changes in blood flow associated with neural activation, often using Blood Oxygen Level Dependent (BOLD) contrast to map function. This makes fMRI particularly valuable for diagnosing conditions such as stroke, epilepsy, and brain tumors.

Over the past decade, brain imaging has undergone significant advancements, particularly through the integration of artificial intelligence (AI) and the development of multi-modal imaging  \cite{zhou2021review,kustner2024intelligent}. These innovations continue to enhance both clinical applications and research into the brain’s complex organization.

\begin{figure}[hbt!]
    \centering
    \includegraphics[width=\linewidth]{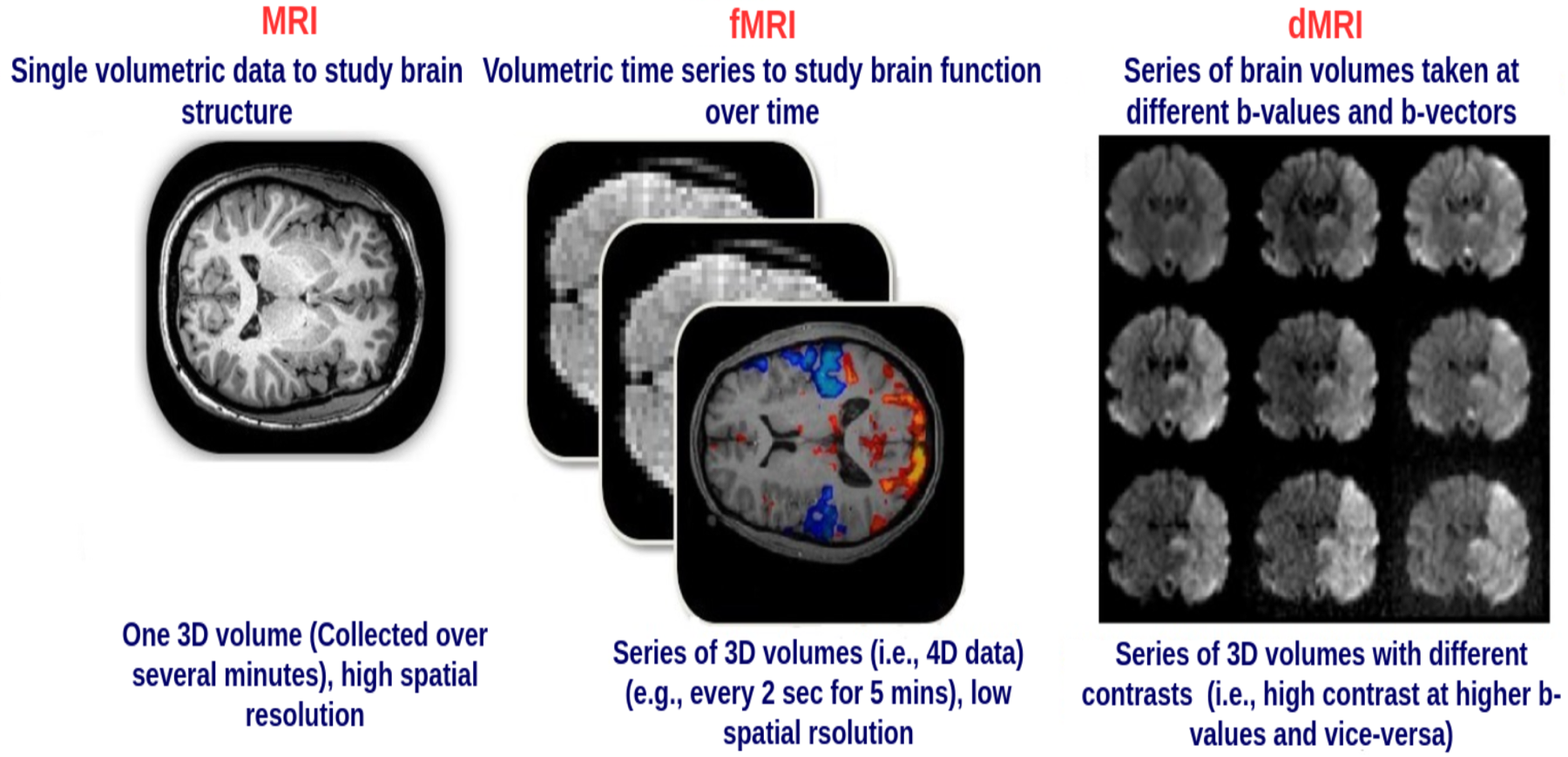}
    \caption{Overview of Structural MRI, Functional MRI (fMRI), and Diffusion MRI (dMRI) Modalities \cite{jha2020advances}}
    \label{fig:dmri}
\end{figure}

\section{Diffusion Magnetic Resonance Imaging}

Diffusion MRI (dMRI) is a non-invasive imaging technique that measures the movement of water molecules within biological tissues. Water diffusion is not completely random but is constrained by cellular structures such as membranes, fibers, and organelles, and therefore varies across tissue types. In the brain, for example, cerebrospinal fluid (CSF) exhibits free isotropic diffusion (Fig. \ref{fig:dti} (i)), grey matter shows relatively isotropic diffusion due to densely packed neuronal cell bodies, while white matter displays highly anisotropic diffusion, with water preferentially moving along axonal fibers (Fig. \ref{fig:dti} (ii)). Capturing these diffusion patterns provides estimates of the underlying tissue microstructure.

\begin{figure}[hbt!]
    \centering
    \includegraphics[width=\linewidth]{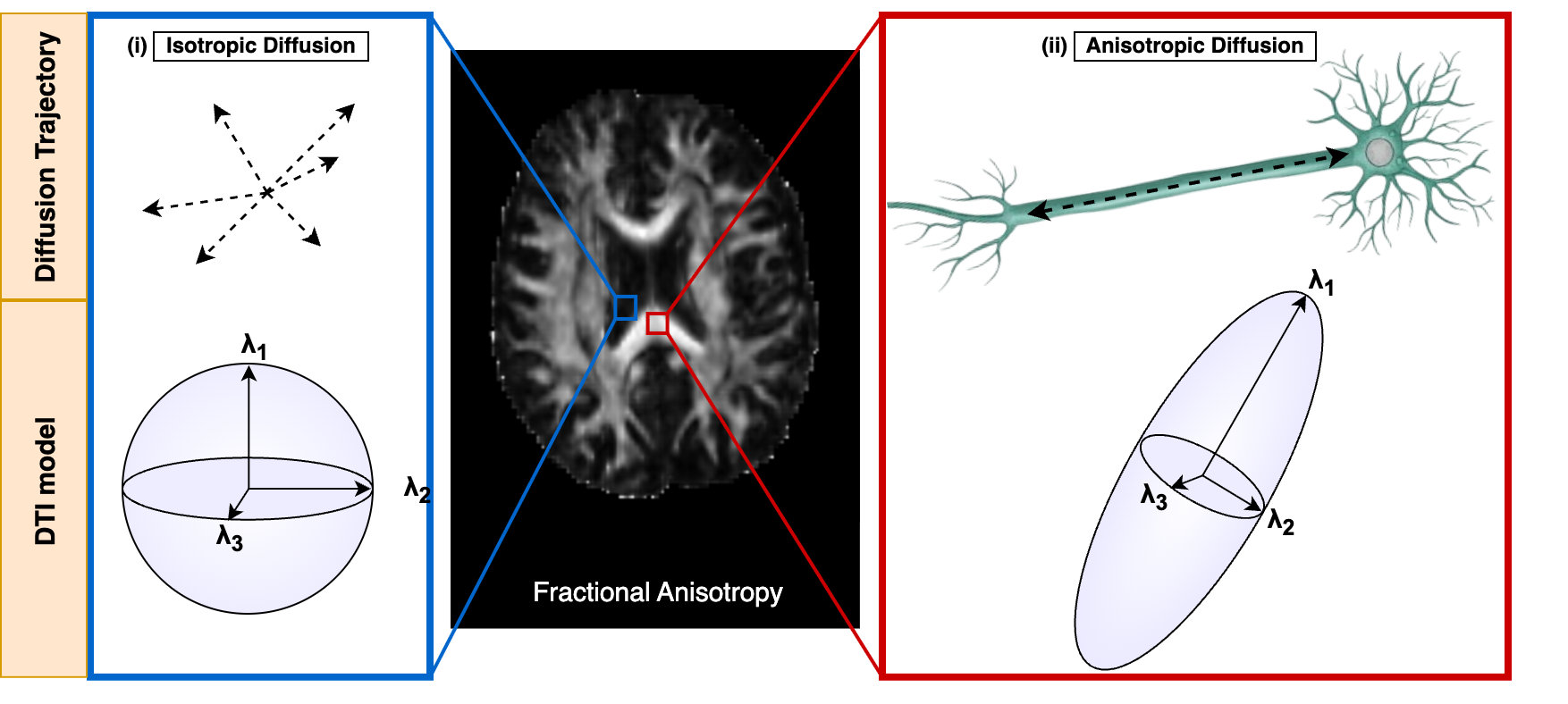}
    \caption{Illustration of (i) free isotropic movement in like CSF. (ii) depicts Anisotropic along the axon of the neuron, which happens in WM.}
    \label{fig:dti}
\end{figure}

\subsection{Acquisition}

During dMRI acquisition, the scanner applies gradients to sensitize the MRI signal to water motion. This produces Diffusion weighted images (DWIs), where contrast is weighted by the degree of water diffusion.
The degree of diffusion sensitivity is determined by the \textit{b-value}, while the gradient direction is defined by the \textit{b-vector}. dMRI can be acquired at different b-values and across multiple directions. 
The data can be represented as a 4D array with dimensions (X, Y, Z, $N_{volumes}$), where $N_{volumes}$ denotes the number of diffusion-weighted images. Each volume is associated with a specific b-value and a corresponding b-vector.  X, Y, and Z represent the spatial dimensions of the brain volume. The accompanying metadata includes b-values, b-vectors, voxel dimensions, and the affine transformation matrix.

Depending on the data (number of b-values and directions acquired), it can be further modeled to infer fiber orientation and other features using approaches such as Diffusion Tensor Imaging (DTI) \cite{basser1994mr} or Constrained Spherical Deconvolution (CSD) \cite{tournier2007robust}, which are described in later sections.

\noindent \textbf{High Angular Resolution Diffusion Imaging (HARDI): } 

HARDI is an acquisition technique in which diffusion images are sampled at around 30 or more directions (up to 100) and at one or more b-values. With HARDI data, advanced modeling methods such as CSD can be applied to estimate the fiber Orientation Distribution Function (fODF). This provides richer angular coverage of the diffusion signal compared to DTI, enabling the resolution of complex fiber structures that DTI cannot capture.  
HARDI thus enables the estimation of the fODF, offering a more accurate description of fiber orientations within a voxel, including regions of fiber crossing, branching, and fanning. It has proven particularly valuable in advanced tractography and connectomics research \cite{tuch2004q, descoteaux2007regularized}. However, HARDI requires longer acquisition times and higher signal-to-noise ratios, which can be challenging in clinical settings.

\subsection{Preprocessing}
\label{subsec:preprocessing-dmri}

Preprocessing of diffusion MRI (dMRI) data is an essential step prior to diffusion modeling, tractography, and connectivity analyses.
It is used to correct the subject motion, eddy current distortions and other artefacts present in the raw dMRI data.
\begin{itemize}
    \item The preprocessing begins with converting the raw data from scanner-specific format DICOM (industry standard) to standard post-processing analysis formats such as NIfTI (Neuroimaging Informatics Technology Initiative) or MINC.
    % \cite{vincent2016minc}.
    \item This is followed by eddy currents and subject motion correction. Eddy currents are induced by rapid gradient switching during acquisition, and can cause geometrical distortions, scaling, and voxel misalignment in the final image. Tools such as FSL Eddy\cite{andersson2016integrated} are widely used to correct these distortions and align diffusion-weighted images accurately.
    \item Bias field is a low frequency signal that causes homogenous tissues to have varying intensities in dMRI scans. These may arise from the scanner’s RF coil sensitivity or the subject’s anatomy. Correction techniques include filtering, surface fitting, segmentation, and histogram-based methods. N4ITK\cite{tustison2010n4itk} is the current state-of-the-art correction technique, integrated into neuroimaging tools such as MRtrix \cite{tournier2019mrtrix3}, SimpleITK\cite{lowekamp2013design}, ANTs\cite{avants2009advanced}, and FreeSurfer\cite{fischl2012freesurfer}.
    \item Subsequently, brain extraction (or skull stripping) is performed to remove non-brain tissues and isolate the brain volume for analysis. 
    \item A final Quality control step is conducted to identify any residual artefacts and signal dropouts, using visual examination or automated inspection tools like SQUAD (Study-wise QUality Assessment for DMRI)\cite{bastiani2019automated} and DTIPrep\cite{oguz2014dtiprep} etc. Together, these preprocessing steps help ensure data reliability for subsequent diffusion analysis and connectivity mapping.
\end{itemize}

\subsection{Modeling diffusion}
\label{modeling-dmri}

The primary dMRI modeling techniques include:

\subsubsection{Diffusion Tensor Imaging}

It is a modeling technique that characterizes water diffusion in tissue as an ellipsoid by fitting a 3D Gaussian tensor.
To estimate the diffusion tensor, at least 6 non-collinear diffusion‐weighted gradient directions are required during MRI acquisition, typically acquired at a single b-value (i.e., a single-shell scheme).

From the fitted diffusion tensor, we obtain three eigenvectors (principal diffusion directions) and their corresponding eigenvalues (\(\lambda_1, \lambda_2, \lambda_3\)) (see Fig. \ref{fig:dti} (ii)), which represent the square of magnitude of diffusion along those directions. These values are used to derive DTI metrics such as Fractional Anisotropy (FA), Mean Diffusivity (MD), Axial Diffusivity (AD), and Radial Diffusivity (RD) \cite{basser1994mr,basser2011microstructural,pierpaoli1996toward}. These metrics are widely used in research, particularly in the case of Traumatic Brain Injury (TBI). However, DTI assumes a single dominant fiber orientation per voxel, which limits its accuracy in regions with complex fiber crossings.

\subsubsection{Fiber Orientation Distribution Function}
\label{subsec:fodf}

The fiber orientation distribution function is a continuous function defined on the unit sphere that describes the probability distribution of fiber orientations within a voxel (a 3D pixel) of a diffusion MRI image.
This is crucial because older models like DTI struggled in areas where fibers cross, as they could only assign a single orientation to a voxel. 
\emph{Constrained Spherical Deconvolution} \cite{tournier2007robust} is a widely used technique to estimate fODF from the raw diffusion-weighted signal. It computes the set of spherical harmonic coefficients (SHCs) of the fODF. Spherical Harmonics are essentially the angular basis functions defined on the surface of a sphere. The spherical harmonic coefficients (SHC) provide a compact parametric representation of the fODF.
In regions of anisotropy such as white matter, the fODF exhibits distinct peaks, each corresponding to a dominant fiber orientation (local maxima of fODF) within the voxel. These principal directions are used as local features to guide the reconstruction of white matter pathways (tractography).

For a DWI image of shape ($X$, $Y$, $Z$, $N_{volumes}$), the fODF is represented by its spherical harmonic (SH) coefficients with dimensions ($X$, $Y$, $Z$, ($l$+1)($l$+2)/2). Each voxel’s SH coefficients correspond to harmonics up to degree $l$ (even integers for fODFs), which determine the angular resolution of the reconstructed distribution. Similarly, the fODF peaks are stored with dimensions ($X$, $Y$, $Z$, 3 × $N_{peaks}$), where $N_{peaks}$ are the total number of peaks and each peak is represented by a 3D vector indicating its spatial direction.

\section{White Matter Fiber Tractography}

\begin{figure}[h]
    \centering
    \includegraphics[width=13cm]{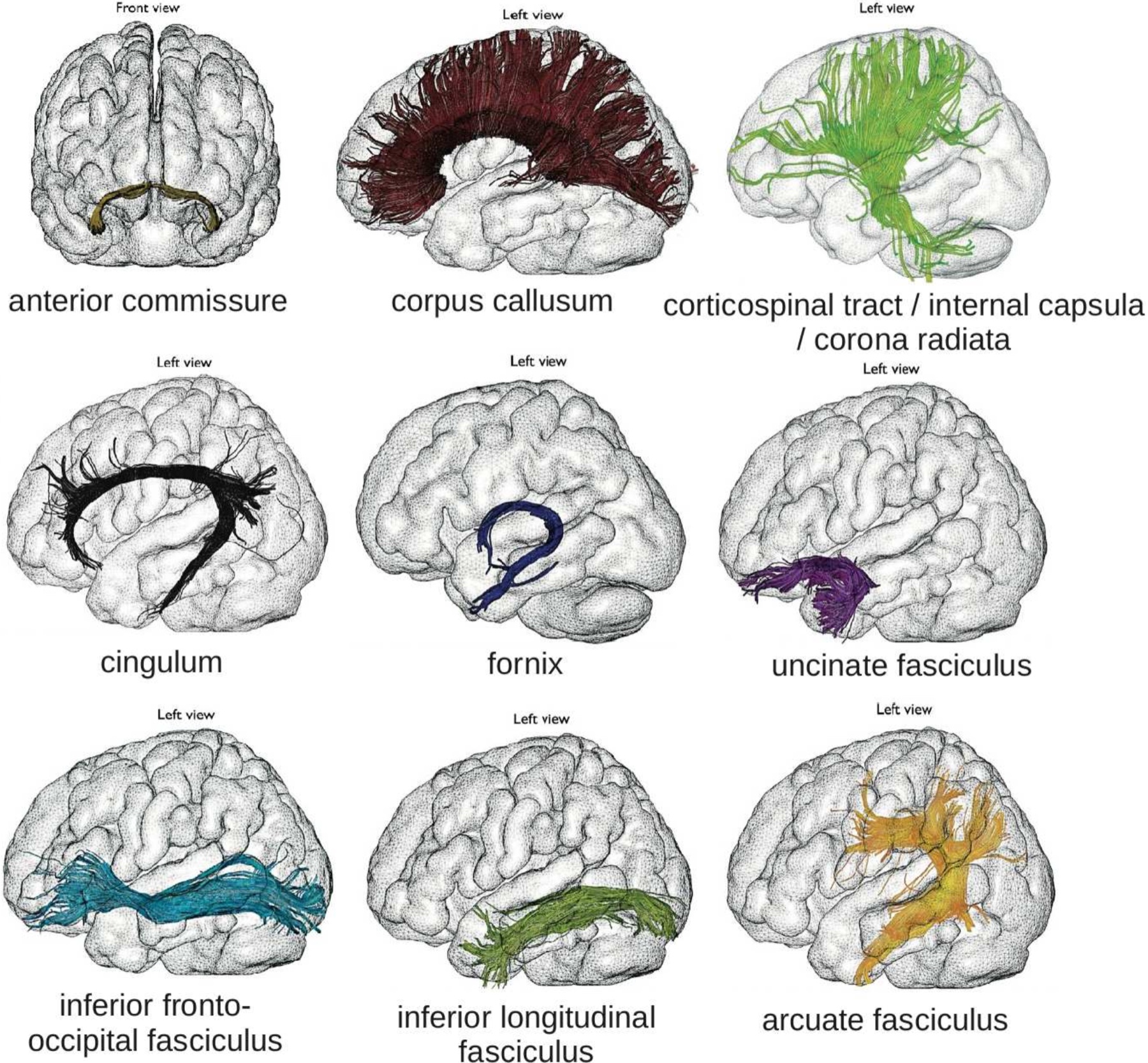}
    \caption{Major White Matter tracts \cite{alvez2011inference}.} 
    % Source : https://www.blockimaging.com/blog/siemens-3t-mri-scanners-compared}
    \label{fig:wm-tracts}
\end{figure}

White matter plays a critical role in interconnecting cortical and subcortical regions of the brain. It constitutes roughly half of the brain’s volume and is composed of myelinated axons that form coherent fiber tracts. These tracts, shown in Fig. \ref{fig:wm-tracts} ( Image taken from \cite{alvez2011inference}), enable signal transmission between different brain regions and the spinal cord.

\begin{figure}[h]
    \centering
    \includegraphics[width=\linewidth]{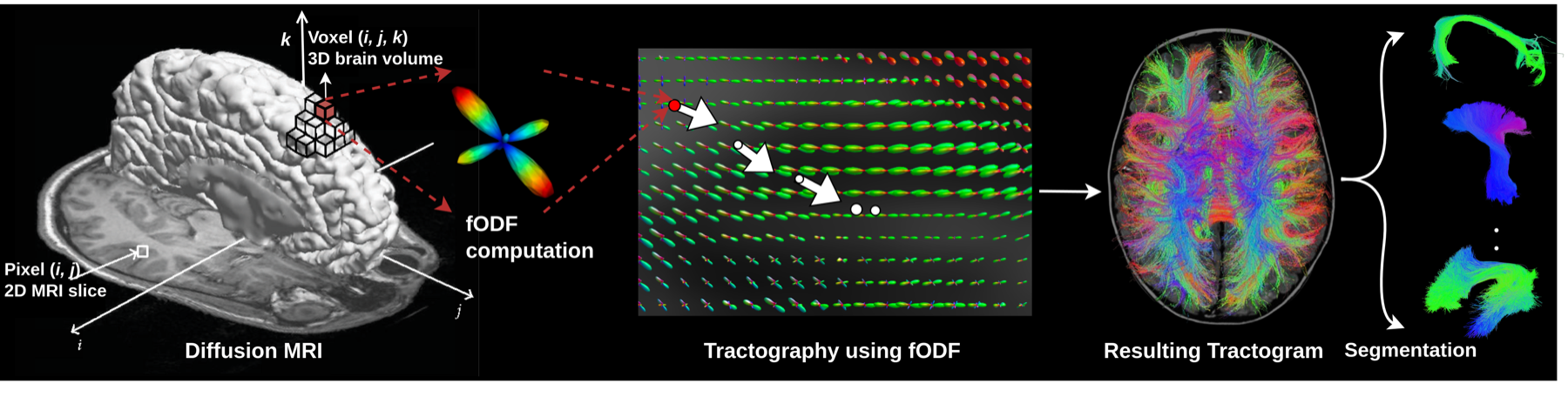}
    % fig-thesis-latesto.png}
    % {images/ch1/thesis-tracto-main-fig.png}
    % {images/ch1/tracto_thesis_fig.png}
    \caption{Tractography process} 
    % Source : https://www.blockimaging.com/blog/siemens-3t-mri-scanners-compared}
    \label{fig:tractography}
\end{figure}

Tractography is the process of mapping these white matter pathways non-invasively using diffusion MRI data. The fODF and its peaks, derived from dMRI data, represent the dominant fiber directions within each voxel and are often used as input to tractography algorithms, as shown in Fig. \ref{fig:tractography}. Building on this local information, tractography estimates long-range fiber pathways that represent global structural connectivity. 

\noindent Tractography can be understood as tracking in 3D space (brain volume), constrained by a tracking mask and anatomy. 
Seed points are placed within this mask, and at each step the algorithm predicts the next direction based on local diffusion information (eg. fiber orientation), scaled by a step size. Starting from each seed point, tracking continues until termination criteria are met, producing the reconstructed tractogram (Fig. \ref{fig:tractography}). A tractogram can be described as a collection of fiber streamlines, where each streamline is a sequence of 3D points of variable length.

\begin{figure}[h]
    \centering
    \includegraphics[width=\linewidth]{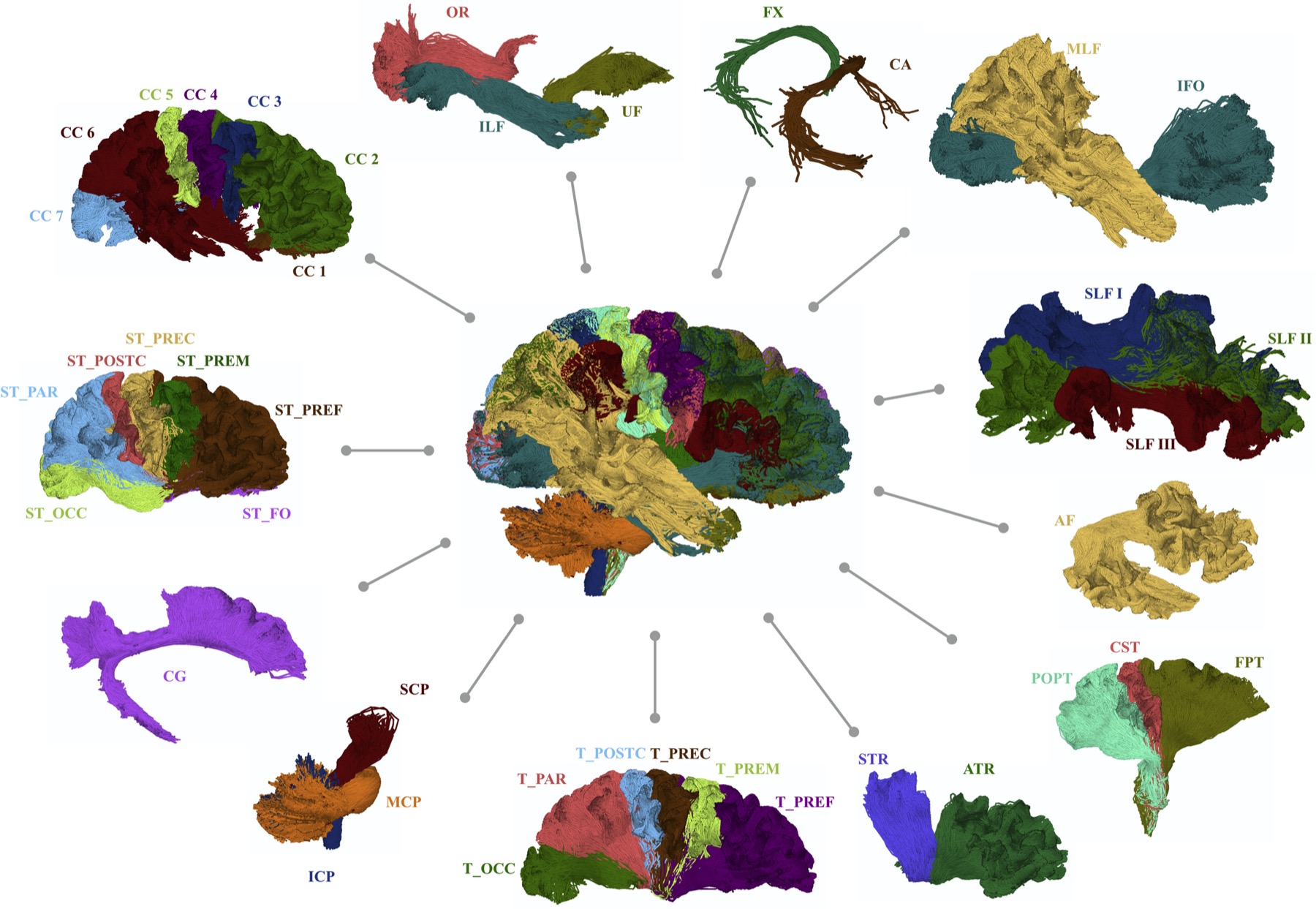}
    \caption{Whole-brain tractogram and its segmentation into anatomically meaningful white-matter tracts. Source: \cite{wasserthal2018tractseg}.}
    \label{fig:tracts-seg}
\end{figure}

After producing a whole-brain tractogram, the next step is tractogram segmentation, which groups fibers that share similar anatomical trajectories or endpoints, producing coherent tracts such as the corticospinal tract, arcuate fasciculus, corpus callosum fibers, etc. displayed in Fig. \ref{fig:tracts-seg} \cite{wasserthal2018tractseg}. By extracting individual tracts, researchers and clinicians can analyze their microstructural properties, assess tract-specific damage or displacement, and relate structural changes to functional or behavioral outcomes. Segmentation process enables tractography to support precise diagnostics, surgical planning, and targeted neuroscientific investigations.

Building on this, Tractography is widely applied in neurosurgical planning, where it allows surgeons to visualize the structure of critical white matter pathways, such as corticospinal tract, and optic radiation, and plan surgical interventions that avoid or minimize damage to these fibers.
It is also used in assessing damage from traumatic brain injury, which may cause fibers to stretch, shear or break down, resulting in reduced anisotropy and dirupted connectivity. It also aids in studying structural disruptions in neurological conditions such as Alzheimer’s and Parkinson’s disease, and is also widely used to explore altered brain connectivity patterns in tumors and psychiatric disorders. 

\subsection{Streamline generation}

Tracking begins from a set of seed points, which can be defined within specific regions of interest (ROI) or across the entire white matter volume.
From each seed, the algorithm iteratively propagates a streamline in small steps. 

At each step, the algorithm predicts the next direction and propagates the streamline by a fixed step size, or directly estimates the next point, based on voxel-wise diffusion information derived from the chosen diffusion model (e.g., diffusion tensor, fODF, SHCs, or fODF peaks).
Several stopping criteria are used to terminate a streamline. These typically include thresholds on fractional anisotropy (FA), maximum curvature angle, and anatomical boundaries. For example, tracking is commonly stopped when FA drops below a predefined limit, indicating regions of isotropic diffusion, or when the angular deviation between successive steps exceeds a threshold, implying an improbable fiber turn.
Fig.~\ref{fig:tractography} shows the workflow of a typical tractography algorithm. It also includes a segmentation step, which is commonly performed to extract specific anatomical tracts from the whole-brain tractogram.

\subsection{Challenges}

Tractography has been described as an ill-posed problem due to the inability of diffusion MRI to uniquely determine the underlying white-matter pathways \cite{maier2017challenge}. This is because multiple anatomically distinct fiber configurations can produce similar voxel-wise diffusion signals, making it impossible to distinguish between these cases purely at the voxel level. This uncertainty at the voxel level can propagate through tractography algorithms, leading to errors in resulting pathways, such as missing true connections (False negatives) or generating spurious streamlines (False positives) \cite{jeurissen2019diffusion}.

\begin{figure}[h]
    \centering
    \includegraphics[width=10cm]{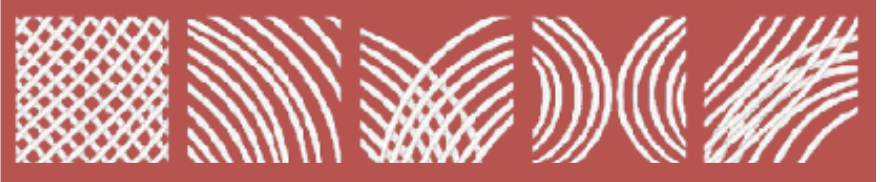}
    \caption{Illustration of common complex fiber configurations, including crossing, branching, and fanning pathways.}
    \label{fig:crossing-fibs}
\end{figure}

Moreover, regions with complex fiber architectures (e.g. crossing, branching, or fanning fibers), shown in Fig. \ref{fig:crossing-fibs} are very prone to errors (false positive and false negative connections). In these cases, algorithmic and methodological choices also play a crucial role. Deterministic tractography methods follow a single, most likely path at each point and can therefore fail to capture branching or crossing fibers, resulting in a large number of false negatives. Whereas probabilistic algorithms generate multiple potential fiber directions, which potentially lead to false positive connections. Deciding the optimal balance between sensitivity and specificity remains a persistent issue \cite{schilling2019challenges}.

% #kross

Furthermore, choices of angular threshold, step size, curvature constraints, fractional anisotropy (FA) and other parameters critically affect tracking results, leading to large inter-study variability. Finally, validation and reproducibility present ongoing challenges. The absence of a definitive in-vivo ground truth for human brain connectivity \cite{poulin2019tractography } makes it difficult to assess the accuracy of tractography results, highlighting the need for more reliable validation methods.

\subsection{Related Work: Tractography Techniques}

Tractography Techniques can be broadly classified into two main approaches: deterministic and probabilistic tracking.
Deterministic methods \cite{mori1999three,basser2000vivo} follow the most likely diffusion direction at each step, producing precise but potentially less robust reconstructions in areas of complex fiber orientations. In contrast, probabilistic methods \cite{behrens2003characterization,tournier2010improved} sample from the distribution of possible orientations, generating multiple potential trajectories to account for uncertainty in the diffusion model.

\subsubsection{Classical Methods}

As discussed in section 1.2.3, before applying any tractography algorithm, local diffusion modeling techniques are first used to estimate fiber orientations within each voxel. Classical tractography methods heavily rely on such mathematical models.

Early tractography approaches were based on the DTI model, which estimates a single principal diffusion direction per voxel \cite{basser1994mr}. Deterministic tractography methods using the DTI model, such as \cite{mori1999three,basser2000vivo} suffer from the “crossing fiber problem,” since DTI leads to missed connections in regions where multiple fiber bundles intersect. To overcome these limitations, probabilistic tractography techniques were developed, introducing uncertainty into the local orientation estimates. Methods such as \cite{parker2003framework,behrens2003non} generate multiple possible streamlines based on the probability distribution of fiber orientations, producing lesser False positives in complex regions. The iFOD2 algorithm \cite{tournier2010improved} represents a second-order integration approach over the Orientation Distribution Functions (ODFs), improving angular accuracy and robustness to noise. Beyond local tracking, global tractography methods attempt to infer the entire set of streamlines simultaneously, optimizing for the configuration that best explains the measured DWI signal across the whole brain. Examples include Gibbs tracking \cite{kreher2008gibbs}, which formulates tractography as an energy minimization problem. Although this framework provides more globally consistent solutions, it is computationally expensive and thus less commonly used in routine applications.

Advances in diffusion imaging have enabled the acquisition of HARDI data, making it possible to employ advanced diffusion models like Q-ball Imaging \cite{tuch2004q} and Constrained Spherical Deconvolution \cite{tournier2007robust}. These models estimate Orientation Distribution Functions (ODFs) that can resolve multiple fiber directions within a voxel, unlike DTI.
As tractography techniques evolved, anatomical priors and tissue constraints were introduced to enhance biological plausibility. Anatomically Constrained Tractography (ACT) \cite{smith2012anatomically} implements this by using tissue-segmented anatomical priors to prevent streamlines from propagating into non-white matter regions. Particle Filtering Tractography (PFT) \cite{girard2014towards} further extends this approach by integrating partial volume estimates (PVEs) into the tracking process, guiding streamline propagation and termination based on the proportions of gray matter, white matter, and cerebrospinal fluid within each voxel. By incorporating anatomical constraints, ACT and PFT enhanced the biological plausibility of streamline reconstruction, bridging the gap between signal-driven and anatomically informed tractography. Even with the advent of machine learning and deep learning methods, classical tractography methods remain strong baselines. They are frequently used in ensemble frameworks for generating reference datasets or tract atlases \cite{maier2017challenge,poulin2022tractoinferno}.

\subsubsection{Machine Learning Methods}

Early learning-based tractography methods focused on supervised classification of fiber orientations. Authors in \cite{neher2015machine} utilized a Random Forest classifier in a supervised setting to identify 25 distinct fiber bundles, leveraging data from the ISMRM 2015 Tractography Challenge dataset \cite{maier2017challenge}. This approach demonstrated the feasibility of applying machine learning to white matter bundle segmentation, though it remained dependent on handcrafted features and labeled training data.
Building on this foundation, research progressed toward regression-based models for streamline tracking. \textit{Learn to Track} \cite{poulin2017learn} proposed the use of a Gated Recurrent Unit (GRU) network to predict the next tracking direction directly from diffusion-weighted signals resampled to 100 gradient directions. The recurrent design enabled modeling of temporal dependencies along streamlines.
\textit{DeepTract} \cite{benou2019deeptract} further advanced this paradigm by employing Gated Recurrent Unit (GRU) for both deterministic and probabilistic tractography. It outputs continuous orientation distribution functions (ODFs) at each step, allowing uncertainty-aware streamline propagation.
In another work, \cite{wegmayr2018data} introduced a data-driven approach using a multilayer perceptron (MLP) that takes as input the local diffusion signal from a $3 \times 3 \times 3$ voxel neighborhood, together with the last four tracking directions, thereby incorporating both spatial and temporal contextual information.
\textit{Entrack} \cite{wegmayr2021entrack} extended this concept through a probabilistic deep learning framework that predicts a full Fisher–von Mises distribution over possible tracking directions, allowing for uncertainty quantification and improved generalization across subjects.
More recently, Deep Reinforcement Learning (DRL) methods such as \textit{Track-to-Learn} \cite{theberge2021track,theberge2024matters} and \textit{TractOracle} \cite{theberge2024tractoracle} have emerged, aiming to remove the dependence on ground-truth streamlines. These approaches frame tractography as a sequential decision-making problem, where an agent learns optimal tracking policies through interaction with diffusion data, guided by anatomically informed reward signals.

\subsubsection{Tract-specific Tractography}

Tract-specific tractography aims to directly reconstruct the tract of interest, focusing on particular white matter pathways.  
Unlike whole-brain tractography, where tracking is conducted within the entire white matter (WM) mask followed by segmentation to obtain individual tracts, tract-specific methods restrict tracking to regions or orientations associated with a specific bundle.
These methods are crucial for clinical surgical planning \cite{bello2008motor}, and research focusing on the microstructural properties and precise geometry of a known white matter pathway.

\textit{TractSeg} \cite{wasserthal2018tractseg} uses Fully connected neural network (FCNN) to create a Tract Orientation Map (TOM) for a specific bundle, where each voxel contains the most likely fiber orientation for that particular bundle. Standard streamline tracking (deterministic or probabilistic) is then run only on this TOM (Wasserthal et al., 2019), bypassing the need for whole-brain tractography. 
On the other hand, \textit{Bundle-Specific Tractography (BST)} \cite{rheault2019bundle} integrates anatomical and orientational prior knowledge, derived from a template or atlas of the target bundle, directly into a probabilistic tracking algorithm (such as iFOD2 \cite{tournier2010improved} or related methods). This guidance is applied during the tracking process to enhance or suppress propagation in certain directions for a given bundle or tract.

\vspace{-0.5cm}
\section{Thesis Motivation}

Classical deterministic and probabilistic tractography methods are highly sensitive to noise, depend heavily on diffusion modeling assumptions, and often produce anatomically implausible or incomplete pathways. Supervised learning approaches, though more data-driven, rely on ground-truth streamlines that are themselves uncertain or unavailable, leading to propagation of labeling biases. 

Deep Reinforcement Learning has recently emerged as a promising alternative by framing tractography as a sequential decision-making process, where an agent learns optimal tracking policies through interaction with diffusion signals rather than supervision from predefined tracts. However, existing DRL-based tractography methods have immense scope for improvement and still face the persistent \textit{sensitivity–specificity} trade-off, where agents either fail to fully capture the extent of the tract or erroneously extend into neighboring tracts or non-white matter regions (overreach).

This thesis addresses these challenges through two key contributions that improve the tractography performance of DRL policies and introduce a policy fusion approach to mitigate the sensitivity–specificity (overlap–overreach) trade-off. Furthermore, the method is designed to be \textit{tract-specific}, directly reconstructing individual bundles of interest, thus eliminating the need for whole-brain tractography followed by segmentation.

\section{Thesis Objectives and Contributions}

The primary aim of this thesis is to address the challenges of applying deep reinforcement learning (DRL) to tractography by improving the performance and robustness of various RL-based tracking policies, such as TD3, SAC, and DDPG, and by fusing their complementary strengths to overcome existing limitations in fiber tractography.
The main contributions of this research are summarized as follows:

\begin{itemize}
    \item We present \textbf{Tract-RLFormer}, a hybrid framework for RL policy refinement in tractography, leveraging GPT-based modeling and deep RL.
    \item We introduce \textbf{TractRLfusion}, a novel RL policy fusion framework for tractography that combines multiple policies to achieve superior performance by balancing the specificity-sensitivity trade-off.
    \item The proposed tract-specific frameworks eliminate the need for a segmentation step. A \textbf{Mask Refinement Module (MRM)} generates the tracking regions of interest (masks). Furthermore, training is conducted without ground-truth annotations and achieve strong generalization performance.

\end{itemize}

\section{Thesis Organization}

This chapter provides an overview of the field of brain imaging, with a particular focus on diffusion MRI-based white matter fiber tractography. It also discusses the motivation behind this thesis. The remainder of the thesis is organized as follows:

\begin{itemize}
    \item \textbf{Chapter 2:} A review of the deep learning methods utilized in this work.
    \item \textbf{Chapter 3:} Introduction of a tract-specific framework that refines RL policies for tractography, thereby improving their tracking performance.

    \item \textbf{Chapter 4:} Presentation of a data-driven fusion strategy that integrates multiple RL policies to improve tractography outcomes.

    \item \textbf{Chapter 5:} Summary of the key contributions of this thesis and a discussion of potential directions for future work.
\end{itemize}
 
\chapter{Preliminaries}\label{prelim}
This chapter presents an overview of the deep learning and deep reinforcement learning foundations relevant to this thesis. We begin by introducing the major learning paradigms, followed by a detailed explanation of the reinforcement learning (RL) framework and the specific algorithms employed in this work. We then discuss sequence modeling and Transformer-based architectures. These preliminaries establish the conceptual basis for Chapters 3 and 4, where deep RL methods and GPT-based models are leveraged to advance the performance of RL algorithms for tractography.

The goal of Machine learning algorithms is to learn the mapping between input and output. There are several different ways to learn it, and these techniques are typically broadly classified into 3 paradigms, namely Supervised, Unsupervised and Reinforcement learning, as shown in Table \ref{table:learning_comparison}.
Supervised learning algorithms require labelled data (input- output examples) to train (mathematical) models. Unsupervised learning algorithms learn to identify hidden patterns, associations, or groupings within data without predefined target outputs.
Reinforcement Learning algorithms learn by exploration.

\begin{table}[h!]
\centering
\renewcommand{\arraystretch}{1.5}
\begin{tabular}{|m{2.1cm}|m{3.5cm}|m{3.5cm}|m{3.82cm}|}
\hline
\textbf{Criteria} & \textbf{Supervised Learning} & \textbf{Unsupervised Learning} & \textbf{Reinforcement Learning} \\ \hline
% \textbf{Definition} & Learns from labeled data to map inputs to outputs. & Learns patterns or structures from unlabeled data. & Learns by interacting with an environment using rewards. \\ \hline
\textbf{Type of Data} & Labeled data & Unlabeled data & Learns by interaction with environment. \\ \hline

\textbf{Goal} & Predict outcomes based on input-output mapping. & Discover hidden patterns or groupings in data. & Maximize cumulative rewards over time. \\ \hline

\textbf{Examples} & Image recognition, price prediction. & Customer segmentation, anomaly detection. & Self-driving cars, game playing(AlphaGo) \\ \hline

\textbf{Challenges} & Requires labeled data; less effective for complex tasks. & May produce less accurate results; computationally intensive. & Requires careful reward design; computationally expensive. \\ \hline
\end{tabular}
\caption{Comparison of Supervised, Unsupervised, and Reinforcement Learning}
\label{table:learning_comparison}
\end{table}

\section{Reinforcement Learning}
In this work, tractography is performed within a reinforcement learning (RL) environment, following previous work \cite{theberge2021track}. The RL formulation of tractography, along with the modifications introduced for tract-specific tracking, are described in detail in Section \ref{subsec:TRLF}.

Reinforcement Learning (RL) is inspired by how humans and animals learn through trial-and-error interactions with their environment. For example, a child learning to ride a bicycle tries various ways of pedaling and balancing, receives feedback (disbalancing, falling, or staying upright), and gradually improves by repeating actions that lead to success. This process of learning from interaction and feedback forms the basis of RL, where an agent observes the state(s) of the environment, takes actions(a), and receives rewards(r) from the environment for its actions.
% based on those actions.
The agent's objective is to learn a policy($\pi$), which is a strategy to select actions that maximize cumulative rewards over time. The agent learns solely from its interactions with the environment and does not rely on external labeled data or ``ground truth''.

\begin{figure}[hbt!]
    \centering
    \includegraphics[width=10cm]{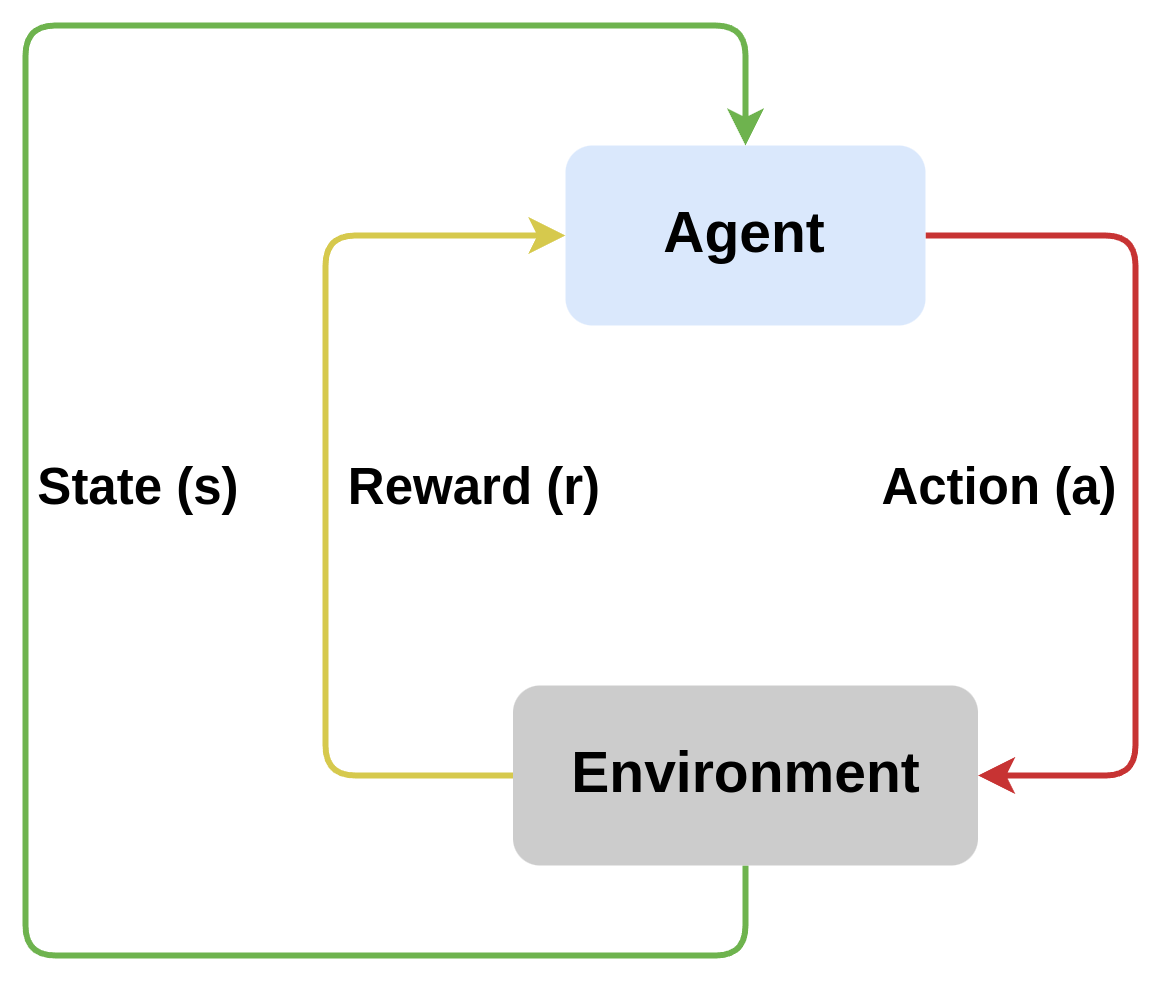}
    \caption[Illustration of a typical reinforcement learning setup]{Illustration of a typical Reinforcement Learning (RL) setup: the agent interacts with the environment by taking actions, receives feedback in the form of rewards, and observes the resulting state to learn an optimal policy.}
    \label{fig:rl}
\end{figure}

\subsection{Fundamentals of RL}

RL is formulated as a Markov Decision Process (MDP), which is a mathematical framework for modeling sequential decision-making under uncertainty. It is defined by the tuple 
\textbf{($S, A, P, R,$ $\gamma$)}. Here, \textit{S} is the set of states, \textit{A} is the set of actions, \textit{P} refers to the transition probability matrix, \textit{R} is the reward space and \textit{$\gamma$} is the discount factor. 
% The 
% Here,
% if plagiarism happens in examples, i can take a simple OpenAI gym env like cartpole and its examples.

Formalizing the definitions of key components of RL:
% using examples of a chess game/ robot navigating in a grid:
\begin{itemize}
    \item \textbf{State (S):} Represent current configuration of the environment (e.g. a robot's position on a grid).
    \item \textbf{Action (A):} Decision taken by agent that transitions it from one state to another. Action space(A) can be both discrete (e.g. choosing \emph{left} or \emph{right}) or continuous (e.g. applying a steering angle or force value), depending on the problem at hand.
    \item \textbf{Transition Probability ($P(s^{'}|s,a)$):} It provides probability of transitioning to state \textit{$s^{'}$} from state \textit{s} by taking an action \textit{a}.
    \item \textbf{Reward (R):} R(s,a,$s^{'}$) provides immediate feedback for transitioning from s to $s^{'}$ via action a.
    \item \textbf{Discount Factor ($\gamma$):} Balances immediate and future rewards ($\gamma \in [0,1]$), prioritizing short-term gains if closer to 0 and long-term if closer to 1.
\end{itemize}

The \textbf{policy} $\pi$ is a function that maps a given input state to an action. The objective of MDP is to find an optimal policy ($\pi^{*}$) that maximizes the \textbf{discounted sum of rewards} (return), denoted by:

\begin{equation}
    G_{t} = r_{t+1} + \gamma r_{t+2} + \gamma^{2}r_{t+3}+ ....
\end{equation}

where \( r_{t} \) denotes the reward received at time step \( t \), that the agent obtains after taking an action in a given state.
A sequence of interactions between the agent and the environment, consisting of states, actions, rewards, and next states is called an \textbf{episode} or a \textbf{trajectory}. Formally, an episode is represented as: $\tau = (s_0, a_0, r_1, s_1, a_1, r_2, \dots, s_T)$, where $T$ is the terminal timestep (either fixed or determined by a stopping condition). Episodes usually terminate due to environment-defined constraints, such as reaching a goal state, exceeding a time limit, or encountering a terminal condition (e.g., agent crashes or fails). For a $T$-length episode, return is defined as:

\begin{equation}
    G({\tau}) = \sum_{t=0}^{T-1} \gamma^t r_{t+1}
    \label{eqn:rl-return_eqn}
\end{equation}

This formulation also includes unknown rewards from future timesteps.
Since future rewards are uncertain (unknown due to stochastic transitions), return is a random variable that cannot be directly estimated or computed. Its value depends on the sequence of states and actions encountered, which are influenced by probabilities. Hence, the expected value of return is utilized \textbf{to} \textbf{evaluate policies}. This expectation is formalized through two functions that estimate the expected cumulative reward:

1. \textbf{State-value function:} Estimates the expected return starting from state s and following policy $\pi$.

\begin{equation}
    V^\pi(s) = \underset{\tau \sim \pi}{\mathbb{E}} \left[ G(\tau) \,\bigg|\, s_0 = s \right]
\end{equation}
Substituing the value of return from Eqn. \ref{eqn:rl-return_eqn}
\begin{equation}
    V^\pi(s) = \underset{\pi}{\mathbb{E}} \left[ \sum_{t=0}^{T-1} \gamma^t r_{t+1} \,\bigg|\, s_0 = s \right]
\end{equation}

2. \textbf{Action-value function:} Estimates expected return obtained by the agent, starting from state s and performing action a and thereafter following policy $\pi$.

\begin{equation}
    Q^\pi(s, a) = \underset{\pi}{\mathbb{E}} \left[ \sum_{t=0}^{T-1} \gamma^t r_{t+1} \,\bigg|\, s_0 = s, a_0 = a \right]
\end{equation}

The \textbf{bellman equation} decomposes the value functions into a recursive formulation of immediate rewards and discounted future values.

\begin{equation}
    V^{\pi}(s) = R(s,a,s^{'}) + \gamma V^{\pi}(s^{'})
\end{equation}

Here, R(s,a,$s^{'}$) refers to the immediate reward for transitioning from state (s) to ($s^{'}$) by taking action a. V($s^{'}$) is the value for the subsequent state $s^{'}$. 
However, the \textbf{environment can be stochastic} meaning the next state $s^{'}$ is uncertain (modeled by the transition probabilities P). Hence,

\begin{equation}
    V^{\pi}(s) = \sum_{s^{'}}P(s^{'}|s,a)[R(s,a,s^{'}) + \gamma V^{\pi}(s^{'})]
\end{equation}

Similarly, incorporating the \textbf{stochastic nature of the policy} ($\pi$) in the Bellman equation gives us:

\begin{equation}
    V^{\pi}(s) = \sum_{a}\pi(a,s) \sum_{s^{'}}P(s^{'}|s,a)[R(s,a,s^{'}) + \gamma V^{\pi}(s^{'})]
\end{equation}

Similarly, the bellman equation for Q function states that:

\begin{equation}
    Q^{\pi}(s,a) = R(s,a,s^{'}) + \gamma Q^{\pi}(s^{'},a^{'})
\end{equation}

\begin{equation}
    Q^{\pi}(s,a) = \sum_{a}\pi(a,s) \sum_{s^{'}}P(s^{'}|s,a)[R(s,a,s^{'}) + \gamma Q^{\pi}(s^{'},a^{'})]
\end{equation}

We know that action a is not chosen by policy $\pi$, only $a^{'}$ and subsequent actions are selected by $\pi$ therefore it can be rewritten as:

\begin{equation}
    Q^{\pi}(s,a) = \sum_{s^{'}}P(s^{'}|s,a)\left[R(s,a,s^{'}) + \gamma\sum_{a^{'}} \pi(a^{'},s^{'})Q^{\pi}(s^{'},a^{'})\right]
\end{equation}

% \begin{equation}
%     Q^{\pi}(s,a) = \sum_{s^{'}}P(s^{'}|s,a)[R(s,a,s^{'}) + \gamma Q^{\pi}(s^{'},a^{'})]
% \end{equation}

The optimal policy ($\pi^*$) is the one that gives the maximum state value.

% recursive

% Optimal State-Value Function
\begin{equation}
    V^*(s) = \max_\pi V^\pi(s)
\end{equation}

To find this optimal policy, \emph{Bellman optimality equation} is used. It helps to find optimal Value and Q function recursively. The Bellman equation for optimality of value function is:

\begin{equation}
    V^*(s) = \max_{a \in \mathcal{A}} \sum_{s'} P(s'|s,a) \left[ R(s,a,s') + \gamma V^*(s') \right]
\end{equation}

Similarly, the bellman optimality equation for maximum Q value, corresponding to the optimal policy ($\pi^*$) is:

\begin{equation}
    Q^*(s, a) = \sum_{s'} P(s'|s,a) \left[ R(s,a,s') + \gamma \max_{a'} Q^*(s', a') \right]
    \label{eqn:bellman-optimality-Q}
\end{equation}

By utilizing the optimal value function $(V^{*}(s^{'}))$, Bellman Optimal Q function can be expressed as: 

\begin{equation}
    Q^*(s, a) = \sum_{s'} P(s'|s,a) \left[ R(s,a,s') + \gamma V^*(s') \right]
\end{equation}

The Bellman optimality equations provide a theoretical foundation for solving MDPs. In traditional tabular reinforcement learning methods, the action-value function $Q(s, a)$ is stored in a Q-table, where each entry corresponds to a specific state-action pair. However, these methods rely on dynamic programming techniques (e.g., value iteration, policy iteration) that assume full knowledge of the environment's dynamics, i.e., $P(s^{'}|s,a)$. Moreover, Q-tables are only practical for environments with small, discrete state-action spaces. These drawbacks are addressed by employing neural networks to approximate $Q(s,a)$, forming the basis for Deep Reinforcement Learning as discussed below.

\subsection{Deep Reinforcement Learning}

Deep Reinforcement Learning (Deep RL) combines reinforcement learning with deep learning, using neural networks to approximate policies, value functions, or models of the environment. This enables RL agents to operate in high-dimensional or continuous state spaces where classical tabular methods become infeasible.

In reinforcement learning, algorithms that require environment dynamics ($P(s^{'}|s,a)$) are classified as \emph{model-based} methods. However, in real-world scenarios, these dynamics are often unknown or hard to model accurately, leading to the widespread use of \emph{model-free} methods, which learn directly from interaction with the environment.
Model-free RL learn to approximate value functions or policies. It is broadly divided into two paradigms, namely \emph{Value-Based}(Q-
learning) Methods and \emph{Policy-Based} (Policy Optimization) Methods \cite{brockman2016openai}. Some Deep RL algorithms simultaneously learn a policy and a value function, known as actor-critic methods. 
\begin{figure}[hbt!]
    \centering
    \includegraphics[width=13cm]{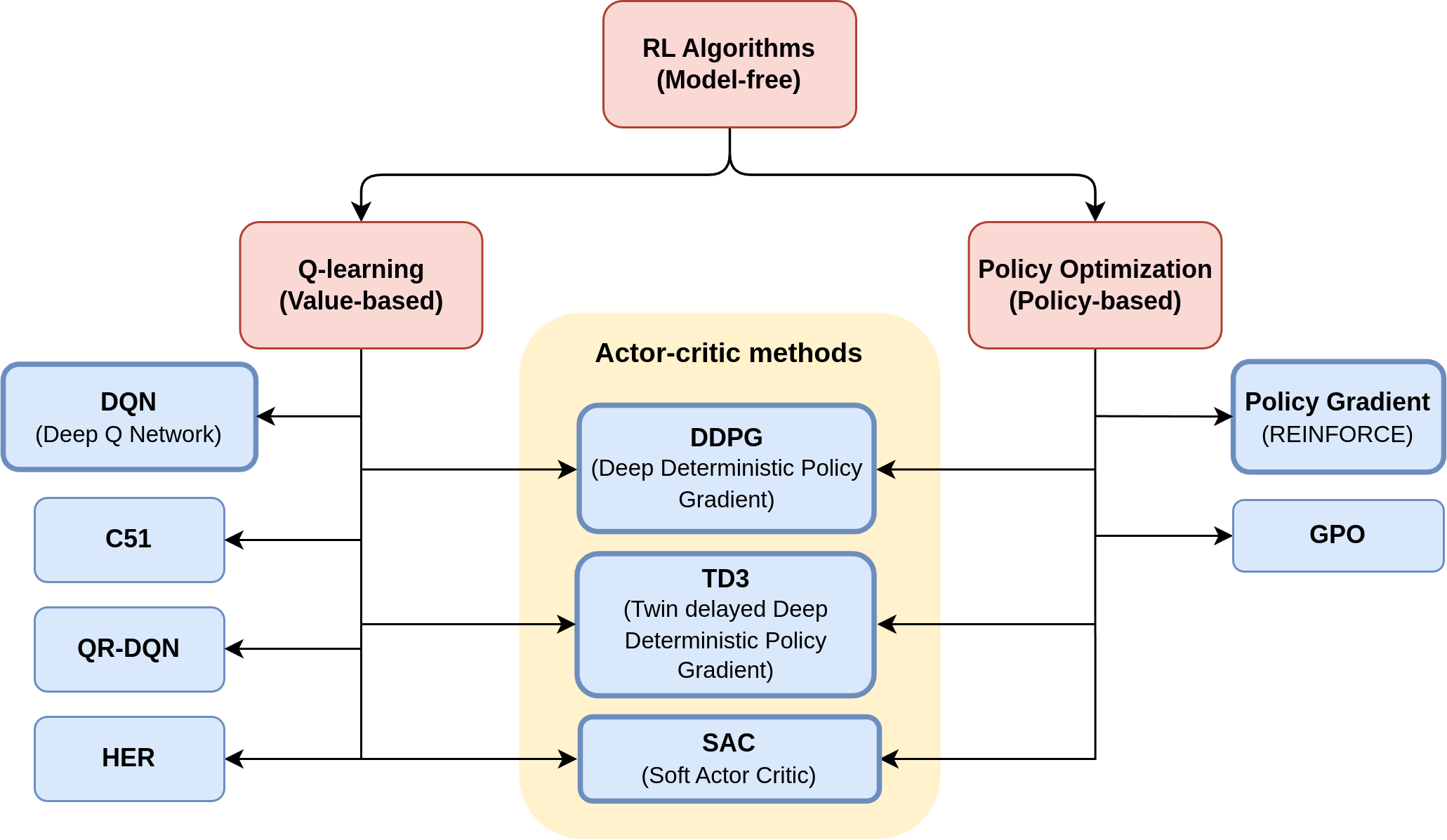}
    \caption{Model-free RL algorithms (non-exhaustive)}
    \label{fig:rl-flow}
\end{figure}
Fig. \ref{fig:rl-flow} shows a non-exhaustive list of Model-free RL algorithms. The methods highlighted in blue are discussed in the following sections.

\subsubsection{Deep Q Networks}

Deep Q-Network(DQN) replaces the Q-table with a neural network that maps states to Q-values for all actions. It is a neural network that approximates the Q value. If the DQN is parametrized by $\theta$, the network is trained to find the optimal parameters $\theta^{*}$, which will provide the optimal Q function ($Q^{*}(s,a)$), which satisfies the Bellman optimality condition (Eqn \ref{eqn:bellman-optimality-Q}). Optimal policy can be found hereafter.

DQN is trained to approximate the Q values for all possible actions for the given input state. The training data for DQN is obtained from replay buffer, which stores the agent's ($s,a,r,s^{'}$) transitions over several episodes. The loss function is mean squared error between the target value (Optimal Q value) and predicted Q value (by DQN).

\begin{equation}
    L(\theta) = \frac{1}{K} \sum_{i=1}^{K} (y_{i} - Q_{\theta} (s_{i}, a_{i}))^{2}
    \label{eqn:dqn}
\end{equation}

Here, $y_{i}$ is the target value defined as $y_{i} = r_{i} +\underset{a^{'}}{\max}Q_{\theta^{'}}(s_{i}, a_{i})$. The target value is estimated using another neural network, which is a time delayed copy of the main network, known as target network (represented by $\theta^{'}$) in order to have target stationarity. The parameters $\theta^{'}$ are copied from the main network parameters ($\theta$) every C steps (where C is a fixed update frequency). This introduces a delay in the target calculation to stabilize the learning process.

\subsubsection{Policy Gradients \& Actor-Critic methods}

Policy-based methods allow us to directly learn or optimize the policy, without requiring the computation of an explicit value function such as the optimal Q function. Unlike value-based methods, which typically struggle with continuous action spaces due to the difficulty of maximizing over an infinite set of actions, policy-based approaches can handle both discrete and continuous action spaces.

In policy gradient methods, we typically learn a stochastic policy $\pi_{\theta}(a|s)$, where the policy network (parameterized by $\theta$) outputs a probability distribution over possible actions for a given state. The policy is trained using data sampled from episodes generated by following the current policy $\pi_{\theta}$. Starting from an untrained policy, the agent gradually learns to select actions that yield higher rewards for a given state.

The core idea behind policy gradient methods is to adjust the policy parameters so as to increase the probability of actions that lead to higher returns, and decrease the probability of actions that lead to lower returns. The objective function is defined as the expected return under the policy:

\begin{equation}
    J(\theta) = E_{\tau \sim \pi_{\theta}}[G(\tau)]
\end{equation}

where $G(\tau)$ is the total discounted return along trajectory $\tau$, with $\tau$ sampled according to the policy $\pi$ (i.e., $\tau \sim \pi_{\theta}$). This objective is maximized via gradient ascent (as $\theta = \theta + \alpha\nabla_{\theta} J(\theta)$) where the policy gradient is given by:

\begin{equation}
    \nabla_{\theta} J(\theta) = E_{\tau \sim \pi_{\theta}}\left[ \sum_{t=0}^{T-1} \nabla_{\theta} \text{ log } \pi_{\theta} (a_{t}|s_{t}) G(\tau) \right]
\end{equation}

Here, $\text{ log }\pi_{\theta}(a_{t}|s_{t})$ is the log probability of taking action $a_{t}$ in state $s_{t}$ under the current policy. Refer to \cite{sutton1999policy} for the details of policy gradient. In practice, this expectation is estimated using average over N trajectories (sampled \emph{on-policy} using $\pi_{\theta}$), also known as Monte Carlo approximation:

\begin{equation}
    \nabla_{\theta} J(\theta) = \frac{1}{N} \sum_{i=1}^{N} \left[ \sum_{t=0}^{T-1} \nabla_{\theta} \text{ log } \pi_{\theta} (a_{t}|s_{t}) G(\tau) \right]
\end{equation}

This algorithm is commonly known as REINFORCE. While REINFORCE and similar policy gradient methods provide a foundation for policy optimization, they often suffer from high variance in gradient estimates. A simple yet effective improvement is the REINFORCE with baseline method, where a baseline value is subtracted from the return to reduce variance. Typically, the baseline is chosen as the value function $V_{\phi}(s_{t})$, which is estimated by a separate neural network (Value network) parameterized by $\phi$. This leads to the following update:

\begin{equation}
    \nabla_{\theta} J(\theta) = \frac{1}{N} \sum_{i=1}^{N} \left[ \sum_{t=0}^{T-1} \nabla_{\theta} \text{ log } \pi_{\theta} (a_{t}|s_{t})  (G(\tau)-V_{\phi}(s_{t}) \right]
    \label{eqn:2.1.20}
\end{equation}

\subsubsection{Actor-Critic Algorithms}
\label{subsubsec-ch2-Actor-critic}

\begin{figure}[H]
    \centering
    \includegraphics[width=8cm]{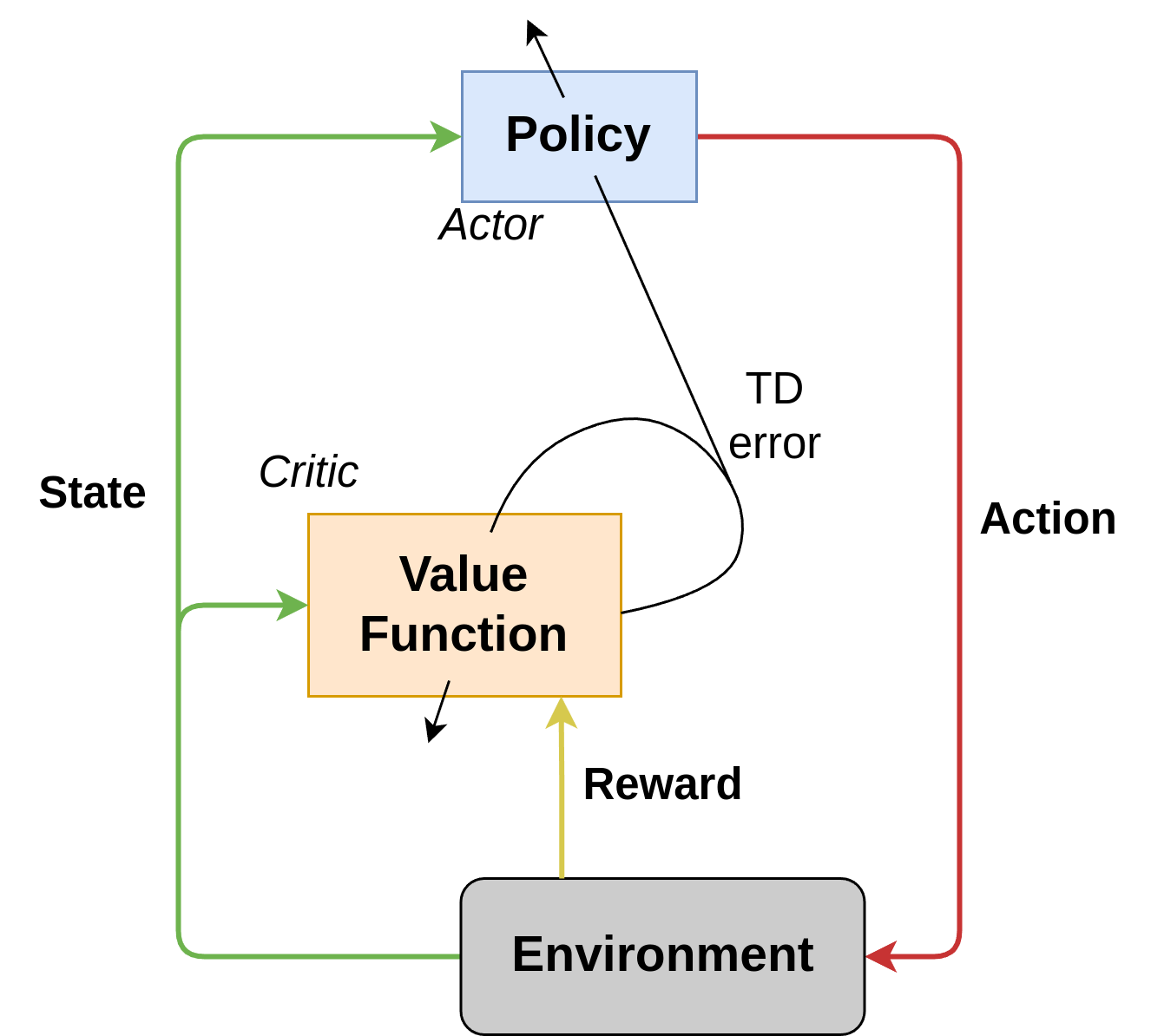}
    \caption{Illustration of actor-critic framework in RL}
    \label{fig:actor-critic}
\end{figure}

Actor-Critic (AC) frameworks combine the strengths of both policy-based and value-based methods to address the limitations of pure policy gradient algorithms such as REINFORCE.
This architecture (as shown in Fig. \ref{fig:actor-critic}) uses two main components:
\begin{itemize}
    \item Actor refers to a policy network that learns a parameterized policy ($\pi_{\theta}$) to select action in each state.
    \item Critic is a value network (similar to REINFORCE with baseline) that estimates the expected return (for example, using the value function $V_{\phi}(s)$ or action-value function $Q_{\phi}(s,a))$ to guide the actor’s learning.
\end{itemize}

The actor is updated using policy gradients with critic value as the baseline (to reduce variance), similar to REINFORCE with baseline (Eqn. \ref{eqn:2.1.20}). But a major difference is that Actor Critic methods are online (updated after each timestep in episode) and therefore return $G(\tau)$ in Eqn. \ref{eqn:2.1.20} is replaced by the bootstrapped estimate as follows:
% as follows:-

\begin{equation}
    \nabla_{\theta} J(\theta) = \frac{1}{N} \sum_{i=1}^{N} \left[ \sum_{t=0}^{T-1} \nabla_{\theta} \text{ log } \pi_{\theta} (a_{t}|s_{t})  (r+\gamma V_{\phi}(s_{t^{'}})-V_{\phi}(s_{t}) \right]
    \label{eqn:2.1.20}
\end{equation}

Critic is updated using Temporal Difference (TD) error which is computed as the difference between the bootstrapped estimate and the current value prediction:

\begin{equation}
    J(\phi) = r+\gamma V_{\phi}(s_{t^{'}})-V_{\phi}(s_{t})
\end{equation}

Actor-critic methods efficiently handle continuous action spaces and provide more stable, lower-variance learning. They enable online, stepwise updates at every timestep, improving sample efficiency and convergence speed.

This section focuses on three prominent off-policy actor-critic algorithms: Deep Deterministic Policy Gradient (DDPG), Twin Delayed Deep Deterministic Policy Gradient (TD3), and Soft Actor-Critic (SAC). Each of these builds upon the limitations of the previous, progressively improving stability and sample efficiency.

(i) \emph{DDPG} \cite{lillicrap2015continuous} extends Actor-Critic to continuous action spaces using deterministic policies and deep neural networks.
It combines elements from DQN and deterministic policy gradients. Specifically, DDPG uses:

\begin{itemize}
    \item A deterministic actor network ($\phi$) and a critic network ($\theta$).
    \item A target actor(parametrized by $\phi^{'}$) and target critic(parametrized by $\theta^{'}$), which are slowly updated to stabilize training, similar to the target networks in DQN.
    % \item An experience replay buffer, which allows the agent to learn from past experiences by sampling mini-batches uniformly.
\end{itemize}

As the name suggests, DDPG learns a deterministic policy ($\mu$), therefore exploration noise is added to the action : $a = \mu_{\phi}(s) + N$.

Critic Loss:
The critic is basically a DQN (value network) which predicts the Q value $(Q_{\theta}(s,a))$. It is trained on transitions sampled from the replay buffer, to minimize the difference between actual and target Q value (refer eqn. \ref{eqn:dqn}).
The target Q value($y_{i}$) in eqn. \ref{eqn:dqn} is obtained using the outputs from target actor ($\phi^{'}$) and target critic ($\theta^{'}$) networks as shown in Eqn. \ref{eqn:2.1.23}. The bellman target is modified to ensure stable convergence in actor-critic learning. Thus critic loss function becomes:

\begin{equation}
    L(\theta) = \frac{1}{K}\sum_{i}\left(r_{i}+\gamma Q_{\theta^{'}}(s_{i},\mu_{\phi^{'}}(s_{i^{'}})) - Q_{\theta}(s_{i},a_{i}) \right)^{2}
\end{equation}
\label{eqn:2.1.23}

Actor Loss:
The actor is a policy network that learns the optimal policy by maximizing the following main objective function:
% , to learn a policy having high Q value:

\begin{equation}
    J(\phi) = \frac{1}{K}\sum_{i}Q_{\theta}(s_{i},a)
\end{equation}

\emph{(ii) TD3} \cite{fujimoto2018addressing} builds on DDPG and was introduced to address DDPG’s \textbf{overestimation bias} and instability through three key enhancements:
\begin{enumerate}
    % \item \textbf{Clipped double Q learning}: Uses two Q-networks/ Critic networks and takes the minimum for target updates, to prevent overestimation bias.
    \item Clipped double Q learning: Uses two Q-networks(2 main critics: $\theta_{1}$ and $\theta_{2}$, and 2 target critics: $\theta_{1^{'}}$ and $\theta_{2^{'}}$) and takes the minimum(of target values $Q_{\theta_{1}^{'}}(s^{'},a^{'})$ and $Q_{\theta_{2}^{'}}(s^{'},a^{'})$) for target updates, to prevent overestimation bias.
    \item Delayed policy updates: Updates the actor less frequently than the critics to stabilize training. 
    \item Target policy smoothing: Adds noise to the target actions to prevent overfitting.
\end{enumerate}

% \textbf{Soft Actor-Critic(SAC)} \cite{haarnoja2018soft} introduces entropy regularization to promote exploration. It is a state-of-the-art off-policy actor-critic algorithm. The objective includes an entropy term 
\emph{(iii) Soft Actor-Critic (SAC)} \cite{haarnoja2018soft} is a state-of-the-art off-policy actor-critic algorithm that incorporates entropy regularization to promote exploration and improve learning stability. The actor (policy) objective includes an entropy term and the objective can be expressed as:
% , encouraging the agent to maintain a stochastic policy and avoid premature convergence to suboptimal deterministic behaviors. The objective can be expressed as:

\begin{equation}
    J_{\pi}(\phi) = \frac{1}{K}\sum_{i}Q_{\theta}(s_{i},a) - \alpha [log \pi_{\phi}(a|s_{i})]
\end{equation}

Here, $log [\pi_{\phi}(a|s_{i})]$ is the entropy term, where higher entropy corresponds to more randomness in action selection. The coefficient $\alpha$, known as the temperature parameter, controls the trade-off between maximizing expected reward and policy entropy, thereby balancing exploration and exploitation.

% Here, $log \pi_{\phi}(a|s_{i})$ is the entropy term (higher entropy means more randomness in action selection).
% and $\alpha$ is called as temperature and it is used to control exploration by setting the importance of entropy term.

SAC employs a stochastic policy, where the actor outputs a probability distribution over actions rather than deterministic actions. To reduce overestimation bias in value estimation, SAC uses two Q-networks (twin critics), similar to TD3, but differs significantly in its use of entropy regularization and stochastic policies.

% Stochastic Policy: The actor outputs a probability distribution over actions

% Similar to TD3, SAC uses two Q-networks to mitigate overestimation bias.

% In contrast to TD3 which uses only the Q function as a critic to evaluate actor's policy, SAC uses both the Q function and the value function.

% In DDPG and TD3, the target Q value in the critic is computed using the Bellman equation

% \section{Transformers}
\section{Transformers for Sequence Modeling}
In this work, Transformer-based models are employed to learn tract-specific tracking policies. The GPT-based actor is trained on RL-generated trajectories, enabling it to refine and fuse the underlying RL policies for tractography by leveraging its strong pre-training and generalization capabilities, further detailed in Chapters 3 and 4.
% GPT based actor learning in an RL paradigm.

% Highlight applications in NLP (translation, text generation) and vision tasks

Prior to transformer models, Recurrent Neural Network (RNN) \cite{rumelhart1985learning} and Long short-term memory (LSTM) \cite{hochreiter1997long} networks were widely popular for sequential modeling. However, they processed input one timestep at a time, maintaining a hidden state to capture the information from previous timesteps in the sequence. This led to a major drawback of sequential computation to preserve the order of elements, which prevented the training parallelization. Other important issues included the difficulty in learning long range dependencies, and the vanishing or exploding gradient problem.

Transformers introduced in \cite{vaswani2017attention}, removed the need for recurrent connections, by utilizing attention mechanisms to model relationships between all elements of the sequence as described below. They can model long-range dependencies, while also enabling parallelization for faster training.
% following line can be commented:-
% Transformers utilize self-attention to capture contextual dependencies across a sequence, without relying on hidden state or recurrence.

% where to add info that: Transformer has an encoder-decoder architecture.

\subsection{Self-Attention Mechanism}

Self attention computes a representation of each element or token with respect to all other tokens in the sequence.
To achieve this, first, each input token is mapped to three d-dimensional vectors, Query (\emph{q}), Key (\emph{k}), and Value (\emph{v}), through learnable linear transformations. For a given $i^{th}$ token, say $t_{i}$ with query vector $q$, attention scores are computed by comparing $q$ with key vectors $k_t$ of all tokens $t=1,2,....,T$ in the sequence. The attention score between $q$ and $k_{t}$ is computed using a scaled dot product:

\begin{equation}
    s_{t} = \frac{q^{T}k_{t}}{\sqrt{d}}
\end{equation}

The scores \(\{s_t\}_{t=1}^{T}\) are passed through a softmax function to obtain a distribution over all tokens in the sequence:

\begin{equation}
\label{sdpa}
    a_{t} = \frac{exp(s_{t})}{\sum_{l=1}^{T}exp(s_{l})}
\end{equation}

This is also known as Scaled dot product Attention (SDPA). Finally, the resulting representation of the token $t_{i}$ is given by the attention ($a_{t}$) weighted sum of its value vectors ($v_{t}$). These representations preserve the sequential information by avoiding recurrence, and enable parallelization.

% Attention score/ similarity is computed for Query vector of each token/element with respect to keys of all tokens.
% Attention score is computed as a scaled dot product, passed through softmax function to get a distribution. The resulting representation of a token/element is given by the attention weighted sum of its value vectors. 
% The attention score between a query vector $q$ and 

% For element i, its attention score with element t (where $t=1,2,...,T$) is computed as:
% For d-dimensional query(q), key($k_{t}$), and value($v_{t}$) vectors, attention score for $t^{th}$ element having key and value vectors $k_{t}$, and $v_{t}$ where is given by:

% For d-dimensional query(q), attention score for $t^{th}$ element having key($k_{t}$) and value($v_{t}$) vectors where $t=1,2,...,T$, is given by:

% \begin{equation}
%     a_{i,t} = softmax \left(\frac{q_{i}^{T}k_{t}}{\sqrt{d}}\right)
% \end{equation}

% attention weights are given by:

% \begin{equation}
    % softmax(a_{i,t}) = \frac{exp(a_{i,t})}{\sum_{t=1}^{T}exp(a_{i,t})}
% \end{equation}

% The attention scores are used to take a weighted sum of value the vectors, which results in the output representation of a given token/element.

% Query, key and Value.

\subsection{Transformer Architecture}

The transformer follows an encoder-decoder framework.
The encoder maps the input sequence into an intermediate representation, and the decoder generates an output sequence conditioned on the encoder's output representations/embedding and the previously generated outputs.
% ToDo: input embedding
% As shown in the figure, N encoders are stacked vertically, where each encoder sends its output to the one above it.

\begin{figure}[hbt!]
    \centering
    \includegraphics[width=\linewidth]{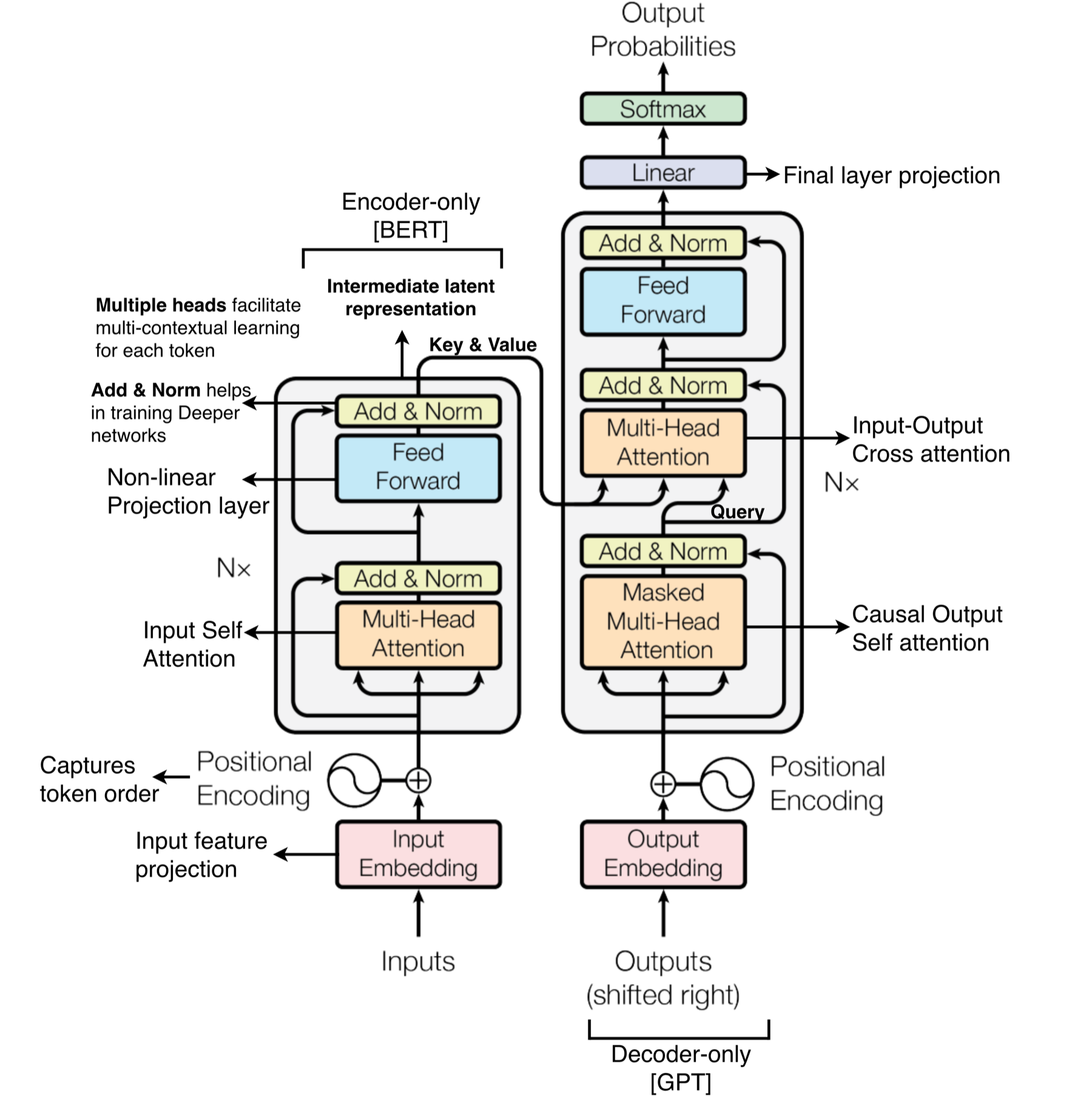}
    \caption{Illustration of Transformer model architecture \cite{vaswani2017attention}}
    \label{fig:tf}
\end{figure}

The encoder is composed of $N$ stacked layers, each consisting of a multi-head self-attention mechanism and a position-wise feedforward neural network, with residual connections followed by layer normalization (Add \& Norm) applied after each sub-layer, as shown in Figure~\ref{fig:tf}.

Before entering the encoder stack, the input tokens are first passed through a learned input embedding layer to obtain dense vector representations. Since the Transformer architecture does not inherently capture the order of tokens, additional positional information is combined with the input embeddings. The positional encodings are either fixed (using sinusoidal functions) or learned during training.
The sinusoidal positional encoding utilized in \cite{vaswani2017attention} is given as follows:

\begin{equation}
PE_{(pos, 2i)} = \sin\left(\frac{pos}{10000^{\frac{2i}{d_{\text{model}}}}}\right)
\end{equation}

\begin{equation}
PE_{(pos, 2i+1)} = \cos\left(\frac{pos}{10000^{\frac{2i}{d_{\text{model}}}}}\right)
\end{equation}

where $pos$ denotes the position index, $i$ denotes the dimension index, and $d_{\text{model}}$ is the dimensionality of the embeddings. The even dimensions use the sine function, and odd dimensions use the cosine function. This continuous and periodic encoding scheme enables the model to capture both absolute and relative positional relationships between tokens.
% This design allows the model to easily learn to attend by relative positions. 

The resulting input representations, after adding input embeddings to positional encodings, are then fed into the encoder layers. The two core components of each encoder layer are:

\subsubsection{Multi-head Self-attention} It runs multiple attention mechanisms or ``heads'' in parallel. Each head learns its own linear projections for queries, keys, and values. This allows for learning different types of relationships. The outputs of all heads are concatenated and passed through a final linear transformation to produce the output of the multi-head attention module. This enables the model to jointly attend to information from different representation subspaces at different positions. Formally, the multi-head attention is defined as:

\begin{equation}
\text{MultiHead}(Q, K, V) = \text{Concat}(\text{head}_1, \text{head}_2, \ldots, \text{head}_h)W^O
\end{equation}

where each attention head is computed as:

% \begin{equation}
% \text{head}_i = \text{Attention}(QW_i^Q, KW_i^K, VW_i^V)
% \end{equation}

\begin{equation}
\text{head}_i = \text{SDPA}(QW_i^Q, KW_i^K, VW_i^V)
\end{equation}

Here:
\begin{itemize}
    \item $Q, K, V$ are the matrices of queries, keys, and values, respectively.
    \item $W_i^Q, W_i^K, W_i^V$ are learnable projection matrices for the $i$-th head.
    \item $W^O$ is the learnable projection matrix applied after concatenating the outputs of all heads.
\end{itemize}
% The $\text{Attention}$ function typically refers to scaled dot-product attention.

The $\text{SDPA}$ function refers to scaled dot-product attention function (eq: \ref{sdpa}).

\subsubsection{Position-wise Feedforward Network} A shared feedforward neural network (FFNN) is used for each token(at each timestep) in the sequence. It captures the non-linear relationships in the data. Each FFNN consists of two dense layers with ReLU activation function, and the parameter weights are shared across all the timesteps in the sequence and different across the `$N$' encoder blocks.

The decoder has a similar structure to the encoder, consisting of $N$ stacked layers. Each decoder layer has three main sub-components: a \emph{masked multi-head self-attention} mechanism, a \emph{multi-head cross-attention} mechanism over the encoder outputs, and a position-wise feedforward network. As in the encoder, each sub-layer is followed by residual connections and layer normalization (Add \& Norm). 

% The cross attention 

The input to the decoder is the target sequence, passed through a learned embedding layer with positional information added. The masked self-attention ensures that each position in the decoder can only attend to earlier positions, maintaining the autoregressive property during generation. The cross-attention layer allows the decoder to attend to the encoder outputs, enabling it to condition its predictions on the encoded input sequence, as shown in Fig. \ref{fig:tf}.

\subsection{Generative Pre-trained Transformer}

The Generative Pre-trained Transformer (GPT) \cite{radford2018improving} is a \textbf{decoder-only} variant of the Transformer architecture, designed specifically for generative tasks. Unlike the original Transformer, which includes both encoder and decoder blocks (as shown in Fig. \ref{fig:tf}), GPT consists only the decoder stack and relies on causal (unidirectional) self-attention to autoregressively predict the next element in a sequence. This means that they generate outputs one element at a time by conditioning on previously generated elements. The model operates within a fixed \textbf{context length}, defined during training, which limits the number of previous tokens it can attend to when predicting the next token. In contrast, BERT (Bidirectional Encoder Representations from Transformers) \cite{devlin2019bert} uses only the Transformer encoder and employs bidirectional self-attention to model full context, making it well-suited for discriminative tasks like classification, question answering, and token-level labeling.

GPT was originally introduced for language modeling, where the objective is to predict the next word (token) in a sentence given the preceding context. However, its autoregressive structure and flexible sequence modeling capabilities make it broadly applicable to a variety of sequential data domains, such as \textit{time-series forecasting} \cite{zhou2021informer}, \textit{protein sequences} \cite{jumper2021highly}, and \textit{code generation} \cite{chen2021evaluating}.

Subsequent versions such as GPT-2, GPT-3, and GPT-4 have demonstrated that increasing the model size (depth, width, and number of parameters), training data scale, and computational resources leads to substantial gains in generalization, reasoning ability, and few-shot learning. These models are pretrained on large unlabelled datasets and can be fine-tuned on smaller, task-specific datasets to improve performance on downstream applications.

While pretraining equips the model with a broad understanding of sequential patterns, fine-tuning helps specialize it for specific tasks, improving task performance, factual grounding, and alignment with human preferences.
The pretraining and fine-tuning stages of GPT training are discussed in detail below:

\subsubsection{Unsupervised Pretraining}

GPT models are pretrained in an unsupervised manner using the Causal Language Modeling (CLM) objective, commonly known as next-token prediction.
Training is conducted in an autoregressive manner on large-scale unlabelled sequential data, using the negative log-likelihood loss to predict the next token in a sequence, given all previous tokens. Thus given a sequence of tokens $x = (x_1, x_2, ..., x_T)$, the model is trained to minimize the following loss function:

\begin{equation}
\mathcal{L}_{\text{CLM}} = -\sum_{t=1}^{T} \log P(x_t \mid x_1, ..., x_{t-1})
\end{equation}

The masked self-attention mechanism ensures that, during training, each token can only attend to previous tokens in the sequence. This preserves the causal (left-to-right) structure required for generative modeling. This unsupervised pretraining approach allows the model to learn transferable representations and temporal dependencies, which can later be adapted to various downstream tasks through fine-tuning.

\subsubsection{Supervised Fine-tuning}

Fine-tuning is performed after pretraining to adapt the model to specific downstream tasks. This typically involves appending task-specific heads (e.g., linear layers) on top of the pretrained transformer and training the model on labeled data typically using supervised loss functions. Depending on the task and the size of the available dataset, either the entire model or only a subset of layers may be updated.

This study focuses on supervised fine-tuning (of GPT) for tractography, as detailed in Chapters 3 and 4.
\chapter{Tract-specific Tractography using GPT-based RL Policy Refinement}\label{trlf}

\section{Introduction}

Tractography is a complex sequential decision-making process. Recent research has increasingly leveraged machine learning and deep learning (DL) approaches to improve tractography accuracy. In particular, Deep Reinforcement Learning (DRL) has emerged as a promising direction, as it does not rely on unreliable or hard-to-obtain ground truth data for training. However, there remains significant room for performance improvement.

\begin{tcolorbox}[
    colback=gray!5,                 % Light background
    colframe=black!60,              % Border color
    coltitle=white,                 % Title text color
    % colbacktitle=black!40,
    colbacktitle=gray!50!black,% Title background: softer grey-black
    enhanced,
    boxrule=0pt,
    sharp corners=south,            % Bottom corners sharp
    rounded corners=north,          % Top corners rounded
    drop shadow southwest,          % Shadow
    title={\bfseries Problem Statement},
    fonttitle=\large\bfseries,
]
To develop a tract-specific framework for white matter fiber tractography. 
The input is diffusion MRI data, and the goal is to reconstruct white matter 
fiber trajectories belonging to a targeted anatomical tract. Each fiber is 
represented as an ordered sequence of 3D points $(x, y, z)$ that follows biologically plausible streamline propagation constraints.

\end{tcolorbox}

Transformers have recently demonstrated outstanding performance across diverse domains, including language modeling \cite{vaswani2017attention}, image recognition \cite{dosovitskiy2020image}, time-series forecasting \cite{zhou2021informer}, and protein structure prediction \cite{jumper2021highly}. Their strong ability to model long-range dependencies and contextual information makes them well-suited for sequence prediction tasks, such as mapping neural pathways in tractography. Building on these successes, Tract-RLFormer integrates the generalization, transfer learning, and autoregressive capabilities of transformers (e.g., GPT models) to enhance the performance of Deep RL policies for tractography.

\section{Data Specifications and Pre-processing}
\label{section:datasets}

Tractography involves using Diffusion MRI data to generate white matter fiber tracts. Tractography datasets include diffusion MRI data of one or more subjects and their corresponding white matter tracts.

In this study, we utilize 3 datasets, namely TractoInferno \cite{poulin2022tractoinferno}, Human Connectome Project (HCP) \cite{van2012human}, and the ISMRM 2015 Tractography Challenge dataset \cite{maier2017challenge} as described in Table \ref{tab:datasets}.
ISMRM is a synthetic phantom dataset consisting of reference tracts for 25 distinct white matter bundles. HCP provides in-vivo diffusion MRI data from healthy adult subjects. TractoInferno is a synthetic benchmark dataset constructed using four tractography algorithms that provide reference tracts for 30 bundles for machine learning and evaluation tasks. For the HCP dataset, we use the reference white matter tracts released as part of \cite{wasserthal2018tractseg}, which provides reference tract segmentations of 72 bundles for 105 subjects of the HCP young adult dataset.

Diffusion MRI undergoes a series of processing steps before it can be effectively used in tractography algorithms. This process typically consists of two main stages, as previously detailed in Sections \ref{subsec:preprocessing-dmri} and \ref{modeling-dmri}:
\begin{enumerate}
    \item \textbf{Preprocessing}, which includes denoising, motion and eddy current correction, and bias field correction.
    \item \textbf{Signal modeling and feature extraction}, where fiber orientation distributions (fODFs) are estimated and local diffusion features like fODF peaks are extracted as inputs to the tractography pipeline.
\end{enumerate}

Table \ref{tab:datasets} summarizes the preprocessing steps that have already been applied to each dataset. The fODF and peaks computation was performed using \textit{scilpy} \cite{scilpy_docs}.
Following preprocessing, fODFs are estimated using constrained spherical deconvolution (CSD) \cite{tournier2007robust} (see Section \ref{subsec:fodf}). Each fODF is represented using 45 coefficients of 8th-order spherical harmonics (SH). From the resulting fODFs, five peaks are extracted per voxel, representing the dominant fiber directions. Table \ref{tab:datasets} also presents the acquisition details of the datasets, along with the number of subjects used for training and testing in this work.

\begin{table}[h]
\centering
\begin{threeparttable}
\caption{Details of the diffusion datasets used in our experiments.}
\label{tab:datasets}
\begin{tabular}{ p{0.19\textwidth}p{0.12\textwidth}p{0.25\textwidth}p{0.32\textwidth} }
 \hline
 \textbf{Dataset} & \textbf{Subjects} & \textbf{DWI data} & \textbf{Distortion Corrections} \\
 \hline
 TractoInferno \cite{poulin2022tractoinferno} & 284\tnote{a} & b=1000 $s/mm^{2}$; resolution=1mm isometric & N4 bias field; eddy-current; head-motion \\

 \hline
 HCP \cite{van2012human} & 1200\tnote{b} & b=1000/2000/3000 $s/mm^{2}$; 270 directions; resolution=1.25mm isometric & EPI; eddy-current; subject-motion \\

 \hline
 ISMRM \cite{maier2017challenge} & 1\tnote{c} & b=1000 $s/mm^{2}$; 32 directions; resolution=2mm isometric & Eddy currents; head motion \textit{(by our preprocessing)} \\
 \hline
\end{tabular}

\vspace{0.2cm}
\begin{tablenotes}
\footnotesize
\item[a] Train IDs: 1030, 1079, 1119, 1180, 1198; Test IDs: 1160, 1078, 1061, 1159, 1171
\item[b] Test IDs: 930449, 959574, 992774, 987983
\item[c] Only one subject available; no ID needed
\end{tablenotes}
\end{threeparttable}
\vspace{-0.5cm}
\end{table}

\section{Proposed Method}

\begin{figure}[hbt!]
    \centering

    \includegraphics[angle=90,height=0.81\textheight, width=13.5cm]{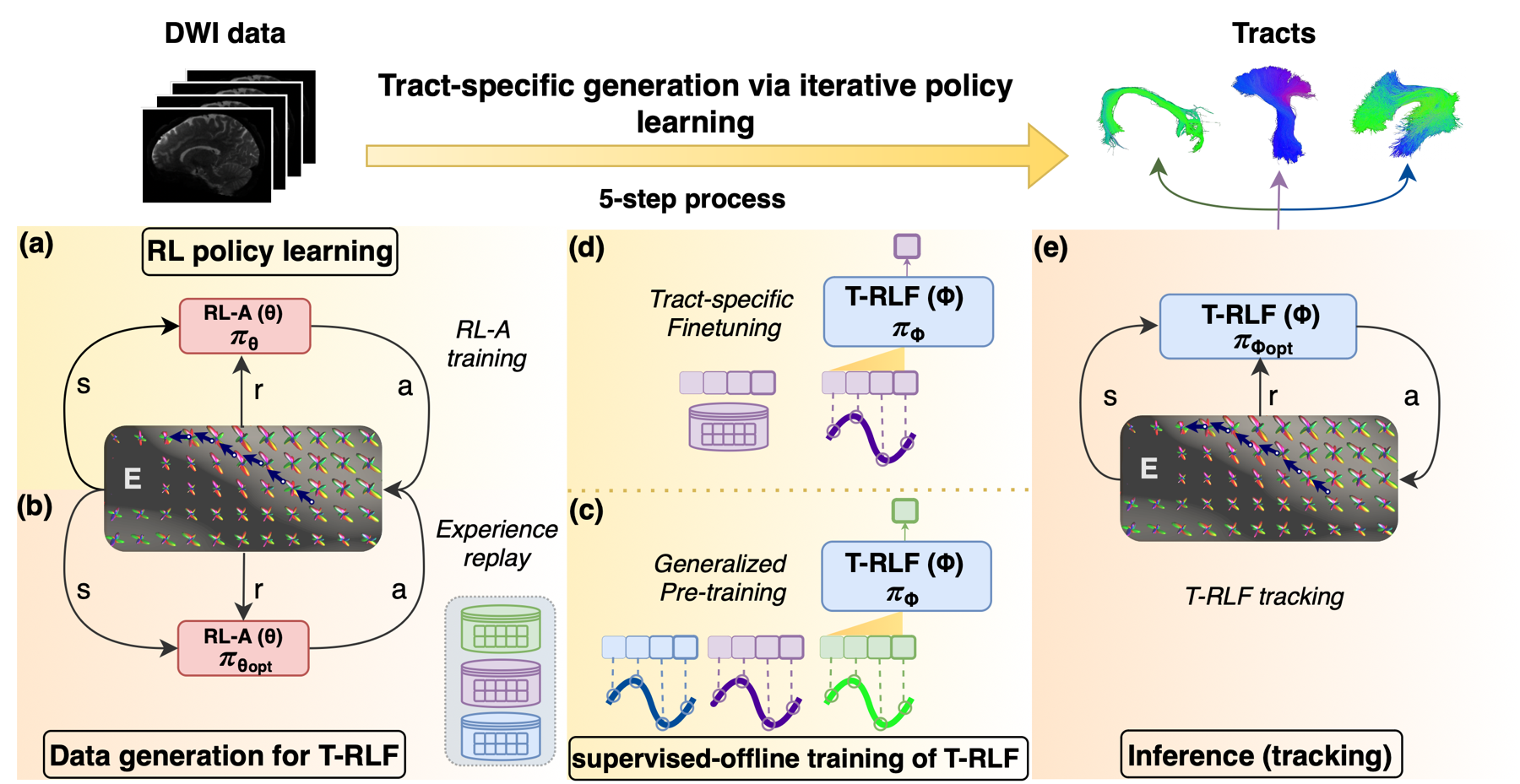}
    \caption[Overview of the proposed Tract-RLFormer framework.]{Overview of the proposed Tract-RLFormer framework. (a) An RL agent ($\pi_{\theta}$) interacts with the environment (E) to learn an optimal \textbf{level-1 policy ($\pi_{\theta opt}$)}. (b) This policy is used to generate tract-specific roll-outs, denoted as 'experience replay'. (c) and (d) illustrate the offline, auto-regressive training of the proposed Tract-RLFormer $\phi$, referred to as T-RLF, over these roll-outs. In (c), T-RLF undergoes general pre-training, while in (d) it is fine-tuned to learn an optimal tract-specific policy ($\pi_{\phi opt}$). (e) shows the testing phase, where T-RLF, which has learned the new \textbf{level-2 policy ($\pi_{\phi_{opt}}$)}, performs tracking in environment $E$ to produce the desired tract. Training and tracking steps are shown in yellow and orange backgrounds, respectively.}
    \label{fig:pipeline-trlf}
\end{figure}

Tract-RLFormer is an \textit{iterative policy learning framework for tract-specific generation}, delineated as a five-step process (see Fig. \ref{fig:pipeline-trlf}).
In this framework, we start by training an RL agent (TD3 \cite{fujimoto2018addressing}) to learn a policy by exploration (within the tracking mask) to generate a tract of interest. We call it as level-1 policy.
Using this initial policy, the agent interacts with the (tracking) environment by taking actions (tracking steps). The agent’s experience (policy rollouts) is collected and sampled to train a refined version of the policy, by the T-RLF model, which learns in a data-driven manner through general pre-training and tract-specific fine-tuning. 
This study focuses on seven principal white matter (WM) tracts: Corpus Callosum (CC), left and right Pyramidal (PYT), Arcuate Fasciculus (AF), and Cingulum (CG) Tracts, shown in Fig. \ref{fig:wm-tracts}. The selection of these seven tracts is based on their clinical significance and frequent analysis as suggested in \cite{rheault2019bundle,theberge2024matters}.
To conduct such tract-specific training and generation, we first compute a tracking region of interest (mask) tailored for each tract using our Mask Refinement Module (MRM), described in \ref{subsec:ch3-mrm}. %below.
Following this, we proceed with the five sequential steps depicted in Fig. \ref{fig:pipeline-trlf}, detailed in subsequent subsections.

\subsection{Tract-RLFormer Framework}
\label{subsec:TRLF}

\subsubsection{RL Environment Setup}
\label{subsubsec:ch3-rl-env}

The reinforcement learning (RL) environment for tractography is designed to: (1) initialize seed points within the white matter mask to initiate tracking; (2) define the state at each voxel within the tracking region; (3) assign rewards and determine the next state based on the agent’s current state and actions; and (4) constrain the tracking process both spatially, within the brain mask and anatomically, according to plausible fiber length and bend.
% white matter pathways.
The RL environment for tractography was first proposed in \cite{theberge2021track}, and since then it has been employed in subsequent works that employ Deep RL for tractography \cite{theberge2024matters,theberge2024tractoracle}. It is implemented using OpenAI Gym \cite{brockman2016openai}. The different elements of the environment are:

\begin{itemize}
    \item \textbf{State:} The state at a given voxel includes the 45 Spherical Harmonic Coefficients ($8^{th}$ order) of that voxel and its six immediate neighbours, along with their white matter mask values. This is followed by appending last four tracking directions(actions) resulting in the final state vector of length 45 × 7 + 7 + 4 × 3 = 334.
    % 45*7+7+4*3= 334.
    % is a 
    \item \textbf{Action:} The action is a 3D vector $a_{t}$ that determines the next tracking direction.
    \item \textbf{Reward:} It is the product between predicted action's alignment (dot product) with the most aligned fODF peak and it's alignment(dot product) with the last tracked streamline/fiber segment.
    It is computed as:
    \begin{equation}
    r_t = \left| \max_{\vec{p}_i} \left( \vec{p}_i \cdot \vec{a}_t \right)\right| \times \left( \vec{a}_t \cdot \vec{u}_{t-1} \right)
    \label{eqn:reward}
    \end{equation}
    where, $\vec{p_{i}}$ is the fODF peak, $\vec{a}_{t}$ is the action taken at timestep $t$, and $\vec{u}_{t-1}$ is the previously tracked step at timestep $t-1$.
    \item \textbf{Episode termination criteria:} An episode terminates when the tracking exceeds the tracking mask, the fiber streamline length exceeds 200 mm, or the fiber curvature exceeds the maximum angle of $30^{\degree}$ or $60^{\degree}$ (depending on RL agent's training algorithm).
\end{itemize}

% \subsubsection{Tracking Mask Generation}
% OR 
{\subsubsection{Mask Refinement Module}
\label{subsec:ch3-mrm}
% ToDo: Give primer/intro in 2-3 lines on what it does on top level before telling how.

In this study, we generate masks for 7 tracts: the Corpus Callosum (CC) and the left and right components of three bilateral tracts, namely the Pyramidal Tract (PYT), Arcuate Fasciculus (AF), and Cingulum (CG).

As discussed in Chapter 1, a tracking mask is required to constrain tractography to white-matter regions, and in our case, to a specific tract of interest. Instead of simply relying on an atlas-based mask, which can be inaccurate across subjects, we trained a neural network to generate subject-wise tract-specific masks that account for inter-subject anatomical variability for reliable tractography.

For a given tract, its reference streamlines are obtained from RecobundlesX Atlas \cite{francois_rheault_2023_7950602}. The atlas reference tract is aligned using ANTs \cite{avants2009advanced} to the given subject space, and then converted into a binary mask, having value of 1 for voxels from which atleast one reference streamline passes, and 0 for the rest. This mask is dilated by 5mm and input to MRM, which prunes the dilated mask to estimate the tract-specific mask.
For each voxel in the dilated mask, the 45 SH coefficients and dilated mask value of the voxel and its 6 immediate neighbour voxels (resulting in a vector of size 45 × 7 + 7 = 322) are input to the MRM network, which outputs the probability of retaining that voxel in the predicted mask. Voxels with an output probability greater than 0.5 are kept in the predicted tract-specific mask, while those with lower probabilities are eliminated. Fig. \ref{fig:mrm} depicts the entire process of mask generation using MRM.

% The 

\begin{figure}[hbt!]
    \centering
    \includegraphics[width=15cm]{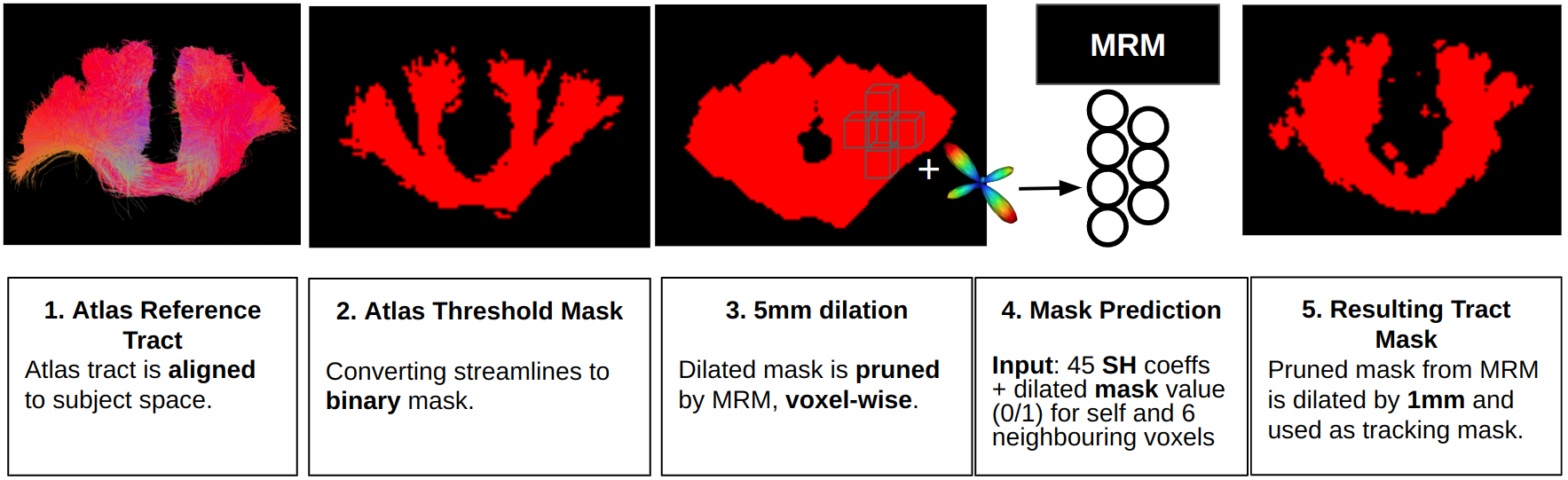}

    \caption[Mask Refinement Module for generating tract-specific masks]{Mask Refinement Module for generating tract-specific masks. (1) The atlas reference tract is aligned to subject space. (2) Streamlines are converted into a binary threshold mask (3) The mask is dilated by 5\,mm and pruned voxel-wise by the MRM. (4) Mask prediction uses as input the 45 SH coefficients together with the dilated mask value (0/1) for the voxel itself and its six neighbouring voxels. (5) The pruned mask is dilated by 1\,mm to produce the final tract-specific mask.}
    \label{fig:mrm}
\end{figure}

\textbf{Architecture and Training details:} The MRM is a Fully connected neural network comprising three hidden layers with 512,
256, and 128 neurons, respectively, and an input dimension of 322. Each layer employs a ReLU activation function and is followed by batch normalization and a dropout layer (0.5). The output layer has a sigmoid activation function, which outputs the probability of retaining voxel in the predicted mask. Training is performed in a voxel-wise manner, using Binary Cross Entropy as the loss function to compare the predicted mask value with the ground truth for each voxel. The model was trained on 50 subjects randomly selected from the TractoInferno dataset over 100 epochs. The resulting mask is then dilated by 1mm to produce the final refined tracking mask for the given subject. In Fig. \ref{fig:mrm}, the apparent disconnected components in the various mask images are due to 2D slice visualization of the volumetric mask. The mask is connected in 3D space across adjacent slices. Visualization in different orientations confirms that the components are connected and anatomically consistent.

\subsection{RL Policy Learning}
\label{subsec:prp}

We utilize TD3 (Twin Delayed Deep Deterministic Policy Gradient) algorithm to train 7 tract-specific agents. It belongs to a class of Actor-Critic algorithms in RL (refer Section \ref{subsubsec-ch2-Actor-critic}) and consists of an Actor and two Critic networks (along- with their time delayed target networks). We adopt the networks similar to \cite{theberge2021track}, where actor and critic networks are both fully-connected neural networks with two ReLU activated hidden layers of 1024 neurons each. The actor has a 334 dimensional input layer and 3-neuron tanh activated output layer to predict the 3-D action, while the critic has a 337 dimensional input layer (corresponding to both 334-D state and 3-D action as input) and a single neuron tanh output layer predicting the Q-value for a given state-action pair.
However, unlike \cite{theberge2021track}, where RL agents are trained for whole-brain tractography, we perform tract-specific training to obtain one policy for each tract of interest.
We train TD3 agent on tract-specific masks obtained from MRM. Each tract-specific RL agent is trained on five randomly selected \textbf{training subjects} from the TractoInferno dataset (1030, 1079, 1119, 1180, and 1198), for 50 batches (of 4096 episodes each) per subject, resulting in a total of 1,024,000 (250*4096) training episodes.

We train the TD3 agent in 5 different instances of the environment specified by each subject’s distinct diffusion data, fODF peaks, and tracking mask. Training is conducted at 7 seeds per voxel and a step size of 0.375mm, with fiber lengths between 20mm and 200mm. The maximum possible episode length is set to 530, corresponding to a 200 mm fiber and a 0.375 mm step size.
Other hyper-parameters include: learning rate: 8.56e-06, Discount factor ($\gamma$): 0.776, and Exploration noise ($\sigma_{train}$): 0.334.

\subsection{Tract-RLFormer for Policy Refinement}

RL agents learn a TD3 policy ($\pi_{\theta opt}$), which can be used to conduct tract-specific tractography. Using rollouts from this level-1 policy, Tract-RLFormer (T-RLF) learns a refined policy ($\pi_{\phi opt}$), by training on trajectories derived from the policy roll-outs of seven trained tract-specific TD3 agents. Each trajectory consists of sequence of state, action, and return-to-go of an episode. It is represented as \textbf{$\tau = (R_0,s_0,a_0,R_1,...,R_T ,s_T ,a_T )$}, as shown in Fig. \ref{fig:trajectory}, where $R_t$ is the scalar sum of rewards from timestep t to the episode’s end, $s_t$ is the 334-dimensional state vector, and $a_{t}$ is the 3-dimensional action as discussed in Section \ref{subsubsec:ch3-rl-env}.

T-RLF is a GPT based model that is trained on these RL-derived trajectories in a two-stage process. (a) Initially, the Tract-RLFormer undergoes a generic, tract-agnostic pre-training, (b) This is followed by fine-tuning for the downstream task of tract-specific generation.
Together, these constitute the next three steps (out of five), namely training data generation (Fig. \ref{fig:pipeline-trlf}(b)) and the two-stage training process (Fig. \ref{fig:pipeline-trlf}(c, d)) of T-RLF.
Each step of the refinement process is discussed in detail below:

\subsubsection{Training Data Generation for \textit{T-RLF}}
\label{subsubsec:ch3-data-gen}
% Training Trajectories generation

For each tract, their trained RL (TD3) agent is used to initiate tracking within its designated masks (obtained from MRM) on the 5 randomly selected training subjects from TractoInferno dataset (Subject IDs: 1030, 1079, 1119, 1180, and 1198). As the policy traverses different voxels within the tracking mask, the corresponding trajectories for those transitions are saved as <$R, s, a$> tuple. Trajectories shorter than 47 transitions, corresponding to a fiber length of approximately 20\,mm given the 1\,mm isotropic voxel resolution of the TractoInferno dataset, are filtered out.
From the remaining trajectories, 10,000 are selected per subject—comprising 5,000 of the longest trajectories and 5,000 chosen at random—resulting in a total of 50,000 trajectories per tract. This forms the tract-specific trajectories dataset, which is used for model finetuning. From the full set of 350,000 trajectories (7 tracts × 50,000 each), a subset of 150,000 is selected to create a mixed, tract-agnostic dataset, denoted as $\tau_{mix}$. This subset includes 75,000 of the longest trajectories and 75,000 randomly selected ones, and is used for generic pre-training.

\begin{figure}[hbt!]
    \centering
    \includegraphics[width=15cm]{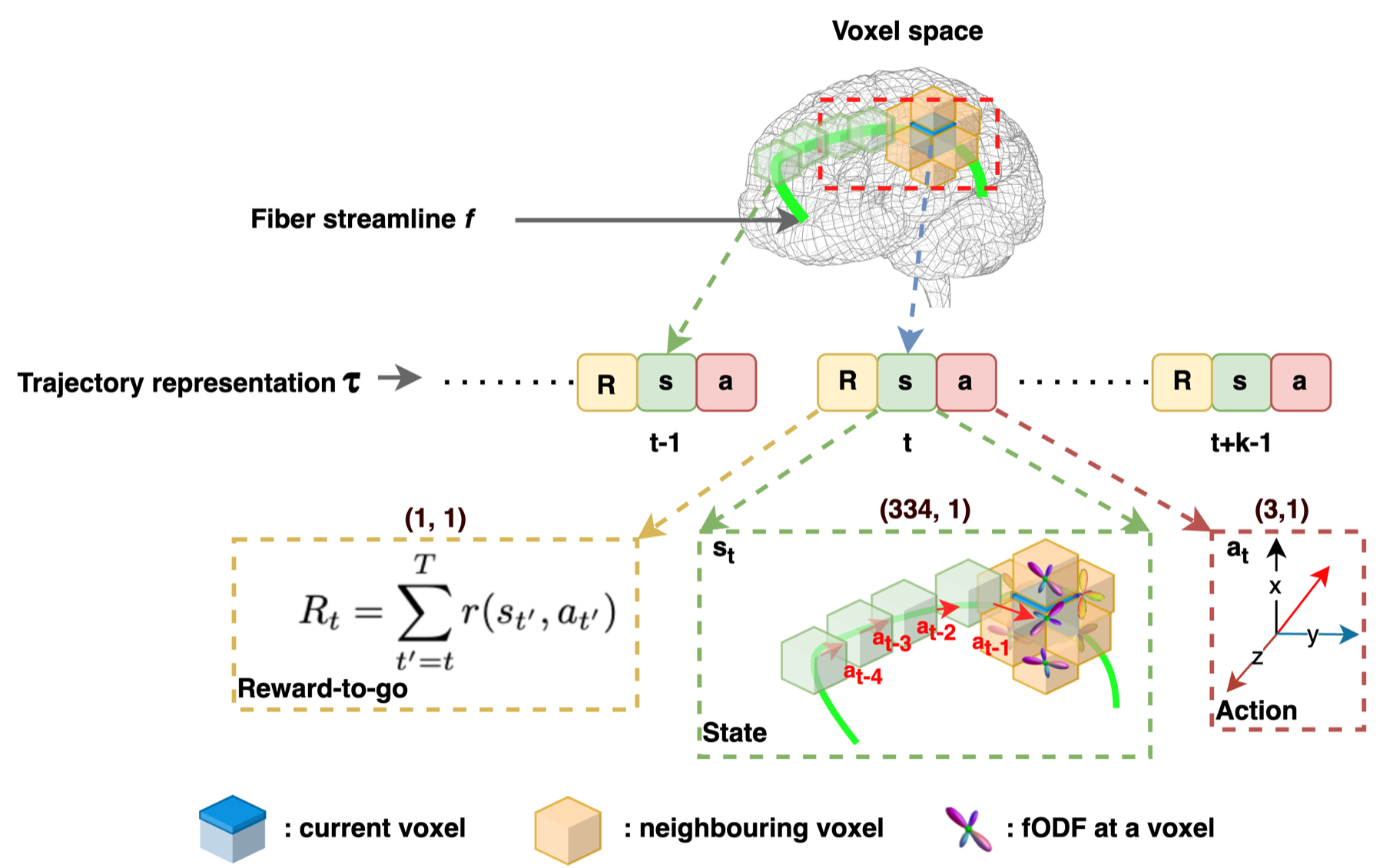}
    \caption[Data Representation for Tract-RLFormer]{Data Representation for T-RLF: Tract specific policy refinement using a trajectory-based approach in an RL agent's experience space. The figure illustrates a $k$ length fiber streamline $f$ in human brain voxel space, represented as a trajectory $\tau = (R_0, s_0, a_0, R_1, s_1, a_1, ..., R_k, s_k, a_k)$. Each point in the streamline corresponds to a state, action, and return-to-go tuple at time-step $t$.}
    \label{fig:trajectory}
\end{figure}

\subsubsection{Architecture of \textit{T-RLF}}
The network architecture of the model is based on a decoder-only Transformer framework (GPT), consisting of 4 sequential decoder layers. It supports a maximum context length of 40 tokens, enabling it to process and generate sequences up to this length at a time. Each timestep processes 3 tokens one each for state, action and return-to-go. 
The model includes a dedicated learned embedding layer of 128 dimensions (as shown in Fig. \ref{fig:trlf-training}) for each component of the trajectory: state (s), action (a), and return-to-go (R). This transforms each input timestep’s state, action and Return-to-go into vectors of dimension 128, which is combined with learned positional embeddings (of dimension 128) to retain order information. The maximum possible episode length $(max\_ep\_len)$ controls length of episode. It is set to 530 because the maximum length of a fiber is 200mm, equivalent to 530 steps for a TractoInferno subject (as 1 step corresponds to 0.375 mm). If an episode exceeds 530 timesteps, it is truncated to this length. 
Each decoder layer contains a masked multi-head self-attention mechanism with 1 attention head, followed by a position-wise feedforward neural network. Layer normalization and residual connections are employed around each sub-layer, like GPT. 
After processing through the decoder layers, the output is passed through an output embedding layer, from which we obtain the predicted action of dimension (3, 1). Dropout with a rate of 0.1 is applied to mitigate any overfitting during training.

\subsubsection{Training Procedure of \textit{T-RLF}}
\label{trlf-training}

Tract-RLFormer model is trained on <state, action, return-to-go> trajectories sampled from TD3 agent (detailed in Section \ref{subsubsec:ch3-data-gen}). 
It follows a two-stage training process: first, a \textit{general pre-training} phase on a mixed tract dataset ($\tau_{\text{mix}}$), followed by \emph{tract-specific fine-tuning} on individual tract datasets ($\tau_i$).

The first three decoder layers are pre-trained over $0.15$ million mixed trajectories (taken from $\tau_{mix}$), containing a total of 30 million transitions for 30 iterations. Later the $4^{th}$ decoder layer is fine-tuned on the tract-specific trajectories buffer for 10 additional iterations. In each iteration, the model undergoes 10,000 training steps, each processing a $batch\_size$= 128 number of K-length trajectories. 

\begin{figure}[H]
    \centering
    \includegraphics[width=\linewidth]{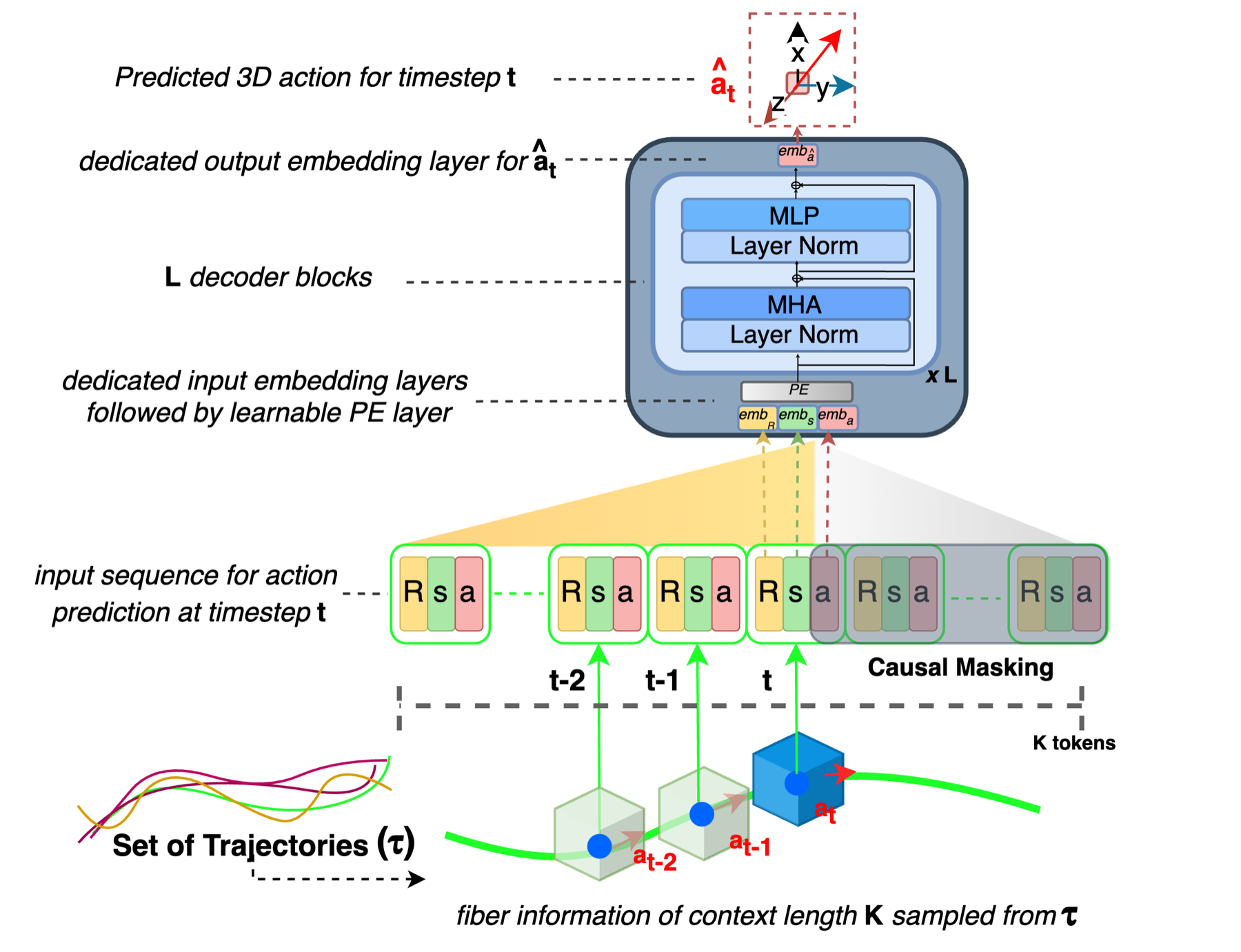}
    \caption[Training Procedure of Tract-RLFOrmer]{T-RLF training: Visual representation of training Tract-RLFormer for action prediction at time-step $t$, using context information from $K$ length fiber. The input sequence tuples <$R$, $s$, $a$> are causally masked from $a_t$ onwards and processed through embedding layers $emb_R$, $emb_s$, and $emb_a$, with a learnable positional encoding layer ($PE$). Embeddings are processed by $L$ decoder blocks ($L=3$ for pre-training, $L=4$ for fine-tuning), incorporating Multi-Head Attention (MHA) and Multi-Layer Perceptron (MLP), to generate predicted action $\hat{a_t}$.   
    }\label{fig:trlf-training}
\end{figure}

A batch of $128$ tokens of $<R_{t}, s, a>$ are sampled from training data ($\tau$) and stacked for a context length ($K = 40$) and fed as an input to the T-RLF. It passes through an embedding layer with 128 dimensions, and positional encoding is added, resulting in a (128 x 120 x 128) matrix and is processed by the 4 decoder layers with causal masking (Fig. \ref{fig:trlf-training}). The decoder output is mapped through an output embedding layer to predict the action. 
Unlike the TD3 agent, T-RLF does not interact with the environment during its training process. Instead, it is trained entirely in an offline mode using only trajectory datasets ($\tau$'s).
For the context length K, a 5-step loss (accounting for current and 2 steps in both forward and backward directions) is computed, aggregating the angular difference between predicted and actual action at each time-step.

\begin{equation}
    L = \sum_{t=2}^{K-2} \left( \sum_{i=-2}^{2} \cos^{-1} \left( \textbf{a}_{t+i} \cdot \hat{\textbf{a}}_{t+i} \right) \right)
\end{equation}

The learning of weights for $\pi_{\phi opt}$ is facilitated by this 5-step loss function, in order to generate more effective and robust actions. Here, $\vec{a}_{t+i}$ and $\vec{\hat{a}}_{t+i}$ are the true and predicted actions at $(t+i)^{th}$ timestep respectively.

Similar to \cite{chen2021decision}, T-RLF training is conditioned to generate action ($a_{t}$) using return ($R_{t}$) at each timestep.
During inference, $R_{t}$ is initialized to an expert return value or the longest trajectory return. In our case, the longest trajectory length is 292, and since the maximum possible reward at each timestep is 1, we initialize $R_{t}$ to 300 ($\thicksim$1x of expert return). This was experimentally verified among various values of $R_{t}$: 100, 200, 300, 500, and 600. For model training, we employed the AdamW optimizer, set with a learning rate of 1e-4 and a weight decay of 1e-4.

Importantly, all tract-specific T-RLF models share the same pre-trained backbone and differ only in their fine-tuning phase. This fine-tuning stage specializes the model for the downstream task of generating accurate and anatomically plausible streamlines for a specific white matter tract. The effectiveness of this two-stage training approach is evaluated in an ablation study, highlighting the contribution of both pre-training and fine-tuning to the model's performance and generalization capabilities across different tracts in section: \ref{subsec:ch3-abl}).

\subsection{Tract generation and cleaning}
\label{subsec:Tract_cleaning}

The final step in our tractography pipeline involves using the trained T-RLF models to perform tracking, followed by cleaning the resulting tracts.
Having learnt the refined policy ($\pi_{\phi opt}$), T-RLF can function autonomously as a tractography agent. 
While the T-RLF model was trained without environmental interaction, once trained (on RL trajectories), it serves as a refined tract-specific policy which can independently conduct tractography by interacting with the tracking environment (refer Section 3.2), similar to original RL policies.

\begin{itemize}
    \item \textbf{Tract generation:} Tracking is executed within tract-specific masks obtained from MRM and is initialised with 7 seeds per voxel, and $R_t$ is set to $R_0=300$. Tracking step size is (empirically  selected) and is dataset-specific, \textbf{$0.375mm$} for the TractoInferno, \textbf{$0.468mm$} for HCP, and \textbf{$0.75mm$} for the ISMRM dataset. At each step, the return-to-go ($R_t$) is reduced by the achieved reward and predicted action($a_t$), new state ($s_t'$), and $R_t$ are appended to the context window to serve as input for the next prediction. This auto-regressive process is used by Tract-RLFormer to generate the fiber tract of interest.
    \item \textbf{Tractogram cleaning:} This is a post-processing step used to filter the resulting tracts. First, length-based cleaning is applied, where streamlines shorter than 20\,mm are removed. Subsequent fine-grained filtering is performed using Fast Streamline Search (FSS) \cite{st2022fast} to eliminate any extraneous fibers, by comparing the predicted tract with the atlas reference tracts (representing general anatomical structure).
    The tracts generated by T-RLF models are confined to the masks generated by the MRM module (tailored to each subject). This approach ensures that tract generation remains confined to regions proximate to the actual neural fibers of the subject, thus mitigating the risk of false positives. Consequently, we can perform a high radius search using FSS, without incurring a major risk of high overreach. This high radius search ensures that accurate fibers are not discarded based on minor deviations from atlas tracts.
\end{itemize}

\section{Result and Performance Analysis}

In this section, we present the outcomes of our evaluation of tract-specific T-RLF models under various experimental setups, including comparative analysis, generalization performance, and an ablation study. We trained TD3 and T-RLF models, on eight tracts — seven principal white matter tracts (refer Section \ref{trlf-training}) and Optical Radius (OR tract) (for analysis in \ref{subsec:comp-analysis}) using five train subjects (id: 1030, 1079, 1119, 1180, and 1198) of the TractoInferno dataset and reported their performance on various test subjects across different datasets in subsequent subsections.
Additionally, we assess their effectiveness relative to supervised approaches and classical tractography methods.
% that do not incorporate learning.

\subsection{Evaluation Parameters}
To evaluate the performance of tractography algorithms, we utilize three evaluation metrics, namely Dice, Overlap, and Overreach. These metrics are computed pairwise between the ground-truth tract and predicted tract in a voxel-wise manner. This means that TP refers to the number of true positive voxels, i.e number of voxels in groundtruth that are also present in predicted tract. Similarly, FP and FN refer to the number of False positive and False negative voxels respectively.
Each metric serves a different purpose. \emph{Overlap (OvL)} computes the intersection of the ground truth and predicted tract. High value indicates better coverage of ground-truth area by predicted tract. It is given by:

\begin{equation}
    OvL = \frac{TP}{TP+FN}    
\end{equation}

\emph{Overreach (OvR)} indicates how much the generated tract exceeds the groundtruth, with lower scores suggesting better performance. It is given by:

\begin{equation}
    OvR = \frac{FP}{TP+FN}  
\end{equation}

\emph{Dice score (D)} assesses both the accurate coverage and the minimization of extraneous coverage
beyond the ground truth area, where values near 1 signify a high similarity level. It is given by:

\begin{equation}
    Dice = \frac{2.TP}{2.TP+FP+FN}  
\end{equation}

% \subsection{Quantitative Comparison}
\subsection{Comparative Analysis}
\label{subsec:comp-analysis}
This section provides a comparative analysis of our model, T-RLF, against supervised learning, traditional tractography, and state-of-the-art (SOTA) reinforcement learning (RL) algorithms, using Dice scores to evaluate performance across three major white matter bundles: $PYT, OR,$ and $CC$.

As presented in Table \ref{table:comparetive}, all methods are tested on subject 1006 from the TractoInferno dataset (similar to \cite{theberge2021track,theberge2024matters} for fair comparison). For the first and third tabular subparts of Table \ref{table:comparetive}, the models are trained on ISMRM data. The second subpart does not involve training (classical methods). These 3 subparts are assessed using whole-brain tractography and segmentation \cite{poulin2022tractoinferno,theberge2024matters}. Additionally, the last subpart details the performance of our T-RLF and the TD3 model, where T-RLF was specifically trained on trajectories derived from the TD3 agent.

% \begin{table}[hbt]
\begin{table}[H]
\centering
% \vspace{-0.2cm}
\caption[Comparative Analysis of Tract-RLFormer with Supervised, Classical, and RL-based algorithms]{Comparison of mean Dice scores for the OR, PYT, and CC tracts for subject 1006 from TractoInferno dataset. Supervised learning scores are from \cite{poulin2022tractoinferno}; RL-based scores, with std. dev., are from \cite{theberge2024matters}. The last 2 rows includes scores for T-RLF and TD3, evaluated using our tract-specific approach. The highest and second highest scores are highlighted in green and red, respectively. `*' denotes tract-specific setting for methods.}
% \vspace{-0.2cm}
\begin{tabular}{ |p{4cm}|p{2.5cm}|p{2.5cm}|p{2.5cm}|  }
 \hline
\textbf{Algorithm}& \textbf{OR} &\textbf{PYT}&\textbf{CC}\\\cline{1-4}
 % \hline
  DET-SE   & 0.569    &0.665 &   0.658\\
  DET-Cosine&   0.598  & 0.708   &0.646\\
  % \hline
Prob-Sphere &0.599 & 0.695&  0.648\\
  Prob-Gaussian    &0.542 & 0.723&  0.668\\
  Prob-Mixture&   0.436  &  0.522& 0.614\\\cline{1-4}
 \hline
 % &Algorithm& Dice &OL&OR\\\cline{2-5}
 % \hline
  DET&   0.516  &  0.475& 0.345\\
  PROB&   0.549  &  0.740& 0.590\\
  PFT   & $0.644 \pm 0.136 $  &$0.753 \pm 0.010 $ &   $\colorbox{SpringGreen}{0.827} \pm 0.008 $\\ \cline{1-4}
  VPG& $0.369 \pm 0.135 $  & $0.434 \pm 0.128 $  & $0.428 \pm 0.182 $ \\
  % \hline
A2C  & $0.225 \pm 0.108 $ & $0.323 \pm 0.082 $&  $0.222 \pm 0.025 $\\
ACKTR  & $0.397 \pm 0.171 $ & $0.559 \pm 0.028 $&  $0.584 \pm 0.054 $\\
TRPO  & $0.330 \pm 0.154 $ & $0.498 \pm 0.062 $&  $0.594 \pm 0.048 $\\
PPO  &$0.440 \pm 0.187 $ & $0.619 \pm 0.042 $&  $0.650 \pm 0.028 $\\
DDPG &$0.612 \pm 0.063 $ & $0.630 \pm 0.045 $& $0.731 \pm 0.006 $ \\
TD3  & $0.555 \pm 0.097 $ & $0.603 \pm 0.045 $&  $0.688 \pm 0.035 $\\
  SAC    & $0.598 \pm 0.098 $ & $0.658 \pm 0.028 $&  $\colorbox{IndianRed1}{0.753} \pm 0.010 $\\
  SAC Auto& $0.608 \pm 0.088 $ & $0.655 \pm 0.032 $ & $0.747 \pm 0.019 $\\\cline{1-4}
 \hline
DET$^*$ & 0.648 & 0.752 & 0.713 \\
PROB$^*$ & \colorbox{IndianRed1}{0.652} & 0.765 & 0.731 \\
TD3$^*$ & 0.644 & \colorbox{IndianRed1}{0.764} & 0.720 \\
  T-RLF (Ours)   & \colorbox{SpringGreen}{0.673}    &\colorbox{SpringGreen}{0.772} &   0.738\\
 \hline
\end{tabular}
% \vspace{-1cm}
\label{table:comparetive}
\end{table}

\noindent Several insightful observations can be reported from the results of table \ref{table:comparetive}.

\noindent \textbf{Comparison with SOTA Classical and RL methods}:

In Table \ref{table:comparetive}, our framework outperforms the state-of-the-art method (PFT) for PYT and OR tracts, demonstrating its robustness in tract-specific tractography. Additionally, T-RLF shows comparable performance to state-of-the-art RL algorithms for CC tract.

\noindent \textbf{Comparison with TD3 agent}:

% Furthermore, the TD3 agent
The TD3 agent demonstrates markedly improved performance within our tract-specific generation framework. 
Testing of tract-specific TD3 on the ISMRM or HCP datasets cannot be conducted due to the absence of the evaluated tracts in these datasets.
However, the enhancement in TD3's performance in our tract-specific setting can be attributed to the training approach rather than dataset consistency. This is evidenced by TD3’s comparable or superior performance on different tracts across the ISMRM and HCP datasets, as detailed further in Tables \ref{table:exp:cg_af}, \ref{table:exp:pyt_cc}.

\noindent \textbf{Comparison with DET and PROB}:

% Moreover, dice
The Dice scores for DET and PROB improved for all tracts in the tract-specific setting, especially for CC, where DET increased by 106.67\% (0.345 to 0.713) and PROB by 23.89\% (0.590 to 0.731). The enhanced tracking performance of DET and PROB, despite not being trained, is indicative of the effectiveness of our tract-specific masks.
Also, in the whole-brain setting, there is a huge difference between DET (0.475) and PROB (0.740) scores on the PYT tract (Table \ref{table:comparetive}), whereas this gap is significantly smaller in the tract-specific setting (marked with `*'), where the tract-specific performance of DET$^*$ (0.752) and PROB$^*$ (0.765) align closely with each other and with the T-RLF and TD3 methods.
The consistency and stability observed for these classical methods are attributed to our tract-specific approach.

\subsection{Generalization Performance Evaluation}

In this section, we present the performance evaluation of our T-RLF model across three distinct datasets (Tables \ref{table:exp:cg_af}, \ref{table:exp:pyt_cc}), demonstrating its generalization performance. 
The results are averaged across test subjects for each dataset including five test subjects from the TractoInferno (TtoI) dataset (id: 1160, 1078, 1159, 1061, and 1171), four from the HCP dataset (id:  930449, 992774, 959574, and 987983), and one from the ISMRM phantom dataset. 

A visual comparison across datasets and subjects is presented in Fig. \ref{fig:result-plot-1}(a). We also compare the performance of T-RLF with classical algorithms, which were employed using tract-specific masks, and the tract-specific TD3 agent, from which the training data for T-RLF was derived (Fig. \ref{fig:result-plot-1}(b)).

\begin{figure}[hbt!]
% \begin{figure}[H]
    \centering
    \includegraphics[angle=90, width=15cm]{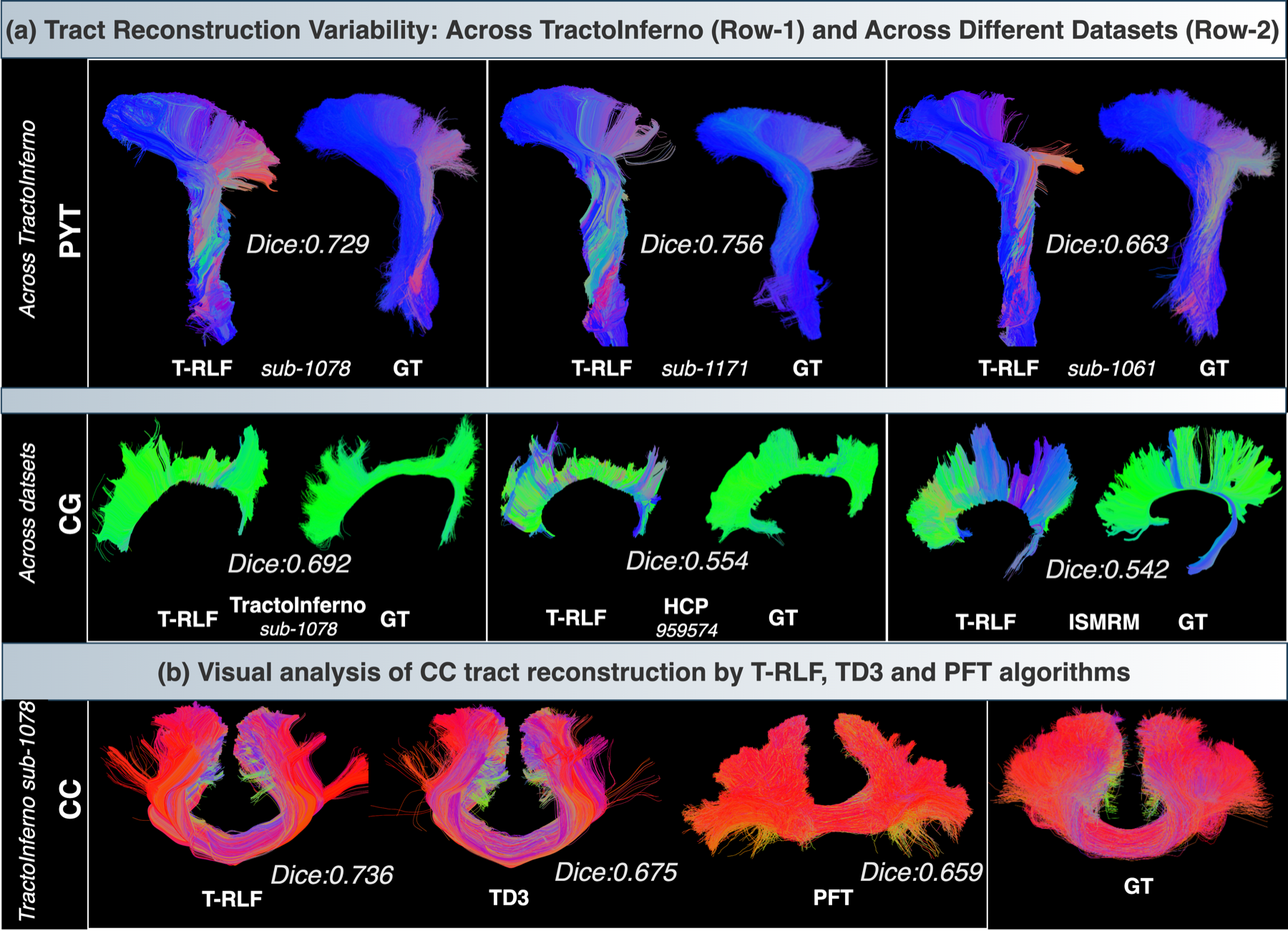}
    \caption[Results of Qualitative Analysis of Tract-RLForemer and other tractography methods]{Visual comparison of reconstructed tracts illustrating (a): Intra-dataset variability, Inter-dataset variability, and (b): Variability across tracts reconstructed by different algorithms. The depicted tracts include the left PYT, CG, and a part of CC. The algorithms evaluated in bottom section of figure are T-RLF (ours), TD3, and PFT.}
    \label{fig:result-plot-1}
\end{figure}

A quantitative comparison of generalization performance is presented in Tables \ref{table:exp:cg_af}, and \ref{table:exp:pyt_cc}, for the left and right Cingulum, Arcuate Fasciculus, Pyramidal Tract and a part of Corpus Callosum tract. The analysis is performed across all three datasets, with each table including only the datasets that contain the corresponding tract(s). We observe that T-RLF model displays a notable generalization performance. Interestingly, the classical deterministic (DET) and probabilistic (PROB) methods exhibit slightly better performance than learnable methods in some cases.

% for the tracts which are available in that dataset (some datasets dont have some tracts).

% In Table \ref{table:exp:cg_af}, we see that T-RLF model displays a notable generalization performance. Interestingly, the classical deterministic (DET) and probabilistic (PROB) methods exhibit slightly better performance than learnable methods in some cases (Tables \ref{table:exp:cg_af},\ref{table:exp:pyt_cc}).

\begin{table}[hbt]
% \vspace{-0.4cm}
\centering
\caption[Generalization performance evaluation of CG and AF tracts]{Performance metrics(in \%) for the CG and AF tracts, trained on the TractoInferno dataset and tested across multiple datasets to evaluate generalization. Tracking is performed using our proposed tract-specific generation method. A dash (`-') indicates the absence of ground-truth tracts in the corresponding dataset, precluding evaluation.
}
\begin{tabular}{ |p{0.1cm}|p{1.28cm}|p{0.6cm}p{0.6cm}p{0.7cm}|p{0.6cm}p{0.6cm}p{0.7cm}|p{0.6cm}p{0.6cm}p{0.7cm}|p{0.6cm}p{0.6cm}p{0.7cm}|  }
 \hline
    & &\multicolumn{6}{c|}{\textbf{Cingulum} (CG)}&\multicolumn{6}{c|}{\textbf{Arcuate Fasciculus}(AF)} \\\cline{3-14}
     & &\multicolumn{3}{c|}{Left} &\multicolumn{3}{c|}{Right} &\multicolumn{3}{c|}{Left} &\multicolumn{3}{c|}{Right} \\\cline{3-14}
\rotatebox{90}{\textbf{Data}} &\textbf{Algo.}& Dice &OvL&OvR& Dice &OvL&OvR& Dice &OvL&OvR& Dice &OvL&OvR\\\cline{1-14}
  &T-RLF   & 53.3  & 42.5 & 16.6 & 45.6   &33.7  & 13.9 & 61.8  & 51.2 & 13.4 & 41.8   &27.9  & 5.40 \\
  &TD3 &  53.0  &42.3   &16.9  &45.2   &33.6  & 14.3 & 61.6  & 51.0 & 13.7 & 41.6   &27.7  & 5.60 \\\cline{2-14}
\parbox[c][0pt][c]{\linewidth}{
  \centering
  \rotatebox[origin=c]{90}{
    \parbox{3cm}{\centering HCP}
  }
}  &DET & 55.2 & 46.4 & 21..5 & 52.6 & 41.7 & 16.3 & 62.8 & 52.5 & 14.0 & 43.9 & 30.0 & 6.6 \\
   &PROB    & 57.6 & 51.8 & 27.9 & 56.1 & 45.5 & 16.6  & 65.5 & 57.9 & 18.3 & 47.4& 33.7 & 8.6 \\
  &PFT  &  67.3 & 55.2 & 7.8 & 59.6  & 45.4  & 6.4  & 71.3  & 71.9 & 29.7 & 69.9  & 71.6 & 33.3 \\\cline{1-14}
  &T-RLF   & 61.0 & 56.8 & 28.6  &  56.5  &  52.6 & 34.9 & 52.7 & 45.1 &  27.8 &  39.5 & 36.5 & 49.8 \\
  &TD3 & 60.0  & 55.1  & 27.3 &  54.9  & 49.8  & 32.4 & 51.8  &  44.3 & 28.2  &  38.4   & 34.9 & 46.9 \\\cline{2-14}
 \parbox[c][0pt][c]{\linewidth}{
  \centering
  \rotatebox[origin=c]{90}{
    \parbox{3cm}{\centering TtoI}
  }
}&DET & 61.2 & 58.8 & 32.3 & 58.2 & 54.7 & 33.7 & 54.6 &46.3 &24.7  & 45.4 & 41.7 & 46.9 \\
   &PROB    & 67.9 & 69.1 & 33.6 & 64.7 & 64.6 & 36.7  & 62.3 & 57.2 & 27.2 & 50.3 & 48.3 & 50.9 \\
  &PFT & 55.9  & 48.8 & 25.4 &  54.5   & 51.3  & 38.6  & 62.8  & 60.1 & 31.5 &  53.9 & 62.8 & 88.0 \\\cline{1-14}
  &T-RLF   & 54.2    & 46.6 & 25.5 & 52.8    & 44.1 &  23.1 &  - & - & - & - & - &- \\
  &TD3 &  53.1  & 44.8   & 23.7  &   51.2  & 41.6   & 21.1  &  - & - & - & - & - &-  \\\cline{2-14}
 \parbox[c][0pt][c]{\linewidth}{
  \centering
  \rotatebox[origin=c]{90}{
    \parbox{3cm}{\centering ISMRM}
  }
} &DET & 57.5 & 51.9&  28.5 &57.7 & 52.4&  29.2 & - & - & - & - & - &- \\
   &PROB    & 61.1 & 59.3&  35.1 &64.0 & 65.4&  39.0  &  - & - & - & - & - &- \\
  &PFT & 55.4  &  49.4& 28.9 &   57.3  &  49.4& 22.9  &  - & - & - & - & - &- \\\cline{1-14}
 \hline
\end{tabular}
\label{table:exp:cg_af}
\vspace{-0.25cm}
\end{table}

\begin{table}[hbt]
% \vspace{-0.5cm}
\centering
\caption[Generalization performance evaluation of PYT and CC tracts]{Results are presented for the left and right parts of PYT and a segment of CC on the TractoInferno (TtoI) dataset. Tracking for all algorithms is conducted using our proposed tract-specific generation method.}
\vspace{-0.2cm}
\begin{tabular}{ |p{1.4cm}|p{1.3cm}|p{0.7cm}p{0.7cm}p{0.7cm}|p{0.7cm}p{0.7cm}p{0.7cm}|p{0.7cm}p{0.7cm}p{0.5cm}|  }
 \hline
    & &\multicolumn{6}{c|}{\textbf{Pyramidal Tract} (PYT)}&\multicolumn{3}{c|}{\textbf{Corpus Callosum}} \\\cline{3-8}
     & &\multicolumn{3}{c|}{Left} &\multicolumn{3}{c|}{Right} &\multicolumn{3}{c|}{(CC)} \\\cline{3-11}
\textbf{Dataset} &\textbf{Algo}.& Dice &OvL&OvR& Dice &OvL&OvR& Dice &OvL&OvR\\\cline{1-11}
  &T-RLF   & 70.3  & 64.1 & 17.2 & 70.1   &63.1  & 16.9 & 70.4  & 71.2 & 32.6  \\
  &TD3 &  69.4  &62.5   &15.9  &69.2   &61.4  & 15.8 & 68.1  & 64.8 & 26.1 \\\cline{2-11}
TtoI  &DET & 72.7 & 79.3 & 38.8 & 70.3 & 76.2 &40.7 & 70.1 & 72.6 & 35.8 \\
   &PROB    & 77.6 & 79.5 & 25.3 &74.8  & 72.5 & 21.3  & 72.6 & 76.4 & 36.7 \\
  &PFT & 66.2  & 55.7 & 12.4 & 65.9  & 57.8  & 17.4  & 54.9 & 51.2 & 36.1 \\
  % \cline{1-11}
 \hline
\end{tabular}
\label{table:exp:pyt_cc}
% \vspace{-0.75cm}
\end{table}

As previously mentioned in section \ref{subsec:comp-analysis}, the consistency observed in Tables \ref{table:exp:cg_af}, \ref{table:exp:pyt_cc} for the classical methods (DET and PROB) is due to our tract-specific approach. This improvement and stabilization may be attributed to the elimination of premature termination issues in narrow and deep WM regions, as described in \cite{girard2014towards}, facilitated by the refined spatial exploration enabled by MRM in our tract-specific approach.
It can be observed from Tables \ref{table:comparetive} and \ref{table:exp:pyt_cc}, that the performance of PFT declined in the tract-aware setting, dropping from 75\% to 66.2\% in the PYT and from 82\% to 55\% in the CC (refer Table \ref{table:exp:pyt_cc}).
This decline can be attributed to use of Continuous Map Criterion (CMC) \cite{girard2014towards} as a stopping criterion for fiber tracking. The CMC terminates fiber tracking based on Partial Volume Estimate (PVE) maps, allowing tractography to continue until the streamline correctly stops in the gray matter. This approach may generate fibers beyond our tract-specific masks, leading to increased overreach (see Fig. \ref{fig:result-plot-1}(b)) and consequently lower Dice scores. Furthermore, fibers generated outside the tracking mask may be erroneous and subsequently filtered or cleaned via FSS, resulting in a lower OvL score.

\textbf{Summarization: }In summary, our results demonstrate that we surpass supervised methods (Table \ref{table:comparetive}). Additionally, we consistently outperform the TD3 model (Tables \ref{table:comparetive}- \ref{table:exp:pyt_cc}), which served as the basis for training T-RLF. Notably, our tract-specific setting not only improves TD3 performance but also the performance of classical methods like DET and PROB compared to the whole-brain setting. This suggests a promising new direction of data driven policy learning for tract specific fiber generation in limited ground truth scenarios that can naturally scale up effectively.

\subsection{Ablation Study}
\label{subsec:ch3-abl}

In this section, we present an ablation study to determine the optimal configuration for our T-RLF model. We also examined the impact of two key components of our work: Mask Refinement Module (MRM) discussed in section \ref{subsec:ch3-mrm}, and the tract-specific policy fine-tuning as detailed in section \ref{subsec:prp}.

% In this section, we present an ablation study to determine the optimal configuration for our T-RLF model. 
Table \ref{table:dt-abl} compares the performance of different configurations of the GPT architecture, varying the number of attention heads (\textit{n\_heads}), context length (\textit{K}), and embedding dimension (\textit{d}). The models were evaluated based on their Dice scores for subject 1006 from the TractoInferno dataset, with scores averaged across the seven white matter tracts (see Section 3.2.1.2).

% \begin{table}[ht!]
\begin{table}[H]
\centering
\caption[Model architecture selection for Tract-RLFormer network]{Dice scores (in \%) averaged over 7 tracts of subject 1006 from TractoInferno dataset, at different values of T-RLF parameters: number of attention heads (\textit{n\_heads}), context length (\textit{K}), and embedding dimension (\textit{d}). Best score is in \textbf{bold}.}
% \vspace{-0.25cm}
\begin{tabular}{|c|c|c|c|c|c|c|}
\hline
\multicolumn{1}{|c|}{} & \multicolumn{2}{c|}{\textbf{$K = 20$}} & \multicolumn{2}{c|}{\textbf{$K = 30$}} & \multicolumn{2}{c|}{\textbf{$K = 40$}} \\
\cline{2-7}
\multicolumn{1}{|c|}{} & \textbf{ $d$=$128$ } & \textbf{ $d$=$512$ } & \textbf{ $d$=$128$ } & \textbf{ $d$=$512$ } & \textbf{ $d$=$128$ } & \textbf{ $d$=$512$ } \\
\hline
$n\_heads$ = 1 & 64.7 & 65.2 & 66.2 & 67.3 & \textbf{68.6} & 67.6 \\
$n\_heads$ = 2 & 63.4 & 66.3 & 66.2 & 67.5 & 68.0 & 68.2 \\
\hline
\end{tabular}
\vspace{-0.25cm}
\label{table:dt-abl}
\end{table}

It can be seen that the highest Dice score is achieved with \textit{n\_heads} = 1, \textit{K} = 40, and \textit{d} = 128, which was therefore selected for T-RLF. Additionally, the results highlight the importance of a larger context in tractography, demonstrating that a broader temporal receptive field can improve the model's ability to generate accurate fiber tracts.

\begin{table}[H]
% \vspace{-0.4cm}
\centering
\caption[Ablation study of the effect of tw-stage training process and MRM masks on Tractography performance]{Average performance metrics (in \%) obtained using Tract-RLFormer highlight the impact of the MRM on test subjects from the TractoInferno dataset. The table also compares the performance of the pre-trained network with the fine-tuned network post MRM application, illustrating the effect of policy fine-tuning on the same dataset.}
\begin{tabular}{ |p{1cm}|p{1cm}p{1cm}p{1cm}|p{1cm}p{1cm}p{1cm}|p{1cm}p{1cm}p{1cm}|  }
 \hline
  \textbf{Tract}  &\multicolumn{3}{c|}{\textbf{Without MRM}} &\multicolumn{6}{c|}{\textbf{With MRM}} \\\cline{2-10}
  & & & & \multicolumn{3}{c|}{Pre-trained} & \multicolumn{3}{c|}{Fine-tuned} \\    \cline{5-10}
 & Dice &OvL&OvR& Dice&OvL&OvR & Dice&OvL&OvR\\
   \cline{5-10}
   \hline
   PYT& 44.1 & 30.4 & 5.6  &65.9 & 55.1 & 11.8 & 70.3 & 67.2 & 24.7 \\
   CG& 41.1  & 45.4 & 80.3 & 51.5 & 45.7 & 30.3 & 58.7 & 54.7 & 31.7 \\
   AF& 34.2  & 34.4 & 65.1  & 45.8& 39.5 & 36.1 & 46.1 & 40.8 & 38.8 \\
   CC& 58.9  & 59.0 & 41.9 & 66.2 & 59.9 & 20.8 & 70.4 & 71.2 & 32.6 \\
   \hline

\end{tabular}
\label{table:abl:mrm}
% \vspace{-0.8cm}
\end{table}

The analysis of the effect of tract-specific masks generated by our Mask Refinement Module (MRM) is presented in Table \ref{table:abl:mrm}. The "Without MRM" scores are obtained by performing tractography within masks derived from converting Atlas streamlines into binary masks, as shown in Step 2 of Fig. \ref{fig:mrm}. We observe that Atlas-derived tracking masks led to a significant overreach (OvR), extending beyond actual region of interest. This OvR was notably reduced after incorporating MRM, leading to improved Dice and overlap metrics across all tracts.

We also present the effect of our two-stage training process, comparing the models obtained after both stages of mixed-tract pretraining and tract-specific finetuning. Table \ref{table:abl:mrm} reports the average scores across five test subjects from the TractoInferno dataset. It can be seen that fine-tuning the network specific to each tract allowed it to learn better and robust tract-specific diffusion characteristics, resulting in additional improvements in the performance metrics.

\section{Summary}

Tract-RLFormer is a data driven RL policy refinement framework for tractography. It is a tract-specific framework that integrates supervised and reinforcement learning paradigms. A distinctive feature of our Tract-RLFormer is its ability to train within the reinforcement learning experience space, independent of ground truth fibers. The fine-tuning stage of T-RLF model focuses and refines its capabilities in generating the tracts of interest. This approach demonstrates its excellent generalization performance across various datasets as well as scalability. This framework has the potential to utilize data from any reinforcement learning agents trained in diverse neuroimaging environments. Moreover, our innovative tract-specific modeling approach simplifies the reconstruction process by directly generating the target tract, thus avoiding the complex and error-prone segmentation step.

% \subsection{Inference, }

% \subsubsection{Evaluation Parameters}

% \section{}

\chapter{Data-driven RL Policy Fusion Framework with Multi-Critic finetuning for Tract-specific Tractography}\label{rlfusion}

\section{Motivation} 

In the previous chapter, we saw that Tract-RLFormer framework enables an effective data-driven RL policy refinement to directly reconstruct the tract of interest. Its GPT-based actor learns a refined policy entirely within the RL experience space, using mixed-tract pretraining and tract-specific fine-tuning to enable strong generalization across datasets and robust performance across diverse tract configurations.
Despite these advantages, our findings and previous RL-based tractography studies \cite{theberge2024matters} revealed that deterministic policies, such as TD3, tend to produce lower tract coverage and therefore higher false-negative rates, whereas stochastic policies, such as SAC, often generate erroneous pathways that result in increased false-positive connections. This trade-off is further discussed in Section \ref{subsec:fusion-results}.

Several studies \cite{schilling2019challenges,kamagata2024advancements} have emphasized this fundamental trade-off in the context of classical tractography: minimizing false negatives (Increasing Overlap) often leads to an increase in false positives (higher overreach), and vice versa.

This observation motivates the central idea of the present chapter: the potential of combining multiple diverse RL policies to achieve a more balanced and robust tractography behavior; a strategy that has shown promise in other domains, such as robotics and game-playing AI, where combining multiple policies has been demonstrated to improve generalization and robustness \cite{lillicrap2015continuous,marinov2024offline}.

Traditional RL policy ensembling strategies, such as decision-level aggregation (e.g., voting or averaging), often fail to leverage  semantic context or state history. On the other hand, complex ensemble techniques, such as Ensemble Policy Gradient or hierarchical/hybrid ensembles, introduce significant system complexity due to the need for agent synchronization, parameter sharing, and joint optimization, making them difficult to implement, debug, and maintain.
To sidestep these challenges, our TractRLFusion adopts a data-driven approach to fusion approach for combining multiple RL policies in tractography.

\section{Proposed \textit{TractRLFusion} Framework}
\textbf{TractRLFusion} is a novel GPT-based multi-policy fusion framework (as shown in Fig. \ref{fig:fusion-overview}). It integrates the following key components:

\begin{itemize}
    \item \textbf{Episodic Data Selection (EDS)} module is used to select tract-specific trajectories with anatomical precision, enabling data-driven fusion.
    \item \textbf{Multi-Critic Policy Fine-Tuning (MCPFT)} module is used for enhancing the policy fusion which enables robust, scalable, and generalizable tractography across diverse datasets.
\end{itemize}

\subsection{Overview and Experimental Setup}

We have utilized TractRLFusion framework to fuse three independently trained policies, namely TD3 \cite{fujimoto2018addressing}, SAC \cite{haarnoja2018soft}, and DDPG \cite{lillicrap2015continuous}, to combine their complementary strengths in tractography. While SAC learns in an exploration-driven manner (leading to high tract coverage at the cost of increased false positives), TD3 and DDPG lean towards conservative tracking (lower coverage with fewer false positives), further elaborated and shown in Section \ref{sec:fusion-results-comp}.

\vspace{-0.25cm}
\begin{figure}[H]
    \centering
    \includegraphics[width=\linewidth]{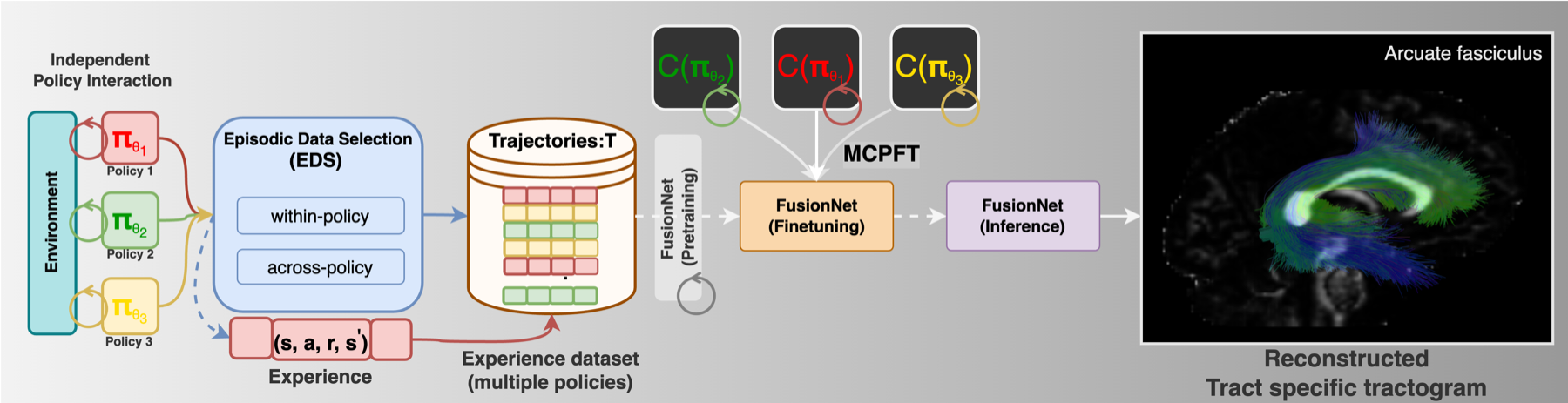}
    \caption[Illustration of TractRLFusion Framework]{\textit{TractRLFusion}: A generative framework where anatomically relevant trajectories with high expected returns from various RL agents(${\pi}_{{\theta}_{i}}$) are combined into a experience dataset (T) to train a generative model. The model leverages critic networks ($C({\pi}_{{\theta}_{i}})$) to align streamline generation with reward estimates, producing highly plausible reconstructions.}
    \label{fig:fusion-overview}
    \vspace{-0.25cm}
\end{figure}

TractRLFusion framework is composed of three key components, Episodic Data Selection (EDS), FusionNet, and Multi-Critic Policy Finetuning (MCPFT). 
To perform a data-driven fusion, we sample trajectories from trained RL policies, where Trajectories refer to sequences of return-to-go($r_{tg}$)--state($s$)--action($a$) obtained from an RL policy. A T-length trajectory is defined as: $\tau = (r_{tg_{0}}, s_{0}, a_{0}, r_{tg_{1}},.... r_{tg_{T}}, s_{T}, a_{T}$).

\textbf{(i)} Firstly EDS samples a mix of trajectories data from each of the 3 policies to train FusionNet on.
\textbf{(ii)} Followed by 2 stage training of our GPT-based FusionNet on EDS data to learn the fusion policy.
\textbf{(iii)} Finally FusionNet policy further undergoes refinement using MCPFT.

FusionNet learns a tract-specific policy for a given white matter tract and we train and test our policies in tract-specific masks obtained from our Mask Refinement Module (MRM) as described in section \ref{subsec:ch3-mrm} \cite{joshi2024tract}. Hence our proposed methods fall into the category of ROI-based tracking \cite{wasserthal2018tractseg}.
Among major tracts we focus on Occipital and pre/post central gyri of Corpus Callosum (CC), left and right Corticospinal Tract (CST), Arcuate Fasciculus (AF), and Cingulum (CG) tracts following prior works\cite{rheault2019bundle,theberge2024matters} in order to have a fair comparison. Datasets used in this study are already detailed in Section \ref{section:datasets}.

\subsection{Tract-Specific RL Policy Training} 
With an environment similar to \cite{theberge2021track} (Table \ref{tab:hyperparam}), our training is conducted via exploration within tract-specific masks of 5 train subjects from TractoInferno dataset, similar to previous chapter. The reward function is defined as the product of the action’s ($\vec{a}_t$) alignment with the most aligned fODF peak ($p$) and its dot product with the previous tracking direction ($\vec{u}_{t-1}$) :

\begin{equation}
    r_t = \left| \max_{\vec{p}_i} \left( \vec{p}_i \cdot \vec{a}_t \right)\right| \times \left( \vec{a}_t \cdot \vec{u}_{t-1} \right)
    \label{eqn:reward}
\end{equation}

Episode termination conditions include (i) exceeding the maximum episode length of 530 (corresponding to max. fiber length=200mm), (ii) exiting tract mask, and (iii) angle $>$ 60 degrees with the previous fiber segment.
The following paragraphs explain the key components of Tract-RLFusion.

\begin{table}[H]
\centering
\begin{tabular}{l|p{8.5cm}}
\hline
\textbf{Category} & \textbf{Details} \\
% \hline
% State Features & 
% - 45 SH \textit{coefficients} (6 neighbors + 1) \newline
% - 4 past directions in 3D making (4$\times$3 = 12 \textit{directions}) \newline
% - 6 neighbors + 1 binary tract-specific mask = 7 \textit{voxels}\newline
% - Total State dimension: ((45+1)$\times$ 7 ) + 12 = 334 \\
\hline
\textbf{Architectures} & 
- TD3/SAC/DDPG (Actor and Critics): \newline 
- 3-layer ReLU-activated FCNNs \newline
- Neurons per layer: 1024 \\
\hline
\textbf{Training Data} & 
- Train Subjects: 5 (IDs: 1030, 1079, 1119, 1180, 1198) \cite{poulin2022tractoinferno}\\
\hline
\textbf{Hyperparameters} & 
- Batches: 50 per subject \newline
- Batch size: 4096 transitions \newline
- Seeds: 7 per voxel \newline
- Step size: 0.375 mm \\
\hline
\textbf{Policy Hyperparameters} & 
- \textbf{TD3}: $lr$ = 8.56e-6, $action\_std$ = 0.334, $\gamma$ = 0.776 \newline
- \textbf{SAC}: $lr$ = 3.7e-5, $action\_std$ = 0.4, $\gamma$ = 0.89, $alpha$ = 0.076 \newline
- \textbf{DDPG}: $lr$ = 8.56e-6, $action\_std$ = 0.35, $\gamma$ = 0.5 \\
% \hline
\end{tabular}
\caption{Summary of Features, Architecture, Training Data, and Hyperparameters for training Tract-specific RL policies.}
\label{tab:hyperparam}
\end{table}

\subsection{Episodic Data Selection}
\label{subsec:EDS}

It is the trajectory data selection process for the pre-training and fine-tuning stages of our FusionNet model (see Fig. \ref{fig:eds}). It involves following steps:
% to obtain "expert" demonstrations from three actor-critic RL policies (TD3, DDPG and SAC).

\textbf{(i)} Tracking is performed using each trained policy (TD3, SAC and DDPG) from the same batch of neighbouring initial seed points (sampled from tract mask of a given subject), resulting in streamlines \emph{(x,y,z)} and the corresponding {$(r_{tg}, s, a)$} trajectories (see Fig. \ref{fig:eds}(i)).

\begin{figure}[H]
    \centering
    \includegraphics[angle=90,height=0.85\textheight]{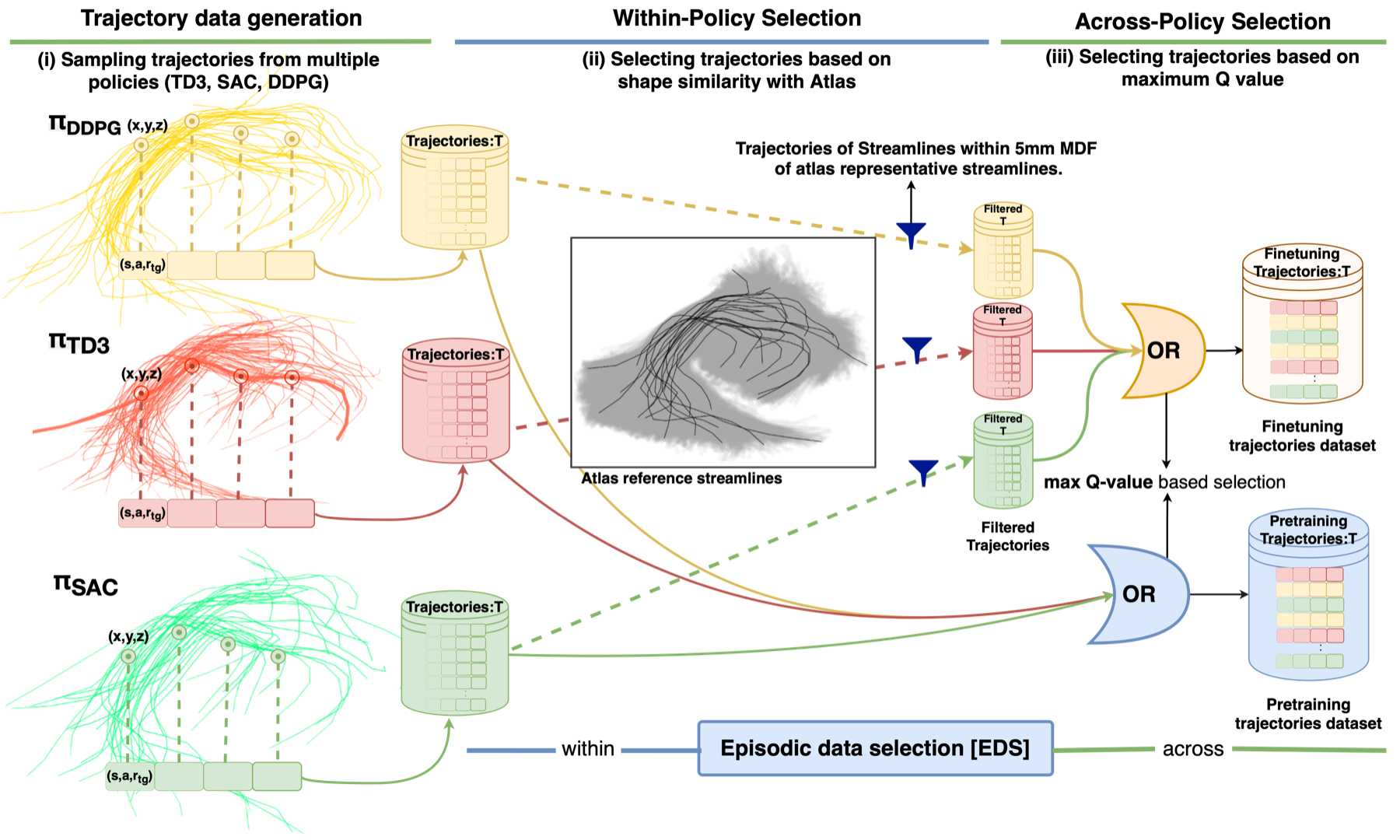}
    \caption[Visual depiction of \textit{Episodic Data Selection}]{Visual depiction of \textit{Episodic Data Selection}. Row 1 illustrates within-policy trajectory selection using the MDF\cite{garyfallidis2012quickbundles} constraint (Eq. 2) with atlas reference streamlines. Row 2 depicts across-policy selection via the Q-value constraint (Eq. 3), where expected returns are compared. For each region $R_i$, trajectories from the policy $\pi$ with the highest expected return is selected to populate the dataset (T).}
    \label{fig:eds}
\end{figure}

% Per subject.
\textbf{(ii)} Trajectories shorter than $47$ transitions (corresponding to approximately $20mm$ of fiber length for a $1mm^{3}$ isometric voxel size and a $0.375mm$ step size) are filtered-out and the rest undergo batch-wise \emph{across-policy} (Fig. \ref{fig:eds} (ii)) selection and \emph{within-policy} (Fig. \ref{fig:eds} (iii)) selection. This threshold was chosen to exclude trajectories corresponding to anatomically implausible short streamlines.

\textbf{(iii)} \textbf{Within-Policy trajectory selection:} Fifteen reference streamlines are selected from the atlas tract using farthest streamline sampling \cite{li2023farthest}. Streamlines for each policy are filtered such that trajectories of streamlines having MDF (Mean Direct Flip) distance \cite{garyfallidis2012quickbundles} less than 5mm with the reference streamlines are selected. This within-policy selection enables the selection of the best trajectories in terms of anatomical shape of any given tract.

\textbf{(iv)} \textbf{Across policy selection} selects one policy out of the three, based on the maximum expected Q-value of its MDF-filtered trajectories. These trajectories are selected for the finetuning dataset. Whereas, for pretraining dataset, Q-value based selection happens directly on unfiltered trajectories, as depicted in Fig. \ref{fig:eds}.

Neighbouring seeds are selected in each batch, enabling EDS to select best trajectories from all regions of the given tract. This is repeated for all tracts of the 5 train subjects, followed by length-based and random selection, similar to chapter 3 (\ref{subsubsec:ch3-data-gen}), resulting in a Pretraining dataset of 150000 trajectories and finetuning dataset of 50000 trajectories. It is important to note that the pretraining dataset is made from trajectories of all tracts, whereas finetuning data consists of tract-specific trajectories.

\subsection{FusionNet Architecture and Training}
\label{subsec:fnet-training}

Central to the TractRLFusion framework, FusionNet, described in Table \ref{tab:fusionnet_params}, is a 4-layer GPT \cite{radford2018improving} that models RL trajectories as sequences which undergoes general pretraining and tract-specific fine-tuning on expert trajectories from EDS. Specifically, the first three layers are trained on mixed trajectories from the EDS pretraining data over 30 iterations, while the final layer is fine-tuned for 10 iterations using tract-specific trajectories from the EDS-Finetune data. This two-stage training strategy is designed to firstly capture broad trajectory patterns and then adapt to the fine-grained tract-specific details. Similar to Chapter 3, we use a 5-step angular loss ($\mathcal{L}_{dist_{cos}}$) between predicted ($\hat{\textbf{a}}$) and actual ($\textbf{a}$) actions, defined as:
% {\setlength{\abovedisplayskip}{5pt}\setlength{\belowdisplayskip}{5pt}
\begin{equation} \label{supervised_loss}
    \mathcal{L}_{dist_{cos}} = \sum_{t=2}^{K-2} \left( \sum_{i=-2}^{2} \cos^{-1} \left( \textbf{a}_{t+i} \cdot \hat{\textbf{a}}_{t+i} \right) \right)
\end{equation}

where $K$ denotes the context length of the FusionNet model.

\begin{table}[hbt!]
\centering
\begin{tabular}{l|p{7.2cm}} % Using p{} column for vertical stacking
\hline
\textbf{Category} & \textbf{Details} \\
\hline
\textbf{Architecture} & 
- GPT Layers: 4  \newline
- Embedding Dim: 128 \newline
- Attention Heads: 1 \newline
- Context Length ($l_{context}$): 40 \newline
- Activation: ReLU \newline
- Dropout: 0.1 \\
\hline
\textbf{Training} & 
- Optimizer: AdamW ; LR: 1e-4 \newline
- Weight Decay: 1e-4 \newline
- Batch Size: 128 \newline
- Iters:30(General),10(Tract-specific) \newline
- Updates/iter: 10000 \newline
- Warmup Steps: 10000 \\
\hline
\textbf{MCPFT} & 
- LR: 1e-4 ; Batch Size: 512 \newline
- Iters: 25 \newline
- Updates/iter: 1000 (actor), 1 (critic) \\
\hline
\textbf{Tracking} & 
- Initial Return-to-Go: 300 \\
\hline
\end{tabular}
\caption{Summary of FusionNet architecture, training, and tracking hyperparameters.}
\label{tab:fusionnet_params}
\end{table}

The model was trained using the AdamW optimizer. Key hyperparameters, summarized in Table \ref{tab:fusionnet_params}, were tuned to maximize Accumulated sum of rewards and Dice Scores.
We observed that further training with the supervised loss (Eq. \ref{supervised_loss}) saturates the performance, showing limited improvement.

\subsection{Multi-critic Policy Fine-tuning}

FusionNet policy is fine-tuned further using our multi-critic setup, known as Multi-critic Policy Fine-tuning (MCPFT). 
% Once FusionNet is trained to learn a fusion policy by offline training on EDS data,
Having learnt a fused policy by offline training on EDS data, FusionNet is further trained in an actor-critic setup using the critic networks from original RL policies of SAC, DDPG, and TD3. Figure \ref{fig:mcpft} shows the MCPFT module in detail.

% \begin{figure}[hbt!]
\begin{figure}[H]
    \centering
    \includegraphics[width=\linewidth]{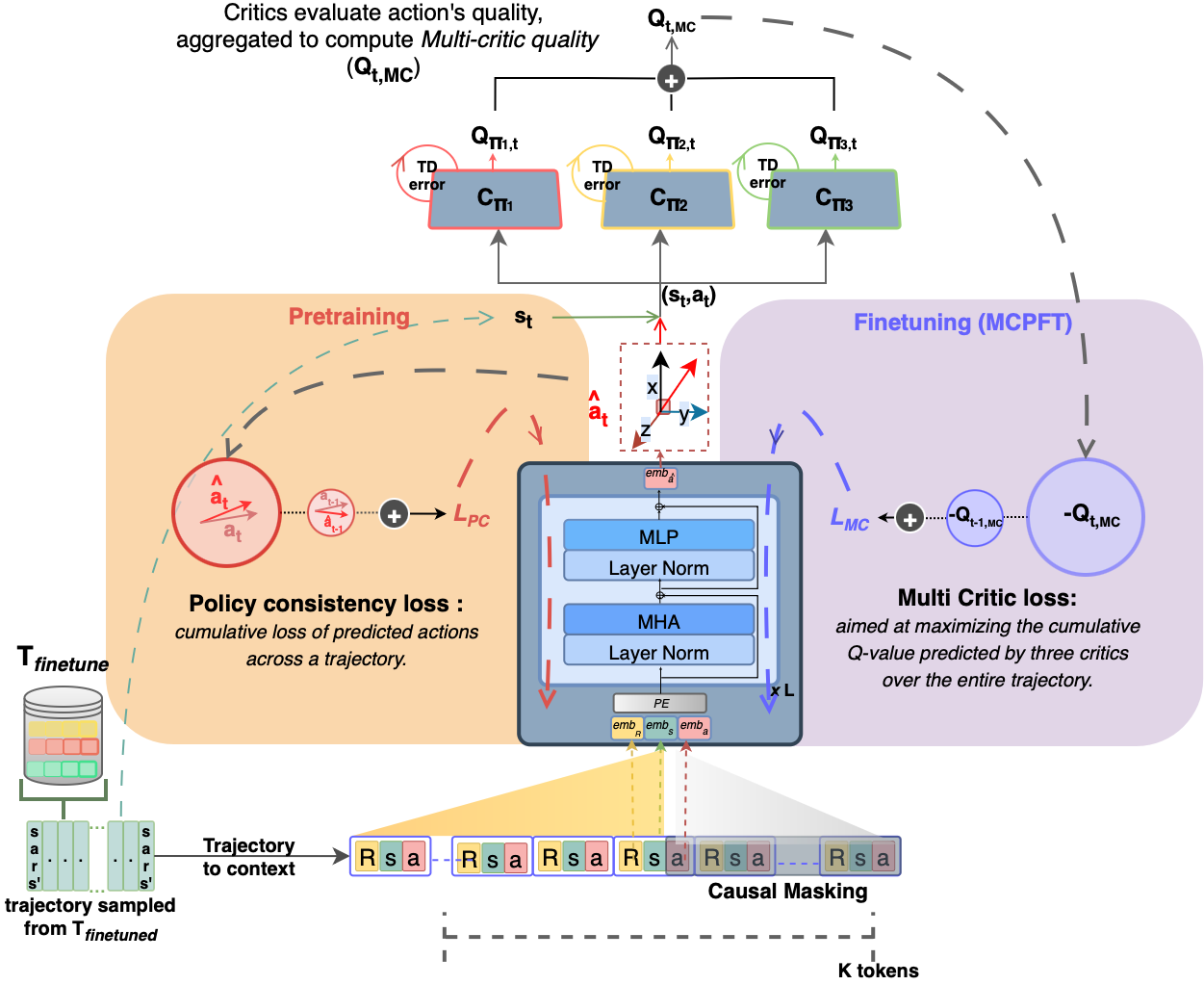}
    \caption[Illustration of Multi-Critic Policy Finetuning Module]{Visualization of MCPFT module: It further finetunes the FusionNet model using aggregated critic loss in addition to the original 5-step loss, interpreted as policy consistency loss.}
    \label{fig:mcpft}
\end{figure}

In this setting, the critics update with their RL losses $\mathcal{L}_{\text{critic}, k}$, while the actor
(FusionNet) optimizes using $\mathcal{L}_{\text{actor}}$ loss as defined below:

\begin{equation}
\mathcal{L}_{\text{actor}} = \mathcal{L}_{dist_{cos}} + \sum_{k=1}^{3} \mathcal{L}_{\text{critic}, k}, \quad \text{where } \mathcal{L}_{\text{critic}, k} = - \sum_{t=0}^{K} Q_{C(\pi_k)}(s_t, \hat{a}_t)
\end{equation}
\label{eqn:mcpft}
% here/ where the k is an iterator of the 3 crtic networks, K is the context length of FusionNet, and $Q_{C(\pi_k)}$ is the Q value obtained from the kth critic, having policy $\pi_k$ for action $\hat{a}_t$ at state $s_{t}$.

where $k$ indexes the three critic networks, $K$ denotes the context length of FusionNet, and $Q_{C(\pi_k)}$ is the Q-value predicted by the $k^{th}$ critic, conditioned on its policy $\pi_k$, for taking action $\hat{a}_t$ at state $s_{t}$.

As discussed in Section~\ref{subsec:fnet-training}, the performance of FusionNet saturates after 40 (pre-training and finetuning) training iterations. 
To further improve the FusionNet policy beyond this saturation point, it is natural to turn to the critics: unlike the 5-step loss, which primarily preserves previously learned behaviour, the critic networks can provide richer information about expected return across the state–action space. Using critic feedback for additional learning can therefore expose the FusionNet actor to value-based gradients that were not fully exploited during the initial training stages.

Moreover, using multiple critics to guide an actor is a common design choice in reinforcement learning. Multi-critic architectures are frequently employed to obtain more reliable value estimates, reduce over or under-estimation bias, or encourage more expressive behaviour \cite{mysore2022multi,haarnoja2018soft,fujimoto2018addressing}. Motivated by this general principle, we explored the integration of the three critics to improve the fused FusionNet policy.

However, critic-driven updates alone were insufficient to preserve the actor’s previously learned behaviour. To address this, we fine-tuned FusionNet using a combined objective that incorporates both our 5-step cosine-distance loss and the aggregated critic loss. Empirically, we observed that the $\mathcal{L}_{dist_{cos}}$ component stabilizes training by retaining the actor’s prior policy structure, effectively anchoring the model to its pre-trained behaviour, while the aggregated Q-value term provides additional information to refine the policy based on value estimates from all three critics. This combination enables FusionNet to benefit from multi-critic feedback without destabilizing the actor, resulting in improved fine-tuning performance compared to using either component alone.

For fine-tuning under this objective (defined in \ref{eqn:mcpft}), we trained the FusionNet model for 25 iterations, with each iteration consisting of 1,000 actor (FusionNet) updates and a single update for each critic. This update schedule intentionally delays critic updates, which we found to mitigate saturation effects caused by rapidly changing or unstable value estimates. A batch size of 512 trajectories was used throughout training.

\subsection{Tract Generation and Cleaning}

Using FusionNet policy, tracking is performed for different tracts within their respective tracking masks, by seeding at 7 seeds per voxel. Return-to-go is initialized (as $r_{tg0}$) to 300 for FusionNet, and dataset-specific step sizes (0.375mm for TractoInferno, 0.468mm for HCP, 0.75mm for ISMRM) are fixed similar to chapter 3. As a postprocessing step, tract filtering is performed using Fast Streamline Search \cite{st2022fast} with atlas reference tracts \cite{francois_rheault_2023_7950602}, as elaborated previously in Section \ref{subsec:Tract_cleaning}.

\section{Experiments and Results}
\label{subsec:fusion-results}

We evaluate our performance using five test subjects (IDs: 1160, 1078, 1061, 1159, 1171) from the TractoInferno dataset. Our generalization performance is assessed on 4 diverse subjects from the HCP \cite{van2012human} dataset (IDs: 930449, 959574, 992774, 987983); (reference tracts from \cite{wasserthal2018tractseg}) and on the ISMRM dataset \cite{maier2017challenge}. Our evaluation compares the Dice, Overlap (OL), and Overreach (OR) scores.

% \textbf{Baselines} \\
We have compared FusionNet against the following \textbf{baselines}.
\begin{itemize}
    \item \textbf{Classical methods} DET \cite{basser2000vivo} and PROB (iFOD1) \cite{tournier2012mrtrix}.
    \item \textbf{Deep RL policies} \cite{theberge2021track,theberge2024matters} SAC, TD3, and DDPG, employed to learn our fusion policy.
    
    \item \textbf{Tract-specific method} TractSeg \cite{wasserthal2018tractseg} was evaluated within its own tract masks. For a fair comparison, FusionNet was also evaluated using the masks generated by TractSeg (presented as FusionNet$^{*}$ in Tables \ref{table:exp:AF_CC} and \ref{table:exp:CG_CST}).

    \item \textbf{Policy Ensembles} $\pi_{maxQ}$ and $\pi_{Avg}$ that combine SAC, TD3, and DDPG at decision level by selecting the action with the max normalized Q-value, and average of the 3 actions, respectively, at each step.
\end{itemize}

\subsection{Comparative Study}
\label{sec:fusion-results-comp}

Tables \ref{table:exp:AF_CC} and \ref{table:exp:CG_CST} present a comprehensive comparison, while Fig. \ref{fig:fusion-plot} shows a qualitative comparison among the best performing RL (SAC) and classical (PROB) algorithm. 
TractOracle \cite{theberge2024tractoracle} was excluded from our comparisons, as it is specifically designed for seeding from the white matter–gray matter (WM/GM) interface.
Although the TractoInferno dataset does not explicitly annotate the corticospinal tract (CST), it does include the pyramidal tract (PYT), which contains the core components of the CST. Prior work by Dumais et al. \cite{dumais2023fiesta} directly compared the PYT (as extracted using the RBx method in TractoInferno) with the CST (as extracted using the TractSeg method in the HCP dataset), demonstrating that the two tracts are anatomically and functionally equivalent for tract-level analyses. Accordingly, in our experiments we report CST results for the HCP and ISMRM datasets, and PYT results for the TractoInferno dataset.

\begin{figure}[h]
    \centering
    \includegraphics[width=\linewidth]{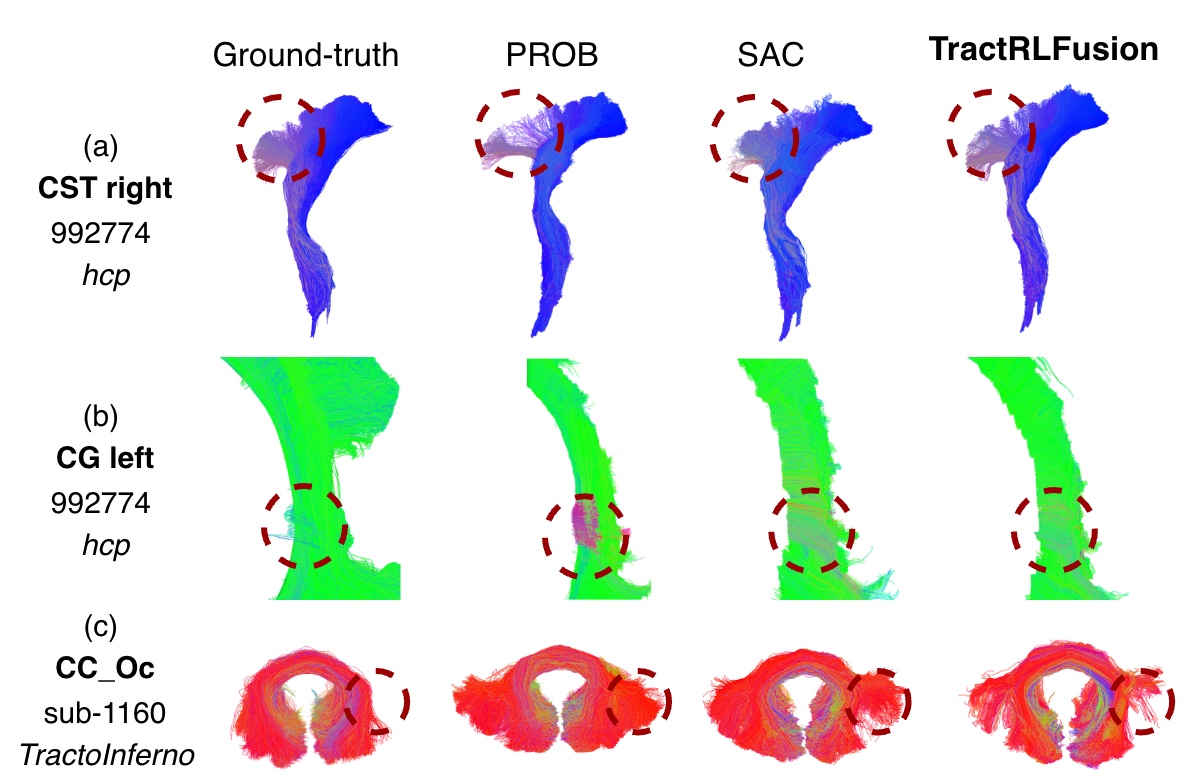}
    \caption[Visualization of Qualitative Comparison of TractRLFusion]{Visualization of (a) right corticospinal tract, and (b) part of the left cingulum tract, from subject 992774 in the HCP dataset. (c) shows the occipital part of the corpus callosum from subject 1160 in the TractoInferno dataset; for comparison of proposed method with PROB and SAC (other top-performing algorithms considered in this study) and ground-truth tracts. Dashed circles highlight key regions for comparison.}
    \label{fig:fusion-plot}
\end{figure}

\begin{table}[hbt]
\centering
\caption{Comparison of performance metrics (\%) for AF and CC, showing generalization across HCP, TractoInferno(TtoI), and ISMRM datasets. Metrics (Dice, OL, OR) are averaged across Left/Right for AF, and across Oc/Pr\_Po for CC.}
\begin{tabular}{ |p{0.5cm}|p{2.1cm}|p{1.5cm}p{1.5cm}p{1.5cm}|p{1.5cm}p{1.5cm}p{1.5cm}| }
 \hline
  &  & \multicolumn{3}{c|}{\textbf{AF}} & \multicolumn{3}{c|}{\textbf{CC}} \\ \cline{3-8}
\rotatebox{90}{\textbf{Data}} & \textbf{Algorithm} & Dice \textuparrow & OL \textuparrow & OR \textdownarrow & Dice \textuparrow & OL \textuparrow & OR \textdownarrow \\ \cline{1-8}
 & DET & 53.4$\substack{+\\-}$5.0 & 41.3$\substack{+\\-}$7.4 & 10.3$\substack{+\\-}$4.6 & 70.9$\substack{+\\-}$1.1 & 77.2$\substack{+\\-}$2.3 & 41.2$\substack{+\\-}$3.4 \\
 & SAC & 54.6$\substack{+\\-}$4.7 & 42.4$\substack{+\\-}$6.5 & 11.4$\substack{+\\-}$3.2 & 70.4$\substack{+\\-}$0.2 & 80.4$\substack{+\\-}$2.9 & 48.1$\substack{+\\-}$5.2 \\
 & DDPG & 50.6$\substack{+\\-}$5.5 & 37.9$\substack{+\\-}$7.6 & 8.3$\substack{+\\-}$3.7 & 65.0$\substack{+\\-}$1.2 & 66.2$\substack{+\\-}$5.4 & 37.3$\substack{+\\-}$6.5 \\
\parbox[c][0pt][c]{\linewidth}{
  \centering
  \rotatebox[origin=c]{90}{
    \parbox{3cm}{\centering HCP}
  }
} & TD3 & 51.7$\substack{+\\-}$5.3 & 39.5$\substack{+\\-}$7.8 & 9.6$\substack{+\\-}$3.9 & 64.9$\substack{+\\-}$1.2 & 66.2$\substack{+\\-}$3.8 & 37.2$\substack{+\\-}$6.6 \\
 % & TRLF$_{S}$ & 50.9 & 38.0 & \textbf{8.2} & 67.6 & 67.5 & \textbf{30.9} \\
 & $\pi_{maxQ}$ & 54.8$\substack{+\\-}$4.7 & 42.3$\substack{+\\-}$7.1 & 10.9$\substack{+\\-}$3.5 & 70.4$\substack{+\\-}$0.6 & 81.6$\substack{+\\-}$3.4 & 50.1$\substack{+\\-}$5.4 \\
 & $\pi_{avg}$ & 52.3$\substack{+\\-}$4.1 & 40.5$\substack{+\\-}$7.7 & 12.7$\substack{+\\-}$6.4 & 70.1$\substack{+\\-}$0.7 & 78.4$\substack{+\\-}$1.9 & 46.1$\substack{+\\-}$2.6 \\ 
 & \textbf{FusionNet} & \textbf{55.5$\substack{+\\-}$4.8} & \textbf{43.8$\substack{+\\-}$6.9} & 12.3$\substack{+\\-}$5.2 & \textbf{72.4$\substack{+\\-}$0.4} & 81.8$\substack{+\\-}$1.7 & 44.8$\substack{+\\-}$2.6 \\ \cline{2-8}
 & TractSeg & 70.9$\substack{+\\-}$1.5 & 59.5$\substack{+\\-}$2.5 & \textbf{8.1$\substack{+\\-}$0.9} & 76.8$\substack{+\\-}$2.2 & 68.5$\substack{+\\-}$3.4 & \textbf{11.7$\substack{+\\-}$1.3} \\
 & \textbf{FusionNet*} & \textbf{74.1$\substack{+\\-}$3.6} & \textbf{66.6$\substack{+\\-}$6.3} & 13.0$\substack{+\\-}$2.2 & \textbf{77.4$\substack{+\\-}$1.8} & \textbf{78.6$\substack{+\\-}$2.8} & 23.9$\substack{+\\-}$3.1 \\
\Xhline{2\arrayrulewidth} 
 & DET & 50.0$\substack{+\\-}$6.5 & 44.0$\substack{+\\-}$3.1 & 35.8$\substack{+\\-}$9.1 & 49.7$\substack{+\\-}$4.8 & 39.5$\substack{+\\-}$5.2 & \textbf{19.6$\substack{+\\-}$8.7} \\
 & SAC & 52.3$\substack{+\\-}$8.9 & \textbf{52.4$\substack{+\\-}$4.7} & 52.1$\substack{+\\-}$9.6 & 54.6$\substack{+\\-}$4.6 & 48.3$\substack{+\\-}$8.6 & 28.6$\substack{+\\-}$8.2 \\
 & DDPG & 46.3$\substack{+\\-}$7.0 & 39.3$\substack{+\\-}$2.5 & \textbf{28.6$\substack{+\\-}$9.9} & 49.6$\substack{+\\-}$5.6 & 40.4$\substack{+\\-}$8.0 & 22.1$\substack{+\\-}$8.9 \\
\parbox[c][0pt][c]{\linewidth}{
  \centering
  \rotatebox[origin=c]{90}{
    \parbox{3cm}{\centering TtoI}
  }
} & TD3 & 45.3$\substack{+\\-}$8.3 & 39.6$\substack{+\\-}$4.2 & 37.6$\substack{+\\-}$9.3 & 48.1$\substack{+\\-}$6.2 & 38.3$\substack{+\\-}$7.4 & 20.9$\substack{+\\-}$5.3 \\
 % & TRLF$_{S}$ & 47.5 & 43.5 & 43.2 & 46.9 & 37.0 & 21.1 \\
 & $\pi_{maxQ}$ & 52.7$\substack{+\\-}$8.5 & 51.9$\substack{+\\-}$4.4 & 47.3$\substack{+\\-}$9.5 & 54.2$\substack{+\\-}$4.1 & 47.0$\substack{+\\-}$7.9 & 30.7$\substack{+\\-}$8.8  \\
 & $\pi_{avg}$ & 52.1$\substack{+\\-}$7.5 & 51.9$\substack{+\\-}$7.6 & 49.5$\substack{+\\-}$8.7 & 53.3$\substack{+\\-}$7.2 & 47.3$\substack{+\\-}$8.6 & 31.5$\substack{+\\-}$5.8  \\
 & \textbf{FusionNet} & \textbf{53.2$\substack{+\\-}$8.9} & 51.3$\substack{+\\-}$4.3 & 39.1$\substack{+\\-}$9.3 & \textbf{56.5$\substack{+\\-}$3.2} & \textbf{50.5$\substack{+\\-}$6.1} & 28.8$\substack{+\\-}$6.6 \\ \cline{1-8}
% \Xhline{2\arrayrulewidth} 
%  & DET & - & - & - & - & - & - \\
%  & PROB & - & - & - & - & - & - \\ \cline{2-8}
%  & SAC & - & - & - & - & - & - \\
%  & DDPG & - & - & - & - & - & - \\
% ISMRM & TD3 & - & - & - & - & - & - \\
%  & TRLF$_{S}$ & - & - & - & - & - & - \\
%  & \textbf{FNet} & - & - & - & - & - & - \\ \cline{1-8}
 \hline
\end{tabular}
\label{table:exp:AF_CC}
% \DrawBox[ultra thick, draw=green]{AF hcp 2 top}{AF hcp 2 bottom}
% \DrawBox[ultra thick, draw=green]{AF ttoi 1 top}{AF ttoi 1 bottom}
% \DrawBox[ultra thick, draw=green]{AF ttoi 2 top}{AF ttoi 2 bottom}
% \DrawBox[ultra thick, draw=green]{CC hcp 1 top}{CC hcp 1 bottom}
% \DrawBox[ultra thick, draw=green]{CC hcp 2 top}{CC hcp 2 bottom}
% \DrawBox[ultra thick, draw=green]{CC ttoi 1 top}{CC ttoi 1 bottom}
% \DrawBox[ultra thick, draw=green]{CC ttoi 2 top}{CC ttoi 2 bottom}
\end{table}

Comparison of Dice, OL, and OR metrics together provides a more comprehensive assessment of performance rather than their individual comparison. A high Dice score indicates a favorable balance between OL and OR, with regards to trade-off between maximizing OL (capturing more true anatomy) and minimizing OR (reducing false positives).

\begin{table}[hbt]
\centering
\caption{Comparison of performance metrics (\%) for CG and CST, showing generalization across HCP, TractoInferno(TtoI), and ISMRM datasets. Metrics (Dice, OL, OR) are averaged across Left/Right.}
\begin{tabular}{ |p{0.5cm}|p{2.1cm}|p{1.5cm}p{1.5cm}p{1.5cm}|p{1.5cm}p{1.5cm}p{1.5cm}| }
 \hline
  &  & \multicolumn{3}{c|}{\textbf{CG}} & \multicolumn{3}{c|}{\textbf{CST/PYT}} \\ \cline{3-8}
\rotatebox{90}{\textbf{Data}} & \textbf{Algorithm} & Dice \textuparrow & OL \textuparrow & OR \textdownarrow & Dice \textuparrow & OL \textuparrow & OR \textdownarrow \\ \cline{1-8}
 & DET & 53.9$\substack{+\\-}$2.4  & 44.1$\substack{+\\-}$3.3 & 18.9$\substack{+\\-}$2.8 & 62.9$\substack{+\\-}$1.7 & 51.8$\substack{+\\-}$4.8 & 11.7$\substack{+\\-}$3.8 \\
 & SAC & 56.2$\substack{+\\-}$1.9 & 47.3$\substack{+\\-}$3.6 & 20.2$\substack{+\\-}$3.7 & 67.7$\substack{+\\-}$0.7 & 60.8$\substack{+\\-}$0.7 & 17.4$\substack{+\\-}$3.2 \\
 & DDPG & 48.0$\substack{+\\-}$3.4 & 35.8$\substack{+\\-}$3.6 & \textbf{12.8$\substack{+\\-}$1.8} & 66.1$\substack{+\\-}$0.4 & 60.4$\substack{+\\-}$4.3 & 20.4$\substack{+\\-}$6.3 \\
\parbox[c][0pt][c]{\linewidth}{
  \centering
  \rotatebox[origin=c]{90}{
    \parbox{3cm}{\centering HCP}
  }
} & TD3 & 49.2$\substack{+\\-}$2.8 & 37.5$\substack{+\\-}$3.5 & 14.7$\substack{+\\-}$2.7 & 55.7$\substack{+\\-}$1.2 & 46.2$\substack{+\\-}$5.4 & 15.8$\substack{+\\-}$6.1 \\
 % & TRLF$_{S}$ & 54.0 & 41.6 & 12.3 & 61.6 & 54.1 & 15.7 \\
 & $\pi_{maxQ}$ & 56.3$\substack{+\\-}$2.5 & \textbf{47.5$\substack{+\\-}$3.8} & 20.7$\substack{+\\-}$3.3 & 67.9$\substack{+\\-}$1.1 & 59.5$\substack{+\\-}$1.8 & 14.9$\substack{+\\-}$4.3 \\
 & $\pi_{avg}$ & 56.0$\substack{+\\-}$1.4 & 46.6$\substack{+\\-}$2.4 & 19.6$\substack{+\\-}$2.9 & 69.2$\substack{+\\-}$0.8 & \textbf{63.7$\substack{+\\-}$0.7} & 17.6$\substack{+\\-}$3.7 \\ 
 & \textbf{FusionNet} & \textbf{56.6$\substack{+\\-}$2.0} & 47.2$\substack{+\\-}$3.1 & 19.2$\substack{+\\-}$2.9 & \textbf{69.4$\substack{+\\-}$1.3} & 62.5$\substack{+\\-}$2.2 & 17.0$\substack{+\\-}$7.1 \\\cline{2-8}
 & TractSeg & 76.8$\substack{+\\-}$3.7 & 73.1$\substack{+\\-}$5.8 & \textbf{17.2$\substack{+\\-}$3.4} & 80.8$\substack{+\\-}$1.7 & 73.2$\substack{+\\-}$3.1 & \textbf{7.8$\substack{+\\-}$2.3} \\
  & \textbf{FusionNet*} & \textbf{77.0$\substack{+\\-}$1.7} & \textbf{86.7$\substack{+\\-}$3.7} & 38.2$\substack{+\\-}$6.5 & \textbf{82.2$\substack{+\\-}$1.1} & \textbf{79.8$\substack{+\\-}$3.2} & 14.4$\substack{+\\-}$3.9\\
\Xhline{2\arrayrulewidth} 
 & DET & 59.7$\substack{+\\-}$5.5 & 56.8$\substack{+\\-}$7.1 & 33.0$\substack{+\\-}$9.6 & 71.5$\substack{+\\-}$4.8 & 77.8$\substack{+\\-}$9.0 & 39.8$\substack{+\\-}$8.5 \\
 & SAC & 63.1$\substack{+\\-}$5.6 & 70.9$\substack{+\\-}$7.9 & 55.9$\substack{+\\-}$2.9 & 73.0$\substack{+\\-}$4.4 & 70.3$\substack{+\\-}$9.2 & 22.0$\substack{+\\-}$8.6 \\
 & DDPG & 59.9$\substack{+\\-}$6.9 & 55.7$\substack{+\\-}$7.2 & 30.4$\substack{+\\-}$9.7 & 69.2$\substack{+\\-}$4.3 & 60.7$\substack{+\\-}$8.7 & \textbf{14.3$\substack{+\\-}$5.8} \\
\parbox[c][0pt][c]{\linewidth}{
  \centering
  \rotatebox[origin=c]{90}{
    \parbox{3cm}{\centering TtoI}
  }
} & TD3 & 57.5$\substack{+\\-}$6.1 & 52.4$\substack{+\\-}$6.3 & \textbf{29.9$\substack{+\\-}$7.4} & 69.6$\substack{+\\-}$5.4 & 62.1$\substack{+\\-}$6.5 & 15.9$\substack{+\\-}$6.3 \\
 % & TRLF$_{S}$ & 62.5 & 63.3 & 39.8 & 69.8 & 64.1 & 19.1 \\
 & $\pi_{maxQ}$ & 63.5$\substack{+\\-}$6.3 & \textbf{71.5$\substack{+\\-}$7.8} & 53.9$\substack{+\\-}$3.6 & 73.9$\substack{+\\-}$4.4 & 70.6$\substack{+\\-}$8.9 & 21.1$\substack{+\\-}$8.3 \\
 & $\pi_{avg}$ & 62.3$\substack{+\\-}$5.3 & 71.2$\substack{+\\-}$8.1 & 60.9$\substack{+\\-}$9.6 & 73.2$\substack{+\\-}$2.4 & 71.4$\substack{+\\-}$4.4 & 23.5$\substack{+\\-}$7.0 \\
 & \textbf{FusionNet} & \textbf{64.0$\substack{+\\-}$6.4} & 65.9$\substack{+\\-}$7.1 & 41.6$\substack{+\\-}$8.9 & \textbf{74.5$\substack{+\\-}$6.9} & \textbf{73.0$\substack{+\\-}$8.1} & 24.9$\substack{+\\-}$8.2 \\
\Xhline{2\arrayrulewidth} 
 & DET & 57.6 & 52.2 & \textbf{28.9} & 43.1 & 38.7 & 40.8 \\
 & PROB & 62.6 & 62.4 & 37.1 & 50.3 & 50.7 & 49.1 \\ \cline{2-8}
 & SAC & 62.7 & 64.9 & 41.5 & 49.7 & 50.3 & 50.8 \\
 & DDPG & 57.7 & 52.9 & 30.0 & 36.2 & 31.4 & 39.9 \\
\parbox[c][0pt][c]{\linewidth}{
  \centering
  \rotatebox[origin=c]{90}{
    \parbox{3cm}{\centering ISMRM}
  }
} & TD3 & 52.2 & 43.2 & 22.3 & 36.9 & 31.6 & \textbf{39.0} \\
 % & TRLF$_{S}$ & 61.1 & 58.2 & 32.1 & 49.1 & 51.1 & 57.3 \\
 & $\pi_{maxQ}$ & 62.8 & 65.1 & 41.6 & 50.2 & 49.8 & 48.8 \\
 & $\pi_{avg}$ & 62.8 & 59.5 & 30.1 & 46.7 & 45.4 & 49.1 \\
 & \textbf{FusionNet} & \tikzmark{CG ISMRM 1 top}\textbf{63.9} & \textbf{65.4} & 39.1\tikzmark{CG ISMRM 1 bottom} & \tikzmark{CST ISMRM 1 top}\textbf{52.6} & \textbf{54.3} & 51.2\tikzmark{CST ISMRM 1 bottom} \\ \cline{1-8}
 \hline
\end{tabular}
\label{table:exp:CG_CST}
% \DrawBox[ultra thick, draw=green]{CG hcp 1 top}{CG hcp 1 bottom}
% \DrawBox[ultra thick, draw=green]{CG hcp 2 top}{CG hcp 2 bottom}
% \DrawBox[ultra thick, draw=green]{CG ttoi 1 top}{CG ttoi 1 bottom}
% \DrawBox[ultra thick, draw=green]{CG ttoi 2 top}{CG ttoi 2 bottom}
% \DrawBox[ultra thick, draw=green]{CG ISMRM 1 top}{CG ISMRM 1 bottom}
% \DrawBox[ultra thick, draw=blue]{CG ISMRM 2 top}{CG ISMRM 2 bottom}
% \DrawBox[ultra thick, draw=green]{CST hcp 1 top}{CST hcp 1 bottom}
% \DrawBox[ultra thick, draw=green]{CST hcp 2 top}{CST hcp 2 bottom}
% \DrawBox[ultra thick, draw=green]{CST ttoi 1 top}{CST ttoi 1 bottom}
% \DrawBox[ultra thick, draw=green]{CST ttoi 2 top}{CST ttoi 2 bottom}
% \DrawBox[ultra thick, draw=green]{CST ISMRM 1 top}{CST ISMRM 1 bottom}
% % \DrawBox[ultra thick, draw=blue]{CST ISMRM 2 top}{CST ISMRM 2 bottom}
\end{table}

\textbf{Observations}: Comparative results in Tables \ref{table:exp:AF_CC}, \ref{table:exp:CG_CST} and qualitative results in Fig. \ref{fig:fusion-plot} reveal the following notable insights:
% the best result is highlighted in green and the second-best in blue. 

% \begin{enumerate}
\begin{itemize}%[label=--, leftmargin=*, itemsep=0.2em]
% above label, leftmargin, itemsep is commented by Ashutosh
    % \item FusionNet (Proposed) consistently ranks either first or second alongside PROB, showcasing best performance in ISMRM dataset CG and CST [Table \ref{table:exp:combined}].
    \item Our proposed FusionNet model consistently ranks best in terms of Dice score across all datasets and tracts.
    \item \textit{Note:} PROB was one of the four tractography algorithms used to generate the ground-truth streamlines in TractoInferno, which inherently gives it an advantage in evaluations involving this dataset. In contrast, the reference streamlines for the HCP dataset in \cite{wasserthal2018tractseg} were generated using iFOD2 \cite{tournier2010improved}, a close successor to PROB, which may cause PROB to systematically underperform on HCP. Therefore we report its performance only on the ISMRM dataset.
    
    \item Fusion policy (FusionNet) outperforms individual RL policies (SAC, TD3, DDPG) and their RL-Ensemble methods ($\pi_{maxQ}$ and $\pi_{avg}$). Individual RL policies show a high OR-OL tradeoff with SAC achieving highest OL with worst OR, and DDPG and TD3 achieving low OL and OR. The Ensemble’s Q-value approach yields suboptimal tracking, while FusionNet’s learned policy better balances OL-OR trade-offs, resulting in the best Dice.
    \item FusionNet* (FusionNet tested on TractSeg masks \cite{wasserthal2018tractseg}) surpasses TractSeg across all HCP tracts, e.g., achieving a Dice score of 74.1 versus TractSeg’s 70.9 for AF, demonstrating superior tracking even within TractSeg’s masks.
    \item FusionNet* and TractSeg achieve highest scores on HCP, as TractSeg was trained on HCP data to obtain those masks which were also involved in making reference streamlines of HCP. However, masks from~\cite{joshi2024tract} were available for all three datasets (TractoInferno, HCP, and ISMRM), preferred for a consistent comparison across these datasets. FusionNet’s superior performance within both masks demonstrates its robustness to variations in mask quality for tract-specific tractography.
    
    \item Spurious connections made by SAC and PROB in left CG and CC\_Oc encircled in red (Fig. \ref{fig:fusion-plot}(a,b)) show lower OR by our method, consistent with the scores in Tables \ref{table:exp:AF_CC} and \ref{table:exp:CG_CST}.
    \item In Fig. \ref{fig:fusion-plot}(a), PROB struggles to reconstruct upper portion of CST right, at pyramidal decussation (intersection of left \& right CST) compared to FusionNet.
\end{itemize}

\textbf{\textit{Note:}} While simpler ensembling methods ($\pi_{maxQ}$ and $\pi_{Avg}$) achieve lower scores than FusionNet, more complex RL ensembling techniques often introduce significant stability and scalability challenges. TractRLFusion addresses this challenge by learning a robust fusion policy in a data-driven manner, which simplifies integration compared to traditional RL ensemble methods. Moreover, while we demonstrate our method using three policies (SAC, DDPG, and TD3) to handle the overlap-overreach trade-off (observed in these individual policies), the proposed framework is modular and scalable. It can be easily extended to include additional policies to further capture diverse tracking behaviors and improve generalization as evident in Fig. \ref{fig:fusion-plot}.

\subsection{Ablation Study}

This section presents an ablation study of the two major components of our work, namely, Episodic Data Selection (EDS) and Multi-Critic Policy Fine-Tuning (MCPFT).

EDS involves both within-policy and across-policy selection of training trajectories, as described in Section \ref{subsec:EDS}.
While across-policy selection ensures a diverse mix of trajectories covering all three policies across different brain regions, we conduct an ablation to assess the contribution of within-policy selection. Hence we compared the full EDS method (which includes both selection steps) against a variant that uses only across-policy selection. These trajectory selection strategies are denoted as EDS and $EDS_{across-\pi}$, respectively, in Table \ref{table:abl:eds}. This comparison is conducted across both the pre-training (PT) and fine-tuning (FT) stages of FusionNet.

% \begin{table}[hbt]
\begin{table}[H]
\centering
\caption[Ablation study of the EDS module highlighting the impact of different training data selection strategies]{Average performance (\%) obtained using Fusion-Net on test subjects from the TtoI. The table compares different pretraining (PT) and finetuning (FT) strategies and highlights the impact of EDS.}
\fontsize{10}{7}\selectfont
\begin{tabular}{|p{0.8cm}|p{1.1cm}p{1.1cm}p{1.15cm}|p{1.1cm}p{1.1cm}p{1.15cm}|p{1.1cm}p{1.1cm}p{1.15cm}|}
\hline
\textbf{Tract} & \multicolumn{3}{c|}{(i){\makecell{\textbf{EDS$_{\text{across}-\pi}$} \\ \textbf{(PT \& FT)}}}} & \multicolumn{3}{c|}{(ii){\makecell{\textbf{EDS} \\ \textbf{(PT \& FT)}}}} & \multicolumn{3}{c|}{(iii)\makecell{\textbf{EDS$_{\text{across}-\pi}$} \textbf{PT} \\  \textbf{\& EDS FT}}} \\
\cline{2-10}
& Dice \textuparrow & OL \textuparrow & OR \textdownarrow & Dice \textuparrow & OL \textuparrow & OR \textdownarrow & Dice \textuparrow & OL \textuparrow & OR \textdownarrow \\
\hline
PYT & 70.4$\substack{+\\-}$5.5 & 67.3$\substack{+\\-}$6.6 & 23.4$\substack{+\\-}$8.4 & 58.6$\substack{+\\-}$1.9 & 45.7$\substack{+\\-}$3.1 & 8.9$\substack{+\\-}$5.2 & 70.9$\substack{+\\-}$5.1 & 68.3$\substack{+\\-}$7.2 & 23.9$\substack{+\\-}$8.5 \\
CG  & 59.2$\substack{+\\-}$7.6 & 58.6$\substack{+\\-}$6.3 & 39.4$\substack{+\\-}$8.2 & 42.9$\substack{+\\-}$6.7 & 32.1$\substack{+\\-}$7.0 & 10.7$\substack{+\\-}$6.3 & 61.4$\substack{+\\-}$8.2 & 59.5$\substack{+\\-}$6.6 & 34.5$\substack{+\\-}$9.3 \\
AF  & 47.3$\substack{+\\-}$8.7 & 39.7$\substack{+\\-}$8.1 & 27.3$\substack{+\\-}$9.8 & 23.3$\substack{+\\-}$4.0 & 14.4$\substack{+\\-}$2.1 & 5.3$\substack{+\\-}$1.5 & 48.7$\substack{+\\-}$8.8 & 43.5$\substack{+\\-}$8.5 & 38.9$\substack{+\\-}$9.4 \\
CC  & 67.7$\substack{+\\-}$4.9 & 74.1$\substack{+\\-}$5.3 & 44.2$\substack{+\\-}$6.9 & 60.6$\substack{+\\-}$3.9 & 50.3$\substack{+\\-}$2.8 & 14.9$\substack{+\\-}$2.2 & 68.6$\substack{+\\-}$4.0 & 70.2$\substack{+\\-}$5.4 & 36.4$\substack{+\\-}$7.2 \\
\hline
\end{tabular}
\label{table:abl:eds}
\end{table}

Table \ref{table:abl:eds} reports the average scores across five test subjects from the TractoInferno dataset under three trajectory selection configurations:

(i) $EDS_{across-\pi}$ (PT \& FT): Both pre-training and fine-tuning trajectories are selected using only across-policy selection.

(ii) EDS (PT \& FT): Both stages utilize the full Episodic Data Selection (EDS) method, incorporating both within-policy and across-policy trajectory selection.

(iii) $EDS_{across-\pi}$ PT \& EDS FT: Pre-training uses across-policy selection only, while fine-tuning applies the full EDS method. This setting corresponds to the proposed EDS configuration used in FusionNet.

Notably, in all 3 configurations, pre-training uses a mixture of trajectories from different tracts, whereas fine-tuning is performed on trajectories from the same tract. It can be observed that performing both pretraining and finetuning on EDS data yields the poorest performance. This is slightly improved when using $EDS_{across-\pi}$ for both stages. The best performance is achieved in case (iii), highlighting the effectiveness of our trajectory selection process.

\begin{table}[H]
\centering
\caption[Ablation study of the MCPFT module]{Average performance metrics (in \%) obtained using Tract-RLFormer highlight the impact of the MCPFT on test subjects from the TractoInferno dataset.}
\begin{tabular}{ |p{1cm}|p{1.4cm}p{1.4cm}p{1.4cm}|p{1.4cm}p{1.4cm}p{1.4cm}|  }
 \hline
  \textbf{Tract}  &\multicolumn{3}{c|}{\textbf{Without MCPFT}} &\multicolumn{3}{c|}{\textbf{With MCPFT}} \\
  \cline{2-7}  % Fixed: Removed misplaced '&' before this line
  & Dice \textuparrow & OL \textuparrow & OR \textdownarrow & Dice \textuparrow & OL \textuparrow & OR \textdownarrow \\  
  \cline{5-7}
  \hline
   % PYT & 70.883 & 68.232 & 23.98 & 73.322 & 70.787 & 23.043 \\
   PYT & 70.9$\substack{+\\-}$5.1 & 68.3$\substack{+\\-}$7.2 & 23.9$\substack{+\\-}$8.5 & 74.5$\substack{+\\-}$6.9 & 73.0$\substack{+\\-}$8.1 & 24.9$\substack{+\\-}$8.2 \\
   % CG  & 61.359 & 59.489 & 34.541 & 63.793 & 63.539 & 36.594 \\
   CG  & 61.4$\substack{+\\-}$8.2 & 59.5$\substack{+\\-}$6.6 & 34.5$\substack{+\\-}$9.3 & 64.0$\substack{+\\-}$6.4 & 65.9$\substack{+\\-}$7.1 & 41.6$\substack{+\\-}$8.9 \\
   % AF  & 48.729 & 43.449 & 38.99  & 52.07 & 45.291 & 27.372 \\
   AF  & 48.7$\substack{+\\-}$8.8 & 43.5$\substack{+\\-}$8.5 & 38.9$\substack{+\\-}$9.4 & 53.2$\substack{+\\-}$8.9 & 51.3$\substack{+\\-}$4.3 & 39.1$\substack{+\\-}$9.3 \\
   CC  & 68.6$\substack{+\\-}$4.0 & 70.2$\substack{+\\-}$5.4 & 36.4$\substack{+\\-}$7.2 & 72.2$\substack{+\\-}$3.2 & 73.2$\substack{+\\-}$6.1 & 32.3$\substack{+\\-}$6.6 \\
   % CC  & 68.575 & 70.134 & 36.456  & 72.167 & 73.213 & 32.397 \\
   \hline
\end{tabular}
\label{table:abl:mcpft}
\end{table}

Moreover, Table \ref{table:abl:mcpft} shows that further finetuning of our FusionNet using MCPFT effectively reduces overreach (OR) while increasing Overlap (OL), a desirable outcome in tractography, leading to a higher Dice score.

\section{Summary}

TractRLFusion is a policy fusion framework for White Matter Tractography, allowing efficient and effective tracking owing 
to its ability to fuse complementary strengths of different actor-critic frameworks. The ablation experiments highlight the effectiveness of our two-stage data curation, where a strong baseline policy is first developed using a combination of broad and anatomically-aware data selection (EDS). This policy is then refined by the MCPFT process. TractRLFusion robustly generalizes well, being trained only on TractoInferno and inferred on HCP and ISMRM.
Finally, TractRLFusion, powered by FusionNet, lays the groundwork for a foundational tractography model, with a potential to span additional datasets and incorporate diverse specialist policies, advancing the field of White Matter Tractography.
\chapter{Conclusion and Future Work}\label{conclusion}

White matter fiber tractography is a core component of modern neuroimaging, enabling detailed mapping of structural brain connectivity that is foundational to both clinical decision-making and neuroscience research. This thesis contributes to advancing reinforcement-learning–based tractography by improving the performance, robustness, and tract-specific adaptability of RL tracking policies, by proposing deep learning based methods that demonstrate robust tractography and strong generalization.

We introduced two hybrid frameworks that leverage GPTs alongside reinforcement learning for tract-specific tractography: Tract-RLFormer, designed for RL policy refinement, and TractRLFusion, aimed at policy fusion. Central to both frameworks is the Mask Refinement Module (MRM), a tract-specific mask generator that facilitates targeted training and reliable white matter tracking.
% reliable bundle reconstruction.

In Tract-RLFormer, pretraining was conducted using a broad and diverse collection of trajectories across multiple tracts, while fine-tuning focused on a single tract of interest. This separation allowed the model to first acquire generalizable tracking knowledge and then specialize effectively. TractRLFusion preserved this two-stage training strategy, but enhanced tract-specificity by selecting fine-tuning trajectories based on their shape similarity to atlas reference fibers. This ensured more precise alignment with the geometry of the target tract and improved downstream tractography accuracy.

Across both frameworks, broad pretraining proved valuable for generalization, whereas tract-specific fine-tuning enabled focused adaptation to the intended bundle. Ablation studies, examining alternative data-selection strategies at each stage, further validated the effectiveness of this design. Quantitative and qualitative evaluations across previously unseen datasets consistently demonstrated robust generalization, improved anatomical plausibility, and reduced false positives, without relying on ground-truth data for training. These conclusions are consistent with insights from foundation-model research \cite{bommasani2021opportunities}, where large-scale pretraining combined with targeted specialization yields strong performance across downstream tasks.

The major contributions of this thesis are summarized below:

\begin{itemize}
    \item We introduced the Mask Refinement Module (MRM) to generate tract-specific masks, accounting for inter-subject variability, enabling anatomically precise targeted tractography.
    \item Tract-RLFormer successfully refined a base RL policy (TD3) and as a GPT-based general backbone for tract-specific agents, Tract-RLFormer also provided the architectural basis for the subsequent fusion framework, TractRLFusion.
    \item TractRLFusion, backed by Episodic Data Selection (EDS), FusionNet, and Multi-critic policy finetuning (MCPFT) demonstrated effective data-driven fusion of TD3, SAC, and DDPG policies, addressing the persistent challenge of False positive-False negative trade-off in tractography. Its scalable design can incorporate additional actor-critic policies, offering a flexible mechanism to capture diverse tracking behaviors.
    \item Extensive experiments on seven major white-matter bundles across the TractoInferno, HCP, and ISMRM datasets confirmed the robustness, generalization capability, and practical utility of both proposed frameworks.

\end{itemize}

\textbf{Future Directions}: 
While this work focused on healthy participants, an important extension is to evaluate the proposed frameworks on diseased or lesion-bearing datasets. Such validation would further establish the clinical relevance of RL-based tracking agents.

Another promising direction is the systematic exploration of tractography hyperparameters, particularly tract-specific masks. Beyond model architecture, the choice of masks, seeding strategies, and stopping criteria substantially influences reconstruction quality. Improving or learning these components, potentially jointly with the RL policy, could yield more reliable and anatomically consistent tractograms.

Finally, although TractRLFusion demonstrated successful fusion of three widely used actor–critic policies, its architecture is inherently extensible. Incorporating additional policies may capture more diverse tracking strategies and further mitigate errors due to erroneous paths or missed fiber paths.

In summary, this thesis demonstrates that combining broad pretraining, tract-specific adaptation, and multi-policy fusion offers a powerful and flexible approach for targeted white-matter tractography. We hope that these contributions will motivate further research at the intersection of reinforcement learning, generative modeling, and neuroimaging, ultimately improving the reliability and clinical utility of white matter fiber tractography.

\bibliographystyle{IEEEtran}

%\chapter*{Publications from thesis}
%$*$ -- Publications directly related to the thesis.
%\renewcommand{\baselinestretch}{0.80}
%\renewcommand\refname{}
\chapter*{List of Publications}
\addcontentsline{toc}{chapter}{List of Publications}
%\section*{Publications Based on the Thesis}

\section*{Publications included in thesis}

% \begin{flushleft}
% {\large \bf Peer-Reviewed Conferences}\\
% \end{flushleft}

% \begin{enumerate}\itemsep0.5em
% 	\item[1] \begin{normalsize}	\textbf{Muneeswaran P}, Jyoti, and Sriram Kailasam, \enquote{A hybrid partitioning strategy
% for distributed FCA,}  \textit{in Proceedings of the 15$^{th}$ International
% Conference on Concept Lattices and Their Applications, Tallinn, Estonia,
% June 29-July 1, 2020}, pp. 71–82, 2020.
% \end{normalsize} 
% \end{enumerate}

\begin{flushleft}
% {\large \bf Refereed Journals}\\
\end{flushleft}

\begin{enumerate}\itemsep0.5em

\item[1] \begin{normalsize} \textbf{Ankita Joshi}, Ashutosh Sharma, Anoushkrit Goel, Ranjeet Ranjan Jha, Chirag Ahuja, Arnav Bhavsar, Aditya Nigam, \textbf{\enquote{TractRLFusion: A GPT-based Multi-Critic Policy Fusion Framework for Fiber Tractography}} \textbf{Accepted} in 2026 IEEE ${23^{rd}}$
International Symposium on Biomedical Imaging \textbf{(ISBI-2026)}.
% to British Machine Vision Conference (\textbf{BMVC-2025}). 
% \textbf{Accepted} in ${28^{th}}$ International Conference on Medical Image Computing
% and Computer Assisted Intervention \textbf{(MICCAI-2025)}.
\end{normalsize}

\item[2] \begin{normalsize} \textbf{Ankita Joshi}, Ashutosh Sharma, Anoushkrit Goel, Ranjeet Ranjan Jha, Chirag Kamal Ahuja, Arnav Bhavsar, Aditya Nigam,  \textbf{\enquote{Tract-RLFormer: A Tract-Specific RL policy based Decoder-only Transformer Network}} in ${27^{th}}$ International Conference on Pattern Recognition \textbf{(ICPR-2024)}.
\end{normalsize}

% \item[3] \begin{normalsize} Anoushkrit Goel, Bipanjit Singh, \textbf{Ankita Joshi}, Ranjeet Ranjan Jha, Chirag Ahuja, Aditya Nigam, Arnav Bhavsar,  \textbf{\enquote{TractoEmbed: Modular Multi-level Embedding framework for white matter tract segmentation,}}  \textbf{Accepted} in ${27^{th}}$ International Conference on Pattern Recognition \textbf{(ICPR-2024)}.
% \end{normalsize}

% \item[4] \begin{normalsize} Ranjeet Ranjan Jha, Hritik Gupta, S Pathak, \textbf{Ankita Joshi}, W Schneider, BVR Kumar, A Bhavsar, A Nigam, \textbf{\enquote{PA-GAN: Parallel Attention Based GAN for Enhancement of fODF}} \textbf{Accepted} in 2023 IEEE International Symposium on Biomedical Imaging \textbf{(ISBI-2023)}.
% \end{normalsize}

\end{enumerate}

\section*{Others}

\begin{enumerate}\itemsep0.5em

\item[3] \begin{normalsize} Anoushkrit Goel, Simroop Singh, \textbf{Ankita Joshi}, Ranjeet Ranjan Jha, Chirag Ahuja, Aditya Nigam, Arnav Bhavsar, \textbf{\enquote{TrackletGPT: A Language-like GPT Framework for White Matter Tract Segmentation}} \textbf{Accepted} in 2026 IEEE ${23^{rd}}$ International Symposium on Biomedical Imaging \textbf{(ISBI-2026)}.
\end{normalsize}

\item[4] \begin{normalsize} Anoushkrit Goel, Simroop Singh, \textbf{Ankita Joshi}, Ranjeet Ranjan Jha, Chirag Ahuja, Aditya Nigam, Arnav Bhavsar, \textbf{\enquote{TractoGPT: A GPT Architecture for White Matter Tract Segmentation}} in 2025 IEEE ${22^{nd}}$ International Symposium on Biomedical Imaging \textbf{(ISBI-2025)}.
\end{normalsize}

\item[5] \begin{normalsize} Anoushkrit Goel, Bipanjit Singh, \textbf{Ankita Joshi}, Ranjeet Ranjan Jha, Chirag Ahuja, Aditya Nigam, Arnav Bhavsar,  \textbf{\enquote{TractoEmbed: Modular Multi-level Embedding framework for white matter tract segmentation}} in ${27^{th}}$ International Conference on Pattern Recognition \textbf{(ICPR-2024)}.
\end{normalsize}

\item[6] \begin{normalsize} Ranjeet Ranjan Jha, Hritik Gupta, S Pathak, \textbf{Ankita Joshi}, W Schneider, BVR Kumar, A Bhavsar, A Nigam, \textbf{\enquote{PA-GAN: Parallel Attention Based GAN for Enhancement of fODF}} in 2023 IEEE ${20^{th}}$ International Symposium on Biomedical Imaging \textbf{(ISBI-2023)}.
\end{normalsize}

\end{enumerate}

\bibliography{mybib}

\end{document}